\documentclass{article}
\usepackage[nonatbib,final]{neurips_2020}

\usepackage[utf8]{inputenc} % allow utf-8 input
\usepackage[T1]{fontenc}    % use 8-bit T1 fonts
\usepackage{url}            % simple URL typesetting
\usepackage{booktabs}       % professional-quality tables
\usepackage{amsfonts}       % blackboard math symbols
\usepackage{nicefrac}       % compact symbols for 1/2, etc.
\usepackage{microtype}      % microtypography
\usepackage{multirow}

\usepackage[export]{adjustbox}
\usepackage{graphicx}
\usepackage{wrapfig}
\usepackage{comment}
\usepackage{amsmath,amssymb,bbm} % define this before the line numbering.
\usepackage{color}
\usepackage{lib}
\usepackage{cite}
\usepackage{booktabs}
\usepackage{alias}
\usepackage{xcolor}
\usepackage[pagebackref=true,breaklinks=true,letterpaper=true,colorlinks,citecolor=blue,bookmarks=false]{hyperref}
\usepackage[labelfont=bf]{caption}

\usepackage{ amssymb }

\newcommand{\blfootnote}[1]{{\renewcommand\thefootnote{}\footnotetext{#1}}}

\author{\textbf{Matthew Chang \quad \quad \quad Arjun Gupta \quad \quad \quad Saurabh Gupta} \\ 
University of Illinois at Urbana-Champaign \\
\texttt{\{mc48, arjung2, saurabhg\}@illinois.edu}}

\begin{document}
\blfootnote{Project website with code, models, and videos:
\url{https://matthewchang.github.io/value-learning-from-videos/}.}
% \title{Learning Actionable Affordances by Watching Egocentric Videos}
% \title{Learning to Navigate to Semantic Targets by Watching Egocentric Videos}
\title{Semantic Visual Navigation by \\ Watching YouTube Videos}
% \title{Learning Spatial Common-sense by Watching Egocentric Videos} 
% Has the issue that we are only learning one specific kind of common sense.
% \title{Learning Value Functions from Egocentric Videos via Latent Action
% Grounding}

\maketitle

\setlength{\abovedisplayskip}{3pt}
\setlength{\belowdisplayskip}{3pt}
\begin{abstract}
Semantic cues and statistical regularities in real-world environment layouts
can improve efficiency for navigation in novel environments. This paper learns
and leverages such semantic cues for navigating to objects of interest in novel
environments, by simply watching YouTube videos. This is challenging because
YouTube videos don't come with labels for actions or goals, and may not even
showcase optimal behavior. Our method tackles these challenges through
the use of Q-learning on pseudo-labeled transition quadruples (image, action,
next image, reward). We show that such off-policy Q-learning from passive data
is able to learn meaningful semantic cues for navigation. These cues, when used
in a hierarchical navigation policy, lead to improved efficiency at the
ObjectGoal task in visually realistic simulations. We observe a relative improvement of $15-83\%$ over end-to-end RL, behavior cloning, and classical methods, while using minimal direct interaction.

\end{abstract}

\section{Introduction}
\seclabel{intro}
Consider the task of finding your way to the bathroom while at a new
restaurant.
% that just opened down the street. 
As humans, we can efficiently solve such
tasks in novel environments in a zero-shot manner. We leverage common sense
patterns in the layout of environments, which we have built from our past experience
of similar environments.
% As human, we leverage common sense patterns
% in environment layouts to efficiently solve such tasks in a zero-shot manner.
For finding a bathroom, such cues will be that they are typically towards the
back of the restaurant, away from the main seating area, behind a corner, and
might have signs pointing to their locations (see \figref{teaser}). 
Building computational systems that can similarly leverage such semantic
regularities for navigation has been a long-standing goal.

% Advances presented in this work enables use of videos to learn such
% semantic regularities. This drastically reduces the sample complexity for
% acquiring such policies. In as few as 100K interaction samples, we can learn
% effective policies for semantic visual navigation. 

Hand-specifying what these semantic cues are, and how they should be used by a
navigation policy is challenging.
% It is challenging to apriori hand-define the set of these semantic
% regularities, as well as how to use them for navigation.
Thus, the dominant paradigm is to directly learn what these cues are, and how
to use them for navigation tasks, in an end-to-end manner via reinforcement
learning. 
% Existing paradigms circumvent these challenges by employing end-to-end policy
% learning through reinforcement learning. 
% Such end-to-end learning through
% direct interaction can automatically learn the appropriate semantic cues and
% how to use them to efficiently solve such navigation tasks. 
While this is a promising approach to this problem, it is sample inefficient,
and requires many million interaction samples with dense reward signals to
learn reasonable policies. 

\begin{figure}[t]
\centering
\insertH{0.29}{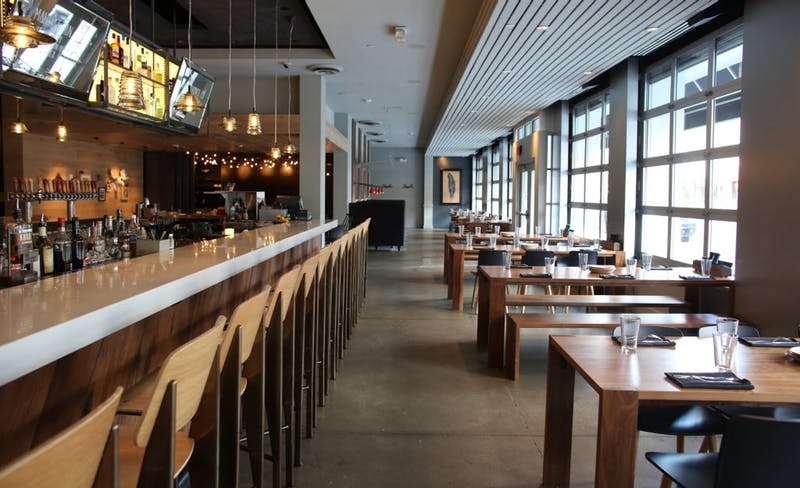}
\insertH{0.29}{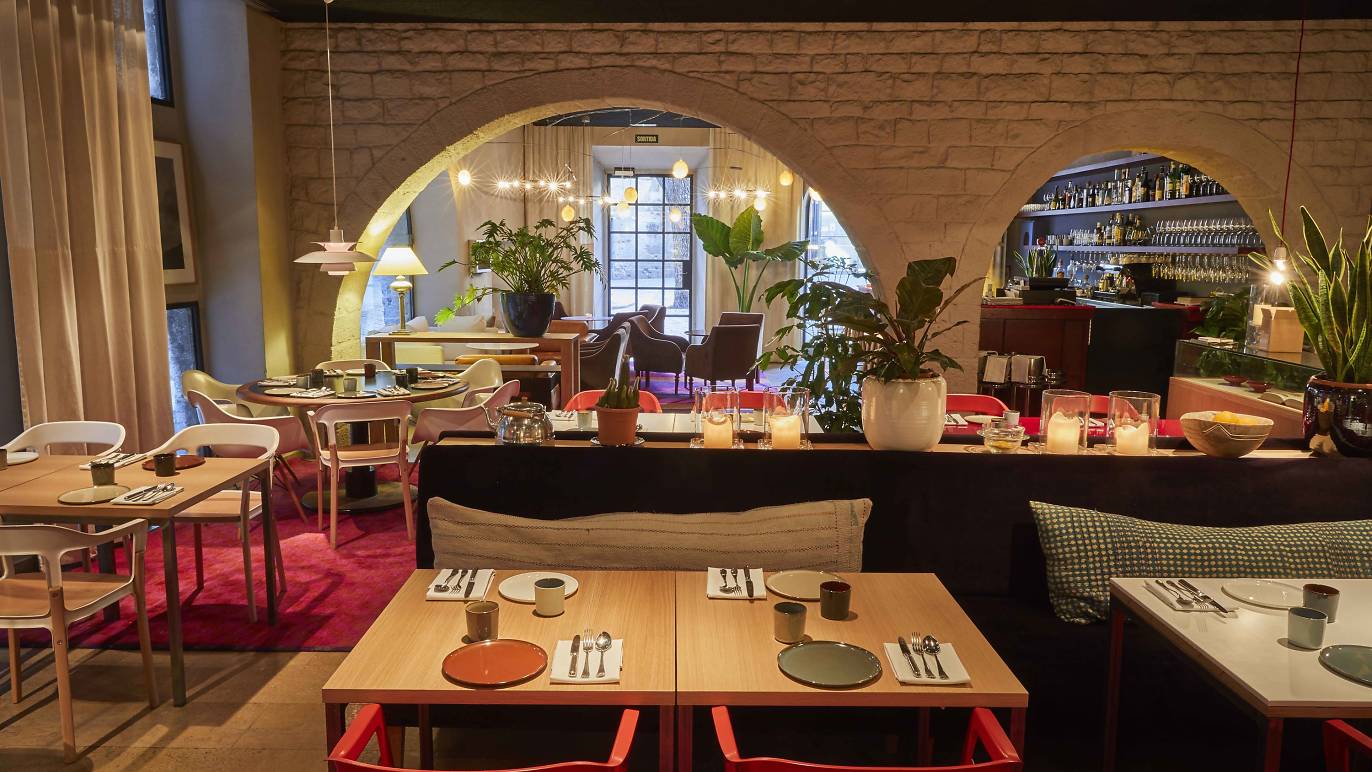}
\caption{\textbf{Semantic Cues for Navigation.} Even though you don't see a
restroom, or a sign pointing to one in either of these images, going straight 
ahead in the left image is more likely to lead to a restroom than going
straight in the right image. This paper seeks to learn and levarage such
semantic cues for finding objects in novel environments, by watching egocentric
YouTube videos.}
\figlabel{teaser}
\end{figure}

But, is this the most direct and efficient way of learning about such semantic
cues? At the end of the day, these semantic cues are just based upon
spatial consistency in co-occurrence of visual patterns next to one another.
That is, if there is always a bathroom around the corner towards the back of
the restaurant, then we can learn to find this bathroom, by simply finding
corners towards the back of the restaurant. This observation motivates our
work, where we pursue an alternate paradigm to learn semantic cues for
navigation: learning about this spatial co-occurrence in indoor environments
through video tours of indoor spaces. People upload such videos to YouTube (see
project video) % \cite{realestate} as an example) 
to showcase real estate for renting and
selling. We develop techniques that leverage such YouTube videos to learn
semantic cues for effective navigation to semantic targets in indoor home
environments (such as finding a \textit{bed} or a \textit{toilet}).

% learning effective navigation policies 
Such use of videos presents three unique and novel challenges, that don't
arise in standard learning from demonstration.  Unlike robotic demonstrations,
videos on the Internet don't come with any action labels. This precludes
learning from demonstration or imitation learning. Furthermore, goals and
intents depicted in videos are not known, \ie, we don't apriori know what each
trajectory is a demonstration for. Even if we were to label this somehow, the
depicted trajectories may not be optimal, 
% for semantic goal reaching
a critical assumption in learning from demonstration~\cite{schaal1997learning}
or inverse reinforcement learning~\cite{ng2000algorithms}.

Our formulation, \textit{Value Learning from Videos} or \textit{VLV},
tackles these problems by \textbf{a)} using pseudo action labels obtained by running an
inverse model, and \textbf{b)} employing Q-learning to learn from video sequences that
have been pseudo-labeled with actions. We follow work from Kumar
\etal~\cite{kumar2019learning} and use a small number of interaction samples
($40K$) to acquire an inverse model. This inverse model is used to
pseudo-label consecutive video frames with the action the robot would have taken to
induce a similar view change. This tackles the problem of missing actions.
Next, we obtain goal labels by classifying video frames based on whether or not they
contain the desired target objects. Such labeling can be done using
off-the shelf object detectors. Use of Q-learning~\cite{watkins1989learning}
with consecutive frames, intervening actions (from inverse model), and rewards
(from object category labels), leads to learning \textit{optimal} Q-functions
for reaching goals~\cite{watkins1989learning, sutton1998rli}. We take the
maximum Q-value over all actions, to obtain value functions. These value
functions are exactly $\gamma^{s}$, where $s$ is the number of steps to the
nearest view location of the object of interest ($\gamma$ is the Q-learning
discount factor). These value functions implicitly learn semantic cues. An
image looking at the corner towards the back of the restaurant will have a
higher value (for \textit{bathroom} as the semantic target) than an image
looking at the entrance of the restaurant. These learned value functions when
used with a hierarchical navigation policy, efficiently guide locomotion
controllers to desired semantic targets in the environment. 

Learning from such videos can have many advantages, some of which
address limitations of learning from direct interaction (such as via RL).
Learning from direct interaction suffers from high sample complexity
(the policy needs to discover high-reward trajectories which may be hard to
find in sparse reward scenarios) and poor generalization (limited number of
instrumented physical environments available for reward-based learning, or
sim2real gap).
Learning from videos side-steps both these issues. 
% Our experiments in visually realistic simulations show a 
We observe a $47-83\%$ relative improvement in performance over
RL and imitation learning methods, while also improving upon strong classical
methods.

\section{Related Work}
\seclabel{related}
This paper tackles semantic visual navigation in novel environments. Our
proposed solution is a hierarchical policy that employs value functions learned
from videos. We survey different navigation tasks, the different
representations used to tackle them, and the different training methodologies
employed to build those representations. 
% \sg{\cite{druon2020visual, nguyen2019reinforcement, yang2018visual,
% das2018embodied, chen2018learning}}

\noindent \textbf{Navigation Tasks.} Navigation tasks take many
forms~\cite{anderson2018evaluation}, but can largely be grouped into
two categories based on whether they require exploration or not. Finding paths
in known environments~\cite{zhu2017target}, or going to a known relative offset
in a previously unknown environment~\cite{gupta2019cognitive}, do not require
very much exploration. On the other hand, tasks such as finding an
object~\cite{gupta2019cognitive} (or a given image
target~\cite{chaplot2020neural}) in a novel
environment, or exhaustively mapping one~\cite{chaplot2020learning,
chen2018learning}, require
exploration and are thus more challenging. Our down-stream task of finding
objects in previously unseen novel environments falls into this second
category.  Most current work~\cite{nguyen2019reinforcement, yang2018visual,
druon2020visual, gupta2019cognitive} on this task employ end-to-end,
interaction-heavy learning to get at necessary semantic cues. Our work
instead seeks to mine them from videos with minimal active interaction. 

\noindent \textbf{Representations.} Solving navigation tasks, requires building
and maintaining representations for space. These range from explicit metric
maps~\cite{elfes1989using, wurm2010octomap, thrun2000probabilistic} or
topological representations~\cite{konolige2010view, savinov2018semi,
chaplot2020neural}, to more abstract learned implicit
representations~\cite{mirowski2017learning}. Such learned representations can
effectively learn about semantic cues. Research has also focused on making
classical metric and topological representations more semantic: explicitly by
storing object detector or scene classifier
outputs~\cite{bowman2017probabilistic, kuipers1993robot, pronobis2011semantic,
henriques2018mapnet, gordon2018iqa, mousavian2019visual, wu2019bayesian}, or
implicitly by storing abstract learned feature vectors useful for the
end-task~\cite{gupta2019cognitive}. In our work, we use a hybrid topological
and metric representation that incorporates implicit semantic information. Our
focus is on investigating alternate ways of learning such semantic information. 

\noindent \textbf{Hierarchical Policies.} Researchers have pursued many
different hierarchical policies~\cite{barto2003recent} for navigation: no
hierarchy~\cite{mirowski2017learning}, macro-actions~\cite{zhu2017target,
gupta2019cognitive}, low-level controllers~\cite{bansal2019combining,
kaufmann2019beauty}, and sub-policies~\cite{chaplot2020learning, gordon2018iqa,
das2018neural}. In particular, Chaplot \etal\cite{chaplot2020learning, chaplot2020neural}
decompose exploration policies into a global policy, for high-level
semantic reasoning, and a local policy, for low-level execution to
achieve short-term goals produced by the global policy.  We follow a similar
decomposition, but tackle a different task (object goal), and investigate
learning from unlabeled passive data \vs active interaction or strong
supervision.

\noindent \textbf{Training Methodology.} Different papers pursue different
strategies for training navigation policies: no
training~\cite{thrun2000probabilistic}, supervised learning for collision
avoidance~\cite{gandhi2017learning, giusti2015machine}, behavior cloning,
DAgger~\cite{ross2011reduction, gupta2019cognitive, kumar2018visual},
reinforcement learning with sparse and dense
rewards~\cite{mirowski2017learning, zhu2017target, chaplot2020learning,
sadeghi2017cad2rl, wijmans2020dd, sadeghi2019divis}, and combinations of imitation and
RL~\cite{das2018embodied, chen2018learning, rajeswaran2017learning}. In
contrast, this paper designs a technique to derive navigation policies by
watching YouTube videos. This is most similar to work from Kumar
\etal~\cite{kumar2019learning} that studies how to learn low-level locomotion
sub-routines from \textit{synthetic} videos. In contrast, we learn high-level
semantic cues from actual YouTube videos.

\noindent \textbf{Learning for Acting from Videos.} Learning about
affordances~\cite{fouhey2012people},
state-transitions~\cite{isola2015discovering, alayrac2017joint}, and
task-solving procedures~\cite{Damen2018EPICKITCHENS}, with the goal of aiding
learning for robots, is a long-standing goal in computer vision. Our work is
also a step in this direction, although our output is directly useful for building navigation policies, and our experiments demonstrate this.

% \cite{pfei, pfeiffer2017perception}.
\noindent \textbf{Learning without Action Labels.} A number of recent papers
focus on learning from observation-only (or state-only) demonstrations (\ie
demonstrations without action labels). Some works focus on directly learning
policies from such data~\cite{torabi2018behavioral, torabi2019generative,
gangwani2020state, edwards2019imitating,
schmeckpeper2019learning,DBLP:journals/corr/abs-1906-07372}, while others focus
on extracting a reward function for subsequent policy learning through
RL~\cite{edwards2019perceptual, aytar2018playing,
DBLP:journals/corr/LiuGAL17, DBLP:conf/icra/SermanetLCHJSLB18,
DBLP:conf/rss/SermanetXL17,DBLP:journals/corr/abs-1808-00928}.
All of these works focus on learning a policy for the \textit{same} task in the
\textit{same} environment that is depicted in the observation-only
demonstrations (with the exception of Gangwani \etal~\cite{gangwani2020state}
who show results in MDPs with different transition dynamics). Our work relaxes
both these assumptions, and we are able to use video sequences to derive cues
that aid solving novel tasks in novel environments.

% - \cite{torabi2019generative} modifies IRL objective to instead use state
%   transitions as opposed to state action pairs. Experiments still show case
%   results in settings where policy is being trained in the same MDP as
%   demonstrations. \\
% - \cite{gangwani2020state} modify this and sow results in MDPs
% with different transition dynamics. Our work here focuses on the setting where
% we do not assume access to the demonstration MDPs. \\

\section{Proposed Approach}
\seclabel{overview}
The final task we tackle is that of reaching semantic goals in a novel
environment, \ie, at test time we will place the agent in a novel environment
and measure how efficiently it can find common house-hold objects (bed, chair,
sofa, table and toilets).

\textbf{Overview.} We design a 2-level hierarchical policy. The
\textit{high-level policy} incrementally builds a topological graph and uses
\textit{semantic reasoning} to pick promising directions of exploration. It
generates a short-term goal (within $2m$) for the \textit{low-level policy},
that achieves it or returns that the short-term goal is infeasible. This
process is repeated till the agent reaches its goal. We 
% provide an overview of this process in \figref{nav-policy} and
describe the details of this hierarchical policy in Supplementary
\secref{nav-policy}.
%Section S1.
Our central contribution is the procedure for learning 
the semantic reasoning function, which we call a \textit{value} function
(following RL terminology~\cite{sutton1998rli}), for the
high-level policy from videos, and we describe this next.
% first in \secref{vfv}.
% (\secref{grounding} and \secref{value}).

\subsection{Value Learning from Videos}
\seclabel{vfv}
\begin{figure}[t]
\centering
\insertW{1.0}{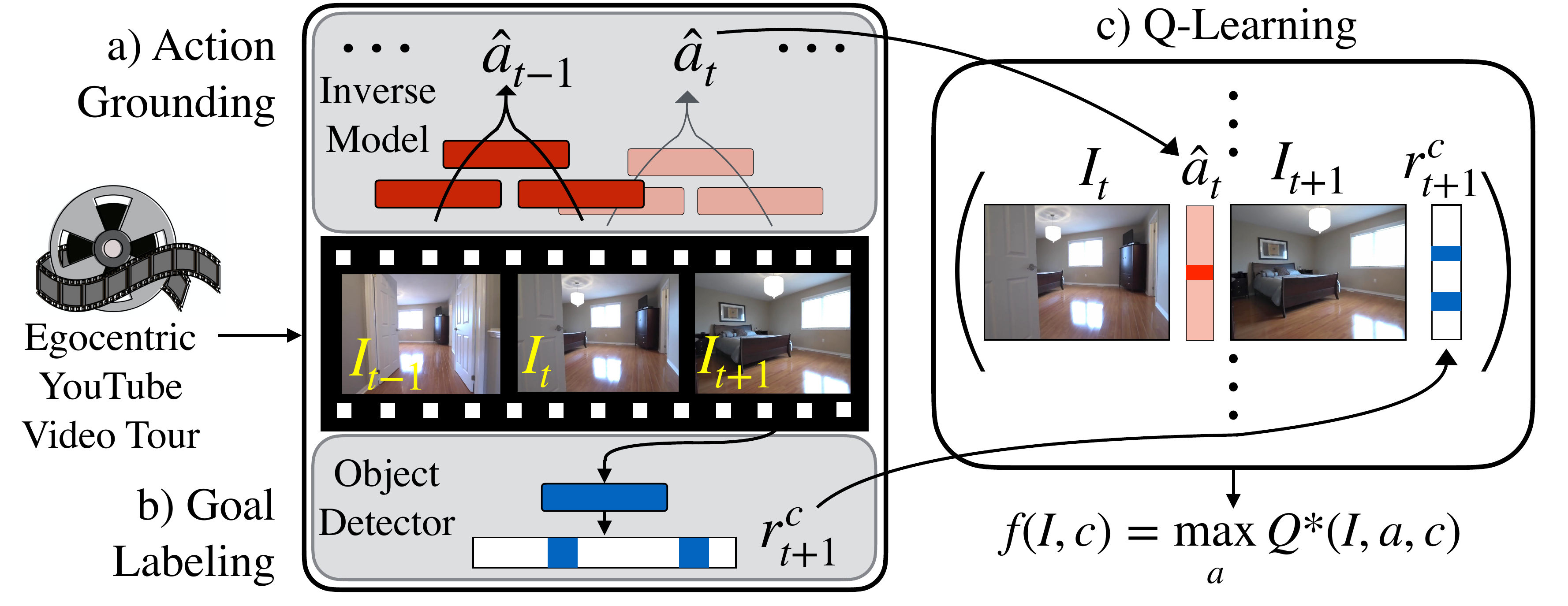}
\caption{\textbf{Learning Values Functions from Videos.} Egocentric videos
tours of indoor spaces are \textbf{a)} grounded in actions (by labeling via an inverse
model), \textbf{b)} labeled with goals (using an object detector). This prepares them for
\textbf{c)} Q-learning, which can extract out optimal Q-functions for reaching goals purely
by watching egocentric videos. See \secref{vfv} for more details.}
\figlabel{vfv}
\end{figure}

Given an image $I$ and a set of object categories \C, we seek to learn a
function $f(I,c)$ that can predict the \textit{value} for image $I$ for
reaching an object of category $c \in \mathcal{C}$. Images that show physical
space close to instances of category $c$ should have a higher value than
images that show regions far away from it.

Let's say we have \V{}, a set of egocentric video tours of indoor
spaces. We seek to learn this function from such videos. We follow a three step
procedure: \textbf{a)} imagining robot actions that convey the robot between intervening
frames, \textbf{b)} labeling video frames of images containing instances of the
desired object category, and \textbf{c)} Q-learning on the resulting reward-labeled
image-action sequence trajectories. \figref{vfv} shows an overview of this
process, we describe it in more detail below.

\textbf{Action grounding.}
Such videos don't come with any information for how one image is related to
another. We follow the pseudo-labeling approach from~\cite{kumar2019learning,
torabi2018behavioral}, to imagine the actions the robotic agent would have
taken to induce the depicted transformation. We collect a small amount of
interaction data, where a robot executes random actions in a handful of
environments. This data is in the form of image action sequences, ${\ldots,
I_t, a_t, I_{t+1}, \ldots}$, and importantly, has information of the action
that was executed to go from $I_t$ to $I_{t+1}$. We use this interaction
dataset to train a \textit{one-step inverse model}
$\psi$~\cite{jordan1992forward, agrawal2016learning} that uses
$I_t$ and $I_{t+1}$ to predict $\hat{a_t} = \psi\left(I_t, I_{t+1}\right)$.
$\psi$ is trained via a cross-entropy loss between its prediction $\hat{a_t}$
and ground truth $a_t$. We use this inverse model $\psi$ to
\textit{pseudo-label} the video dataset \V{} with action labels to obtain
\Vhat{}.

\textbf{Labeling Video Frames with Goals.} Our next step involves labeling
video frames with the presence of object instances from categories in \C.
This can simply be done by using an off-the-shelf object detector \D{} (such as
Mask RCNN~\cite{he2017mask}) trained on the MS-COCO
dataset~\cite{lin2014microsoft}. We assign a binary reward value $r^c(I)$ for
each category $c$ for each video frame $I$: $+1$ if object detected, and $0$ otherwise. 

\textbf{Value Learning via Off-policy Q-Learning.}
Our next step is to derive value function $f(I,c)$ for the different
categories. The above two steps, generate reward-labeled, image-action trajectories
for traversals in indoor environments. For each category $c \in $ \C, these are
in the form of quadruples $\left(I_t, \hat{a_t}, I_{t+1}, r^c_{t+1}\right)$,
where $I_t$ and $I_{t+1}$ are consecutive frames, $\hat{a_t}$ is the
pseudo-label as predicted from the inverse model $\psi$, and $r^c_{t+1}$ is the
label for category $c$ for image $I_{t+1}$. These quadruples can be thought of
as transitions from a Markov Decision Process (MDP)~\cite{sutton1998rli}, where the agent
gets $+1$ reward for entering into a location close to the desired target
object, and $0$ reward otherwise. 

Thus, even though we don't have access to the physical environment, a simple
video traversal of an indoor space can be pseudo-labeled to obtain transition
samples from the underlying MDP operating in this environment. Under mild
conditions, such samples are all that are necessary for learning
\textit{optimal} value functions via Q-learning~\cite{watkins1989learning}.
Thus, instead of directly learning the value function $f(I,c)$, we learn a
Q-function $Q(I,c,a)$ that predicts the $Q$ value of executing action $a$ when
at image $I$ and seeking to find object from category $c$. 

Q-learning takes the following form, where we seek to learn the fixed point,
$Q^*$ of the following Bellman equation (for each category $c$):
$Q^*(I_t, a_t, c) = \max_{a'}{\left(r^c_{t+1} + \gamma Q^*\left(I_{t+1}, a',
c\right)\right)}$.
% \begin{equation}
% Q^*(I_t, a_t, c) = \max_{a'}{\left(r^c_{t+1} + \gamma Q^*\left(I_{t+1}, a', c\right)\right)}.
% \end{equation}
% which in our case becomes:
% \begin{equation}
% Q^*(I_t, a_t, c) = r^c_{t+1} + \gamma\max_{a'}{Q^*\left(I_{t+1}, a', c\right)}.
% \end{equation}
This is done by finding the $Q$ that minimizes the following objective
function, over transition quadruples from \Vhat{} (we
parameterize $Q$ as a convolutional neural network (more details in
\secref{experiments})): 

%, and use gradient based methods to optimize:
\begin{equation}
\sum_{\text{\Vhat{}}}{\left[Q(I_t,a_t,c) - \left(r^c_{t+1} + \gamma\max_{a'}{Q\left(I_{t+1}, a', c\right)}\right) \right]^2}.
\end{equation}
Value function $f(I,c)$ can be obtained by simply taking a maximum of the
Q-values over all actions, \ie, $f(I,c) = \max_aQ(I,a,c)$. This gives us our
desired value function.

% A number of variations of the basic Q-learning formulation have been
% studied in literature~\cite{}. While all of them are applicable here, but we
% found good performance with the Double DQN approach~\cite{ddqn}.

Note, Q-learning can learn optimal Q-functions independent of where
transition quadruples come from (as long as they cover the space), and in
particular, can learn from off-policy data. This allows us to learn
\textit{optimal} value functions even though the video dataset may not follow
optimal paths to any targets. 
% Furthermore, learning can also happen on trajectories that don't even go to the
% desired object at all.
This also leads us to favor Q-learning over the simpler alternative of
employing Monte Carlo or TD(0) \textit{policy evaluation}~\cite{sutton1998rli}.
Policy evaluation is simpler as it does not involve reasoning about intervening
actions, but consequently only learns the value of the underlying policy
depicted in the video, rather than the optimal policy. Our experiments
demonstrate this contrast between these two design choices, in scenarios where
videos don't show the optimal goal reaching behavior. 

% This also precludes the simpler alternative of not reasoning about actions
% and employing \textit{policy evaluation} for learning value functions. Such
% policy evaluation will only predict the value of the underlying policy
% depicted in the video, instead of the optimal values recovered by our
% proposed use of Q-learning. In fact, our experiments demonstrate this
% contrast in scenarios where videos doesn't show the optimal behavior. 

% Lastly, an alternate is to not reason about actions, but simply use value
% learning on the video data labeled with object detectors. Such value learning
% does not recover the \textit{optimal} value, but only the value of the
% underlying policy depicted in the video. We experimentally show that Q-learning
% does recover the optimal value function and value learning fails to, in scenarios
% where the video doesn't necessarily show the optimal behavior. 

The learned Q-function, and the associated value function $f(I,c)$, implicitly
learn semantic cues for navigation. They can learn what images lead to the
desired category, and what don't. Relative magnitude of their prediction can be
used to pick directions for exploration.  It is worth noting, this obtained
value function is the optimal value function under the dynamics of the agent
recording the video. We are implicitly assuming that optimal value function
under the robot's action space or dynamics would be similar enough. This
assumption may not always be true (specially at fine temporal scales), but is
true in a number of situations at coarser time scales. 

\section{Experiments}
\seclabel{experiments}
We show results on the ObjectGoal task in novel
environments~\cite{anderson2018evaluation}. Our experiments test the extent to
which we are able to learn semantic cues for navigation by watching
videos, and how this compares to alternate techniques for learning such cues
via direct interaction. We also compare against alternate ways of learning from passive video
data, and show visualizations of our learned value functions.

\pp{Video Dataset.} We mined for real estate tours from YouTube. This
\textit{\textbf{\YTD}} consists of 1387 videos with a total run length of 119
hours. A sample video is shown in supplementary video. We sample a frame
every 1.5 seconds resulting in 550K transitions tuples $I_t, I_{t+1}$ for
Q-learning (after removing outdoor scenes and people).  We 
denote this dataset as \V{yt}.

% \input{results-slam}
% \input{results-slam-updated}
% In our experiments we wish to show that our approach is able to internalize
% useful visuospatial knowledge more efficiently, and resulting in better
% performance than contemporary methods.
% 
% \noindent \textbf{What all do we want to show?}
% \begin{enumerate}
%     \item Utility of semantic cues for object goal. We compare oracle with one that is acting randomly (but with some structure being built up over time.
%     \item We can learn semantic cues simply by watching videos. Ours \vs random.
%     \item This way of leveraging demonstrations is better than other
%     alternatives. What would be another alternative here: a) behavior closing
%     with an LSTM, b) value prediction from optimal trajectories. 
%     \item Such learning is even better than pure end-to-end RL that uses interactions in the training set for learning.
%     \item Learned value function works for a number of different agents with different underlying action space.
% \end{enumerate}

\textbf{Experimental Setup.} We work with a simulated robot in visually
realistic simulation environments. We use the Habitat simulator~\cite{habitat}
with the Gibson environments~\cite{gibson} (100 training environments from the
medium split, and the 5 validation environments from the tiny split). These
environments are derived from scans of real world environments, and thus retain
the visual and layout complexity of the real world, but at the same time allow
for systematic experimentation.

We split the 105 environments into three sets: \E{train}, \E{test}, and
\E{video} with 15, 5, and 85 environments respectively.  The robot has access
to, and can directly interact with environments in \E{train}.  \E{test} is same
as the official Gibson tiny validation set that comes with human verified
semantic class labels~\cite{armeni20193d}. It is used to setup downstream semantic navigation tasks
for evaluation. \E{train} and \V{yt} are used for learning via our proposed
formulation. Learned policies are evaluated on \E{test}.
For some additional control experiments, we also create a dataset of synthetic
videos \V{syn} using the 85 environments in \E{video} (generation procedure
described in supplementary). 
Our splitting procedure ensures: \textbf{a)} final testing happens in novel,
previously unseen environments, and \textbf{b)} the robot does not have direct
access to environments in which videos were shot (neither the \E{video} used to generate
\V{syn}, nor the real estate shown in \YTD \V{yt}).

\pp{Robot Model.} We use a simplified robot action
space with four actions: move forward by $25cm$, rotate left $30^\circ$, rotate
right $30^\circ$ and stop. We assume perfect localization, that is, the robot
exactly knows where it is relative to its previous location. This can be achieved by running a SLAM system, or using additional
sensors such as an IMU. The robot is a $1.25m$ long cylinder of radius $10cm$, and has a RGB-D camera with
$90^\circ$ field of view, mounted at a height of $1.25m$.

\pp{Semantic Visual Navigation Task.} We set up the ObjectGoal
task~\cite{anderson2018evaluation} in \E{test} for testing different models.
Note that \E{test} is same as the Gibson tiny validation set (and does not
overlap with environments in \E{train} or \E{video}), and comes with
human-verified annotations for semantic classes. We use these semantic
annotations to set up the ObjectGoal task for 5 categories: bed, chair, couch,
dining table, and toilet. We sample 1075 test episodes, equally
distributed among these 5 classes. For each episode, the agent is
initialized at the starting location, and asked to go to the chosen object
category. An episode is considered successfully solved if the agent reaches
within $1m$ of \textit{any} instance of the target category. We report both the
success rate and SPL~\cite{anderson2018evaluation}. Minimum geodesic distance
to \textit{any} instance of the target category, is used as the reference path
length for computing SPL. 
We consider two settings: \textit{Oracle Stop} (episode is automatically
terminated and deemed successful when the agent is within $1m$ of the target
category), and \textit{Policy Stop} (agent needs to indicate that it has reached
the goal). We report results along with a 90\% bootstrap confidence interval.

% We group these episodes into 3 difficulty levels,
% based on distance to the nearest instance of the target category: \textit{easy}
% ($\leq 2m$), \textit{medium} ($2m$ to $5m$), and \textit{hard} ($5m$ to $15m$). 

% \underline{\textbf{Implementation Details.}} \\ 
\subsection{Implementation Details}
% We describe implementation details for our model.
\pp{Action Grounding.} Inverse model $\psi$ processes RGB images $I_t$
and $I_{t+1}$ using a ResNet-18 model~\cite{resnet}, stacks the resulting
convolutional feature maps, and further processes using 2 convolutional layers,
and 2 fully connected layers to obtain the final prediction for the intervening
action. We train $\psi$ on 40K interaction frames gathered by
randomly executing actions in \E{train}. 
This is an easy learning task, we obtain close to $96\%$ classification accuracy on a held-out validation set.
We use this inverse model to pseudo-label video dataset \V{yt} and \V{syn} to
obtain \Vhat{yt} and \Vhat{syn}.

\pp{Object Detectors.} We use Mask RCNN~\cite{he2017mask} trained on MS-COCO
dataset~\cite{lin2014microsoft} as our detector \D{coco}. Frames with
detections with score in the top $10\%$ are labeled as +1 reward frames.
\D{coco} also predicts a foreground mask for each detection. We use it to
evaluate a stopping criterion at test time.

\pp{Q-Learning.} We represent our Q-function with ResNet 18 models, followed by
1 convolutional layer, and 2 fully connected layers with ReLU non-linearities.
We use Double DQN (to prevent chronic over-estimation~\cite{ddqn}) with
Adam~\cite{adam} for training the Q-networks, and set $\gamma=0.99$. As our
reward is bounded between $0$ and $1$, clipping target value between $0$
and $1$ led to more stable training.

\pp{Semantic Navigation Policy.}
\underline{High-level policy} stores 12 images for each node in the topological
graph (obtained by rotating 12 times by $30^\circ$ each). It uses the learned
value function, $f(I,c)$, to score these $12n$ images (for a $n$ node
topological graph), and samples the most promising direction for seeking
objects of category $c$.  The sampled direction is converted into a short-term goal
by sampling a location at an offset of $1.5m$ from the chosen node's location,
in the chosen view's direction. \underline{Low-level
policy}~\cite{map-plan-baseline} uses occupancy maps (built using depth
images)~\cite{elfes1989using} with fast marching planning~\cite{fmm} to execute
robot actions to reach the short-term goal. It returns control on success /
failure / timeout. The High-level policy also factors in the distance to the
sampled direction, and score from \D{coco} while sampling directions.
\underline{Stopping criterion:} The agent chooses to stop if \D{coco} fires with
confidence $\geq \tau_c$ and median depth value in the predicted mask is $\leq d_c$
distance.  More details are provided in Supplementary 
\secref{nav-policy}.

\subsection{Results}
% \underline{\pp{Results}} \\
\seclabel{results}
\tableref{results-eval} reports performance on the ObjectGoal task for our method and
compares it to other methods for solving this task. An important aspect to
consider is the amount and type of supervision being used by different methods.
We explicitly note the scale (number of frames, environments) and type (reward
signals) of active interaction used by the different methods.  For
\textit{Policy Stop} setting, for all methods, we found our stopping criterion
to work much better than using the method's own stop signal. We use it for
all methods.
Using only 40K reward-less interaction samples from \E{train}, along with
in-the-wild YouTube videos our proposed method is able to achieve an OS-SPL
(Oracle Stop SPL) of 0.53 and PS-SPL (Policy Stop SPL) of 0.22 respectively in
the Oracle and Policy stop settings.  We put this in context of results from
other methods. 
 
\pp{Topological Exploration} exhaustively explores the environment. It uses our
hierarchical policy but replaces $f(I,c)$ with a random function, and ignores
scores from \D{coco} to score different directions. As the topological map
grows, this baselines systematically and exhaustively explores the environment.
Thus, this is quite a bit stronger than executing random actions (OS-SPL of $0.15$). 
It is able to find objects often (67\%), though is inefficient
with OS-SPL of 0.30.

\pp{Detection Seeker} also does topological exploration, but additionally also
uses scores from \D{coco} to seek the object once it has been detected.  This
performs quite a bit better at 0.46 SPL. This indicates that object detectors
provide a non-trivial signal for object goal navigation. Even lower confidence
detection scores for more distant but partially visible objects will guide the
agent in the right direction.  Our method captures more out of view context,
and consequently does better across all settings.

% We replace our learned value function with a trivial value function that
% outputs random values (this amounts to exploring randomly to find the
% object).  Everything else in the hierarchical navigation policy is kept the
% same. Our learned value functions perform much better across the board,
% improving SPL by 58\% relative in the oracle stop case and nearly doubling
% the SPL when the policy has to stop by itself (we stop based on our predicted
% values for both models). This demonstrates that we are able to learn
% meaningful semantic cues that improve the efficiency of goal reaching in
% novel environments.

\renewcommand{\arraystretch}{1.1}
\begin{table}
\setlength{\tabcolsep}{3pt}
\centering
\caption{\textbf{Results}: Performance for ObjectGoal in novel environments
\E{test}.  Details in \secref{results}.}
% in oracle stop and policy stop settings.
\tablelabel{results-eval}
\resizebox{\textwidth}{!}{
\begin{tabular}{lccccccc}
\toprule
\multirow{2}{*}{\textbf{Method}}
& \multicolumn{3}{c}{\textbf{Training Supervision}} & \multicolumn{2}{c}{\textbf{Oracle Stop}} & \multicolumn{2}{c}{\textbf{Policy Stop (using \D{coco})}} \\
\cmidrule(lr){2-4} \cmidrule(lr){5-6} \cmidrule(lr){7-8}
& \textbf{\# Active Frames} & \textbf{Reward} & \textbf{Other} & \textbf{SPL} & \textbf{Success (SR)} & \textbf{SPL} & \textbf{Success (SR)} \\
% & Frames & & & & & & \\
\midrule
Topological Exploration           & -                                                        & -               & -               & 0.30 $\pm$ 0.02          & 0.67 $\pm$ 0.02          & 0.13 $\pm$ 0.01          & 0.29 $\pm$ 0.02 \\
Detection Seeker                  & -                                                        & -               & -               & 0.46 $\pm$ 0.02          & 0.75 $\pm$ 0.02          & 0.19 $\pm$ 0.02          & 0.37 $\pm$ 0.02 \\
RL (RGB-D ResNet+3CNN)            & 100K (\E{train})                                         & Sparse          & -               & 0.17 $\pm$ 0.01          & 0.37 $\pm$ 0.02          &                          & \\
RL (RGB-D ResNet+3CNN)            & 10M (\E{train} $\cup$ \E{video})                         & Dense           & -               & 0.26 $\pm$ 0.02          & 0.54 $\pm$ 0.02          &                          & \\
RL (RGB-D 3CNN)                   & 38M (\E{train} $\cup$ \E{video})                         & Dense           & -               & 0.28 $\pm$ 0.02          & 0.57 $\pm$ 0.03          &                          & \\
RL (RGB ResNet)                   & 20M (\E{train})                                          & Dense           & -               & 0.29 $\pm$ 0.02          & 0.56 $\pm$ 0.03          & 0.08 $\pm$ 0.01          & 0.21 $\pm$ 0.02 \\
RL (Depth 3CNN)                   & 38M (\E{train})                                          & Dense           & -               & 0.25 $\pm$ 0.02          & 0.52 $\pm$ 0.02          &                          & \\
Behavior Cloning                  & 40K (\E{train})                                          & -               & \Vhat{yt}       & 0.25 $\pm$ 0.02          & 0.53 $\pm$ 0.03          & 0.08 $\pm$ 0.01          & 0.20 $\pm$ 0.02\\
Behavior Cloning + RL             & 12M (\E{train})                                          & Dense               & \Vhat{yt}       & 0.24 $\pm$ 0.02          & 0.58 $\pm$ 0.02          &                          & \\
Our (Value Learning from Videos)  & 40K (\E{train})                                          & -               & \Vhat{yt}       & \textbf{0.53} $\pm$ 0.02 & \textbf{0.79} $\pm$ 0.02 & \textbf{0.22} $\pm$ 0.02 & \textbf{0.39} $\pm$ 0.03\\ \midrule
Behavior Cloning                  & 40K (\E{train})                                          & -               & \Vhat{syn}      & 0.36 $\pm$ 0.02          & 0.71 $\pm$ 0.02          & 0.10 $\pm$ 0.01          & 0.26 $\pm$ 0.02\\
Behavior Cloning + RL             & 12M (\E{train})                                          & Dense               & \Vhat{syn}      & 0.24 $\pm$ 0.02          & 0.55 $\pm$ 0.03          &                          & \\
Our (Value Learning from Videos)  & 40K (\E{train})                                          & -               & \Vhat{syn}      & 0.48 $\pm$ 0.02          & 0.75 $\pm$ 0.02          & 0.21 $\pm$ 0.02          & 0.38 $\pm$ 0.03\\ \midrule
  Strong Supervision Values       & \multicolumn{3}{c}{Labeled Maps (\E{video})}             & 0.55 $\pm$ 0.02 & 0.81 $\pm$ 0.02 & 0.24 $\pm$ 0.02          & 0.43 $\pm$ 0.02  \\
  Strong Supervision + VLV (Ours) & \multicolumn{3}{c}{Labeled Maps (\E{video}) + \Vhat{yt}} & 0.57 $\pm$ 0.02 & 0.82 $\pm$ 0.02 & 0.23 $\pm$ 0.02          & 0.41 $\pm$ 0.02 \\
\bottomrule
\vspace{-15pt}
\end{tabular}}
\end{table}

%RL (RGB-D)              & 38M (\E{train})                  & Dense  & -          & 0.25 $\pm$ 0.02 & 0.53 $\pm$ 0.02 &                 & \\
%RL (RGB-D)              & 38M (\E{train})                  & Sparse & -          & 0.15 $\pm$ 0.01 & 0.32 $\pm$ 0.02 &                 & \\
% 1. RL(RGB-D ResNet + SimpleCNN, 100K, sparse, 15), SG trains.
% 2. RL(RGB-D SimpleCNN + SimpleCNN, 38M, 100) Finished training
% 3. RL(Depth SimpleCNN, 38M, 15), Already in paper : in
% 4. RL(RGB-D, ResNet + SimpleCNN, 10M, dense, 100), Finish training in 2 hours : in
% 5. RL(RGB ResNet, 20M, dense, 15), Already trained. :in

% rl rgbd sparse
% & 0.28 $\pm$ 0.04 & 0.55 $\pm$ 0.05& 0.16 $\pm$ 0.02 & 0.36 $\pm$ 0.04& 0.06 $\pm$ 0.01 & 0.16 $\pm$ 0.03& 0.00 $\pm$ 0.01 & 0.01 $\pm$ 0.02& 0.15 $\pm$ 0.01 & 0.32 $\pm$ 0.02

\pp{End-to-end RL.} We also compare against many variants of end-to-end RL
policies trained via direct interaction. We use the
PPO~\cite{schulman2017proximal} implementation for CNN+GRU policies that are
implemented in Habitat~\cite{habitat}. We modify them to work with ObjectGoal
tasks (feeding in one-hot vector for target class, modifying rewards), and most
importantly adapt them to use ImageNet initialized ResNet-18
models~\cite{resnet} for RGB (given no standard initialization for Depth image,
it is still processed using the original 3-layer CNN in Habitat code-base). The
fairest comparison is to train using sparse rewards (dense rewards will require
environment instrumentation not needed for our method) in \E{train} for 40K
interaction samples with RGB-D sensors. This unsurprisingly did not work
(OS-SPL: 0.17 and OS-SR: 37\%). Thus, we aided this baseline by providing it
combinations of more environments (\E{train} $\cup$ \E{video}), many times more
samples, and dense rewards. Even in these more favorable settings, end-to-end
RL didn't perform well. The best model had a OS-SPL of 0.29 and OS-SR of 56\%
(\vs 0.50 and 75\% for our method), even when given interaction access to
$6\times$ more environments, 250$\times$ more interaction, and dense rewards
(\vs no rewards). This demonstrates the power of our proposed formulation that
leverages YouTube videos for learning about spatial layout of environments.
Policy stop evaluation is computationally expensive so, we report the score only
for the strongest model.
% 1. RL(RGB-D ResNet + SimpleCNN, 100K, sparse, 15), SG trains.
% 2. RL(RGB-D SimpleCNN + SimpleCNN, 38M, 100) Finished training
% 3. RL(Depth SimpleCNN, 38M, 15), Already in paper
% 4. RL(RGB-D, ResNet + SimpleCNN, 10M, dense, 100), Finish training in 2 hours
% 5. RL(RGB ResNet, 20M, dense, 15), Already trained.
% in \E{train} environments. Note that
% those are the only environments that are available to the agent for training.
% Generating dense rewards in physical environments can be hard, thus we only use
% sparse rewards (reward of +10 if agent arrives at a location within $1m$ of the
% target object. We use the default CNN policy from the Habitat
% codebase~\cite{habitat}, and consider RGB-D and Depth inputs. We train each of
% these polices with $5M$ interaction frames. Even with use of $25\times$ more
% active interactions, these end-to-end RL policies perform markedly worse than
% our policies (SPL of 0.25 \vs 0.36 for our). This is remarkable as our method
% derives this semantic signal largely from passive videos, rather than active 
% interaction.
% \sg{Part of the gain comes from use of
% hierarchies, and path planning on occupancy maps (as also observed by Chaplot
% \etal~\cite{chaplot2020learning} for more geometric tasks). Our policy
% design allows us to leverage this gain for semantic tasks; and we learn these
% semantic cues purely from passive data.} 

% Training such RL policies with the \textit{stop} action is hard (as the agent
% quickly learns to stop to avoid time penalty), and resulted in very low
% performance, and hence we omit it from the results table.

\pp{Behavior Cloning (BC) on Pseudo-Labeled Videos \Vhat{}.} We pre-process the
videos to find trajectories that lead to objects of interest (as determined by
\D{coco}). We train CNN+GRU models to predict the pseudo-labeled action labels
on these trajectories. As this is passive data that has already been collected,
we are limited to using behavior cloning wth \rgb input as opposed to richer
inputs or the more sophisticated DAgger~\cite{ross2011reduction}. This is effectively
the BCO(0)~\cite{torabi2018behavioral} algorithm. This performs fairly similarly to RL methods and with negligible
sample complexity, though still lags far behind our proposed method that
utilizes the exact same supervision. Perhaps this is because our proposed
method uses pseudo-labeled action indirectly and is more tolerant to mismatch
in action space. In contrast, behavior cloning is critically reliant on action
space similarity. This is brought out when we use \Vhat{syn} instead of \Vhat{yt}
where the action space is more closely matched. Behavior cloning performs much
better at 0.36 OS-SPL, though our method still performs better than all the
baselines even when trained on videos in \Vhat{syn}.

\pp{Behavior Cloning+RL.} 
We also experimented with combining behavior cloning and RL. We use the
behavior cloning policies obtained above, and finetune them with RL. For the same
reasons as above, this policy is limited to use of \rgb inputs. When finetuning
from behavior cloning policy trained on \Vhat{yt} we found performance to
remain about the same (OS-SPL 0.24). When starting off from a policy trained on
\Vhat{syn}, we found the performance to drop to OS-SPL of 0.24. We believe that
the dense reward shaped RL may be learning a qualitatively different policy
than one obtained from behavior cloning. Furthermore, use of dense
rewards for RL, may limit the benefit of a good initialization. 

\pp{Strong Supervision Value Function.} While our focus is on learning purely
from passive data, our semantic navigation policy can also be trained using
strong supervision obtained using semantically labeled maps. We train $f(I,c)$
to predict `ground-truth' Q-values computed using the number of steps to the
nearest instance of category $c$ on the meshes from environments in \E{video}.
This model is strong at OS-SPL of 0.55. This serves as a very competitive ObjectNav 
policy in the regime where we allow such strong supervision. 
Our proposed method that uses significantly less supervision
(in-the-wild videos from YouTube \vs environment scans) is still close to the performance of this strongly
supervised method (OS-SPL 0.53). When we combine the two by training the strongly supervised objective jointly with
our Q-learning based objective, performance is even stronger at OS-SPL of 0.57 (significant at a p-value of 0.025).

Thus, in conclusion, value functions learned via our approach from YouTube
video tours of indoor spaces are effective and efficient for semantic
navigation to objects of interest in novel environments. They compare favorably
to competing reinforcement learning based methods, behavior cloning approaches,
and strong exploration baselines, across all metrics.

% \noindent \textbf{Pseudo-labeling performance.}
% \noindent 
% \textbf{Experiments}
% \begin{enumerate}
%   \item Effectiveness of pseudo-labeling.
%   \item Visualization of learned value function maps (with and without ground
%   truth actions).
%   \item Panorama visualizations.
%   \item Solving down-stream tasks with learned value functions, comparing to
%   random as a baseline, and a oracle trained with the ground truth actions.
%   \item Reading off other things from the learned model: a) t-SNE embedding
%   of learned representations inside the Q-function, see if we can extract out
%   a semantic graph from it, b) simple readout functions trained on top of our
%   features (and if they can predict the next observation better than ImageNet
%   features or something).
% \end{enumerate}

\subsection{Ablations}
\seclabel{ablations}
We present ablations when testing policies on \E{train} in
Oracle Stop setting. Note \E{train} was only used to train the inverse model,
and not the Q-learning models that we seek to compare. The \textit{base setting}
from which we ablate corresponds to training $f(I,c)$ on \Vhat{syn} with
pseudo-labeled actions, \D{coco} based reward labels, and the use of $f(I,c)$ and
spatial consistency for sampling short-term goals. This achieves an OS-SPL of
$0.40\pm0.02$. We summarize results below, table in supplementary.

We notice only a minor impact in performance when \textbf{a)} using true actions as
opposed to actions from inverse model $\psi$ ($0.41\pm0.03$), \textbf{b)} using
true detections as opposed to detections from \D{coco} ($0.40\pm0.03$), \textbf{c)}
using true reward locations as opposed to frames from which object is visible
as per \D{coco} ($0.41\pm0.03$) (the proposed scheme treats frames with high-scoring
detections as reward frames as opposed to true object locations), and \textbf{d)}
using optimal trajectories as opposed to noisy trajectories ($0.43\pm0.03$).
Albeit on simulated data, this analysis suggests that there is only a minor
degradation in performance when using inferred estimates in place of ground
truth values.

Perhaps, a more interesting observation is that there is a solid improvement
when we additionally use \D{coco} score to sample short-term goals
($0.46\pm0.03$). 
We believe use of \D{coco} produces a more peak-y
directional signal when the object is in direct sight, where as differences in
$f(I,c)$ are more useful at long-range. 
Secondly, we found that use of
$360^\circ$ images at training time also leads to a strong improvement
($0.47\pm0.02$). We believe use of $360^\circ$ images at training time
prevents \textit{perceptual aliasing} during Q-learning. In the base setting,
Q-values can erroneously propagate via an image that looks directly at a
wall. Presence of $360^\circ$ context prevents this.  While this is useful for
future research, we stick with the base setting as we are limited by what
videos we could find on YouTube.

\begin{wrapfigure}[11]{R}{0.25\textwidth}
\vspace{-11pt}
\includegraphics[trim=164pt 532pt 877pt 40pt, clip, width=1.0\linewidth, frame=0.5pt]{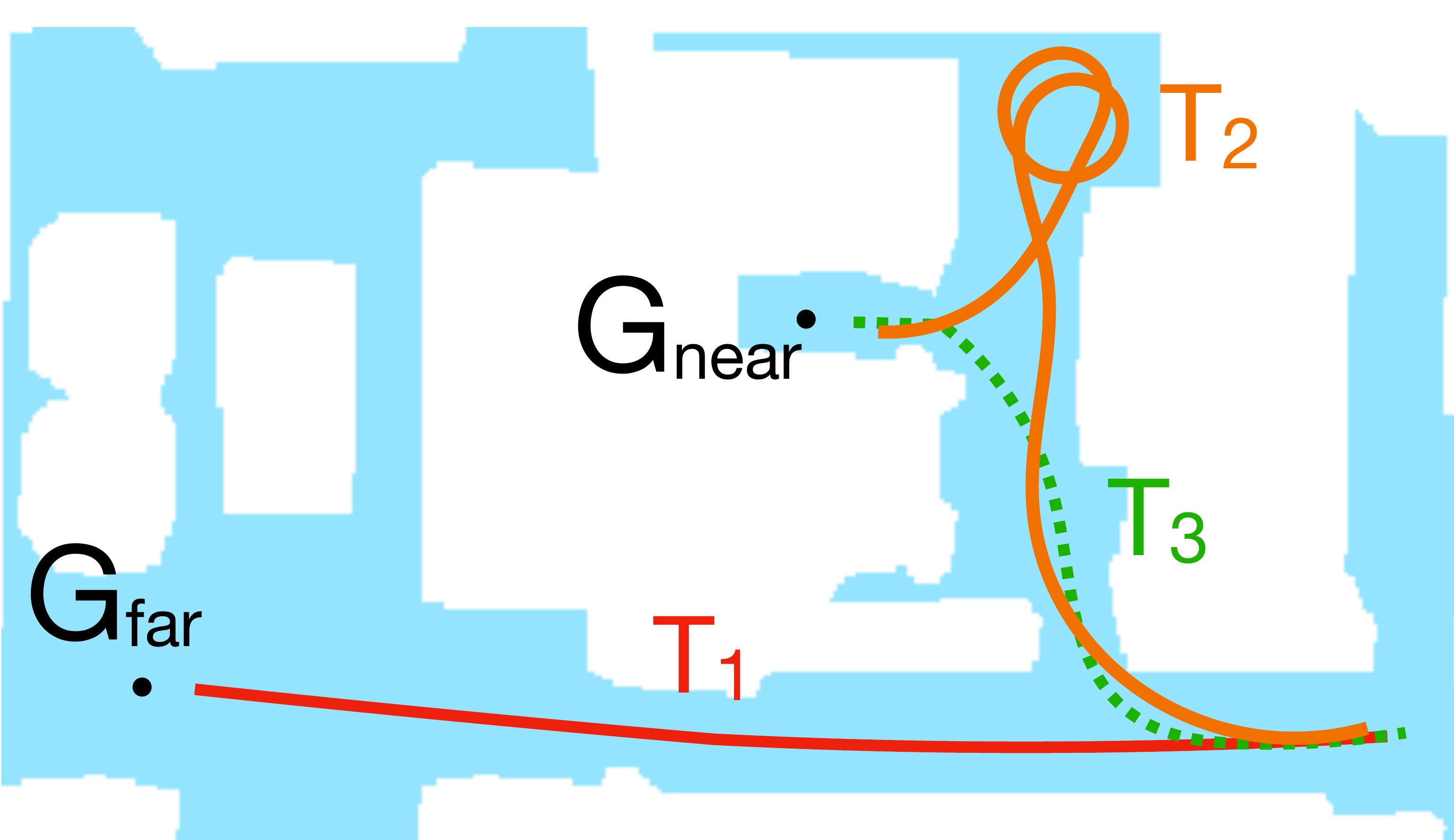}
\includegraphics[trim=164pt 532pt 877pt 40pt, clip, width=1.0\linewidth, frame=0.5pt]{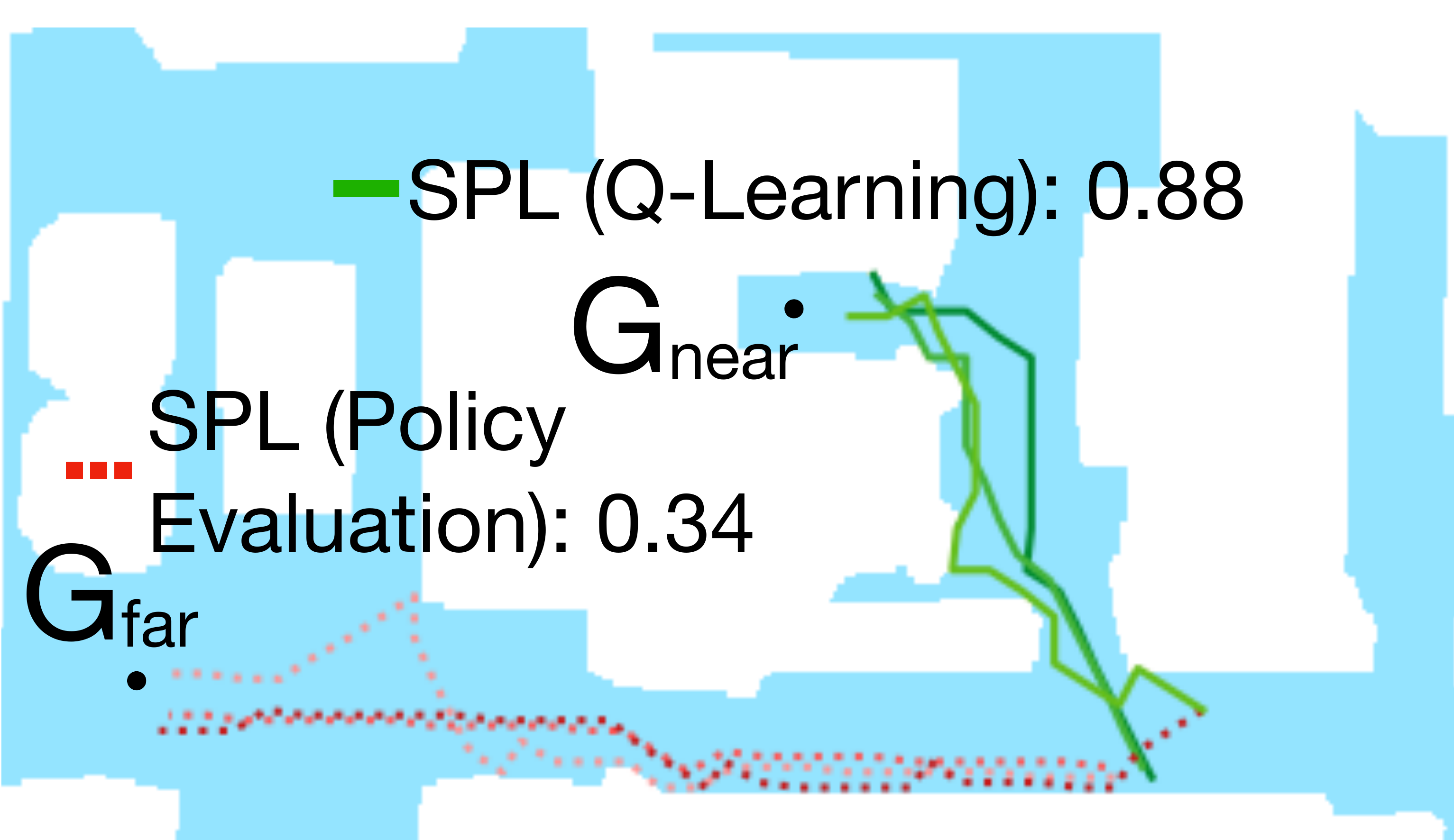}
% \insertWL{1.00}{figures/branching_setup.pdf}
% \insertWL{1.00}{figures/branching/output.pdf}
% \caption{\sg{Shorter figures than this placeholder.}}
% \figlabel{branching}
\end{wrapfigure}
% \end{figure}

\pp{Is action pseudo-labeling necessary?}
As discussed in \secref{vfv}, we favored use of Q-learning over action agnostic
methods, such as policy evaluation, as this allows us to learn optimal value
functions as opposed to value of the policy depicted in the video. To test
this, we train different methods in the \textit{branching environment} as shown
in the figure on the right (top). Desired goal locations are labeled by \G{near} and
\G{far}.  We investigate the learned behavior at the branch point $B$, by
initializing the agent at random locations in the circle $S$. Desired behavior
is for agent to reach \G{near}. In departure from all other
experiments, here we train and test in the same branching
environment. This is a deliberate choice as we seek to understand how different
methods interpret the training data.

Videos in this branching environment are a $50-49.5-0.5\%$ mix of trajectories
$T_1$, $T_2$, and $T_3$. $T_1$ and $T_2$ are sub-optimal trajectories to reach
\G{near} and \G{far} respectively, while $T_3$ is the optimal trajectory to
reach \G{near}. The policy evaluation method doesn't use any action labels, and
correctly infers the values for the policy from which videos are sampled. As
expected, this causes it to pursue the sub-optimal goal
(red paths in bottom figure). In contrast, Q-learning with pseudo-labeled
actions, estimates the \textit{optimal} value function, and consistently
reaches \G{near} (green paths).

% Furthermore, we found that an arbitrary partition of the action space (tuples
% belonging to each action class being arbitrarily assigned into finer classes),
% also resulted in learning of the same optimal behavior. This may indicate that
% we may be able to learn meaningful value functions even when the pseudo action
% label class doesn't exactly match the actual action space used in videos.
% Further empirical and theoretical investigation of this is future work.

% \input{visualizations}
\section{Discussion}
\seclabel{discussion}
We presented a technique to enable learning of semantic cues for finding
objects in novel environments from in-the-wild YouTube videos. Our proposed
technique employs Q-learning on pseudo-labeled transition quadruples.  This
allows learning of effective semantic cues even in the absence of action
grounding and goal-directed optimal behavior. When coupled with a hierarchical
navigation policy, these cues convey the agent to desired objects more
effectively than competitive exploration baselines and RL methods at a fraction
of interaction cost. In the future, we will test our policies on real robots
and extend to other navigation tasks.
%Based on the ablation results we do not significantly degrade performance by using an inverse model
%to pseudo-label actions, or by labeling goal frames with a detector. Comparing the strong supervision result
%to ours \V{syn} (0.55 vs 0.48), Q-learning causes the greatest loss in performance. However, Q-learning is the
%strongest of the tested alternatives (see \secref{ablations}).
% 
% 
% We
% collected a new dataset to 
% We focused on learning and leveraging semantic cues for the task
% of finding objects of interest in novel environments, from YouTube videos.
% Learned semantic cues when used with a hierarchical policy leads to better
% performance at a fraction of interaction cost. 
% 
% 
% by simply watching
% YouTube videos. 
% Our proposed approach effectively leverages such data and reduces
% the number of interaction samples required to learn such semantic cues, and
% compares favorably to competing end-to-end RL approaches. Future work should
% focus on relaxing some of our assumptions, testing with real robots and
% in-the-wild vid
% We collected the \YTD, and showed that
% 
% 
% . We proposed a way to
% learn these from passive data in the form of egocentric tours of indoor
% environments. Our proposed approach effectively leverages such data and reduces
% the number of interaction samples required to learn such semantic cues, and
% compares favorably to competing end-to-end RL approaches. Future work should
% focus on relaxing some of our assumptions, testing with real robots and
% in-the-wild videos from the web. 

\begin{figure}
\centering
\includegraphics[height=0.3\linewidth]{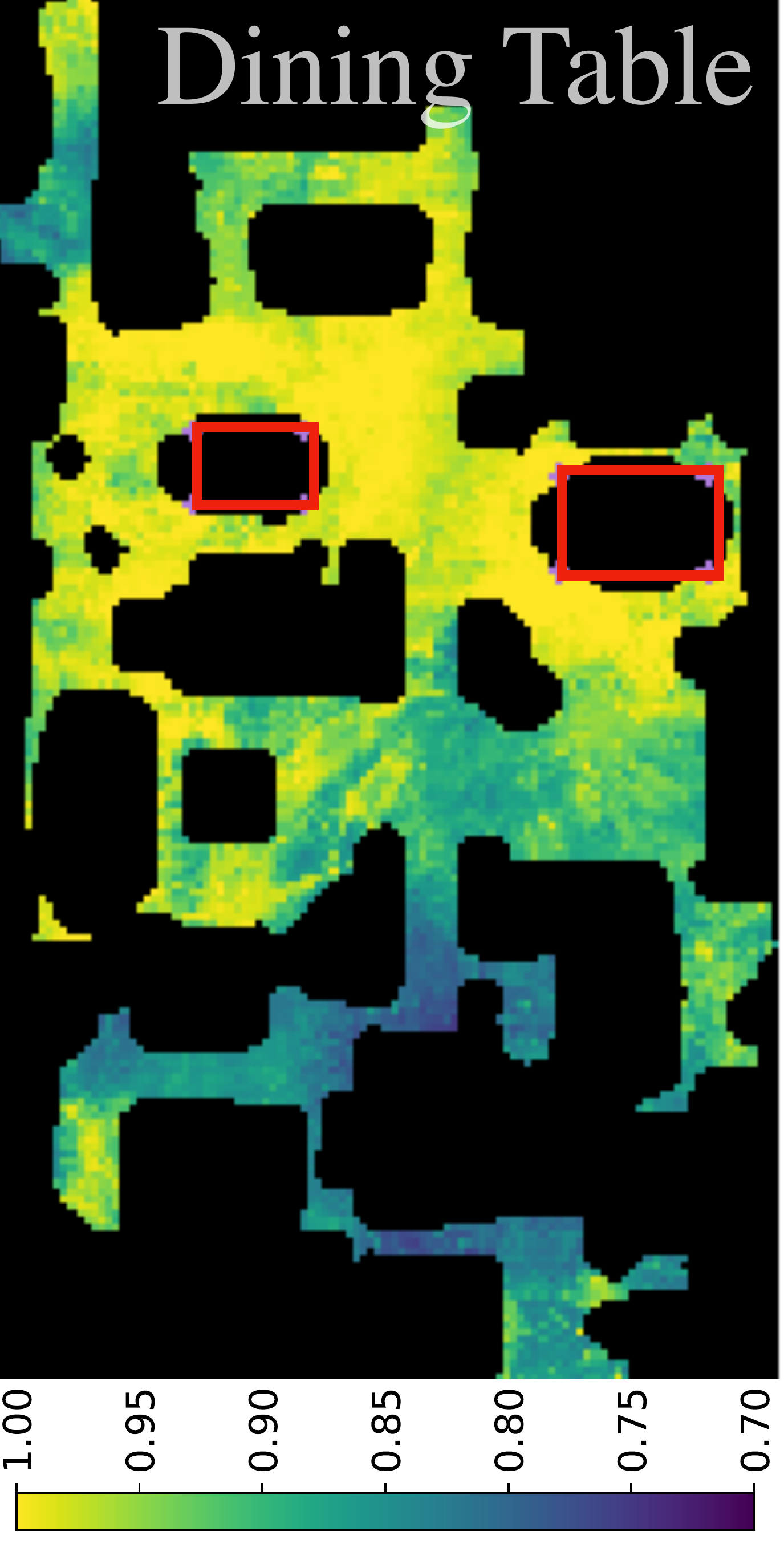}
\unskip\ \vrule\ 
\insertH{0.3}{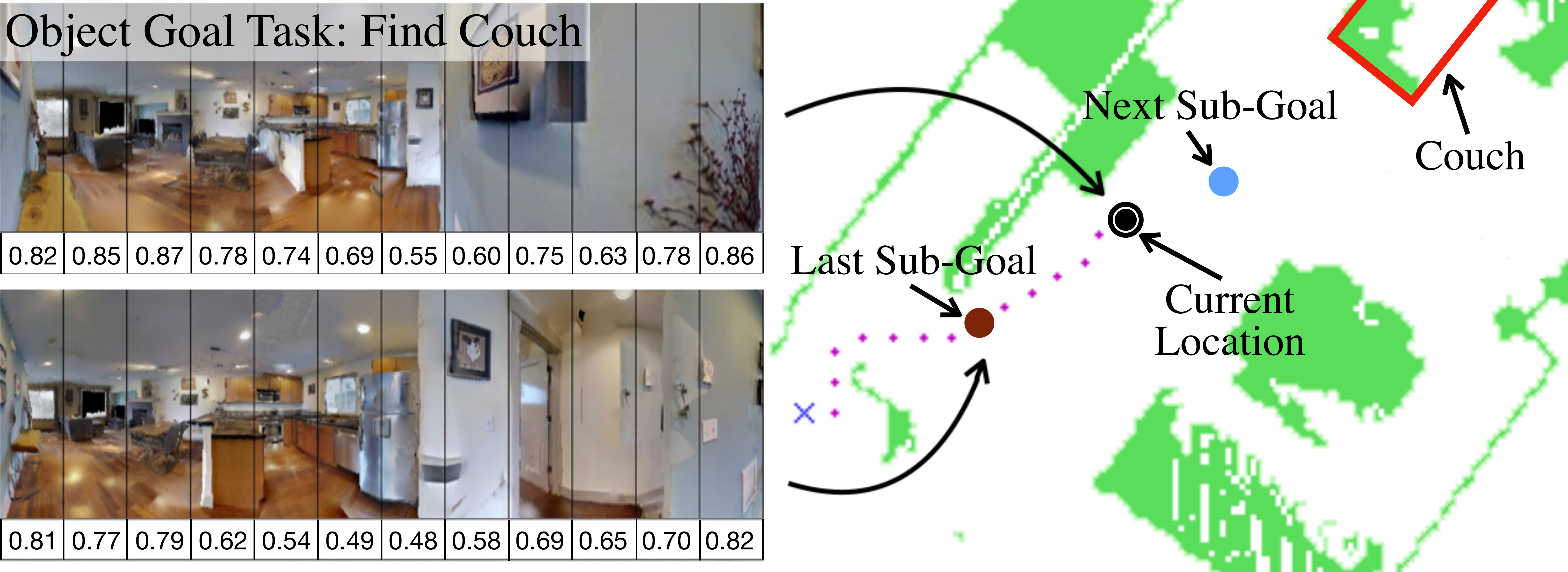} %vis_map.png}
\caption{Left figure shows predicted values for reaching a \textit{dining
table} at different points on the top-view map in a novel environment. Values
are high near the dining tables (denoted by the red boxes), and smoothly
bleed out to farther away regions.  Right shows a sample execution of our
navigation policy finding a \textit{couch} in a novel environment. More in
Supplementary.}
% \insertH[0.3]{figures/syn_value_map.png}
% \caption{Example trajectory from the full evaluation. Radially uniform images are scored at each short-term goal to determine which direction to pursue for the next short-term goal. Previous sub-goals are shown in dark blue with the next sub-goal in light blue. The goal object (couch) is outlined in red.}
\figlabel{trajectory}
\end{figure}

\section*{Broader Impact}
Our specific research in this paper lowers barriers for the training of navigation policies. 
Instead of needing fully instrumented environments, or
  large-scale 3D scans, we can now train using video tours of indoor spaces.
  This significantly expands the environments that such methods can be trained
  on. Existing datasets~\cite{gibson,chang2017matterport3d} used for training
  current systems have a bias towards expensive houses. This is because sensors
  and services involved in constructing such scans are expensive. While our
  current YouTube Walks dataset also has some of this bias, a video tour can be
  collected merely by using a phone with a camera. This will allow training of
  navigation policies that will work well in more typical environments, and
  will democratize the use of learning-based policies for navigation. We also
  acknowledge that the use of publicly available data from the Internet (in our
  case YouTube videos) raises questions about privacy and consent. These issues
  require a broader discussion.
% What form of consent from the uploaders is required for this use case, which is likely outside 
% the post's original scope? Additionally, while broader data collection may remove bias 
% present in other curated datasets, it may add bias of its own.

Our broader research aims to improve policies for navigation in unstructured
environments. This by itself has numerous desirable applications (such as
automated delivery, search and monitoring in hazardous environments, automated
crop inspection and mechanical weeding via under-canopy robots). Such
applications can save lives, prevent food shortage (by preventing herbicide
resistance), and enable development of other automation technologies. 
% prevent food-shortages (one way to tackle
% herbicide resistance~\cite{gressel2017well} is mechanical weeding, but manual
% mechanical weeding is too labor-intensive to be practical), and save lives.

While there are a number of critical applications that our research can
potentially enable, we acknowledge that our research falls under automation,
and as with all other research in this area, in the future it could replace jobs
currently performed by humans.  However, this must be viewed in context of the
critical applications described above.  Resolving or even fully understanding
this trade-off will need a much broader discussion.

% Authors are required to include a statement of the broader impact of their
% work, including its ethical aspects and future societal consequences.  Authors
% should discuss both positive and negative outcomes, if any. For instance,
% authors should discuss a) who may benefit from this research, b) who may be put
% at disadvantage from this research, c) what are the consequences of failure of
% the system, and d) whether the task/method leverages biases in the data. If
% authors believe this is not applicable to them, authors can simply state this.
% 
% Use unnumbered first level headings for this section, which should go at the
% end of the paper. {\bf Note that this section does not count towards the eight
% pages of content that are allowed.}

\textbf{Acknowledgement:} We thank Sanjeev Venkatesan for help with
data collection. We also thank Rishabh Goyal, Ashish Kumar, and Tanmay Gupta
for feedback on the paper. This material is based upon work supported by 
NSF under Grant No. IIS-2007035, and DARPA Machine Common Sense.

\textbf{Gibson dataset license:}
\url{http://svl.stanford.edu/gibson2/assets/GDS_agreement.pdf}

{
\small
\bibliographystyle{plain}
\bibliography{refs}
}

\clearpage

\setcounter{section}{0}
\setcounter{figure}{0}
\setcounter{table}{0}
\setcounter{equation}{0}
\renewcommand{\thesection}{S\arabic{section}}
\renewcommand{\thefigure}{S\arabic{figure}}
\renewcommand{\thetable}{S\arabic{table}}
\renewcommand{\theequation}{S\arabic{equation}}

\section{Hierarchical Policies for Semantic Navigation}
\seclabel{nav-policy}
% \begin{wrapfigure}[20]{R}{0.55\textwidth}
% \vspace{-5pt}
\begin{figure}[h]
\centering 
\insertW{1.0}{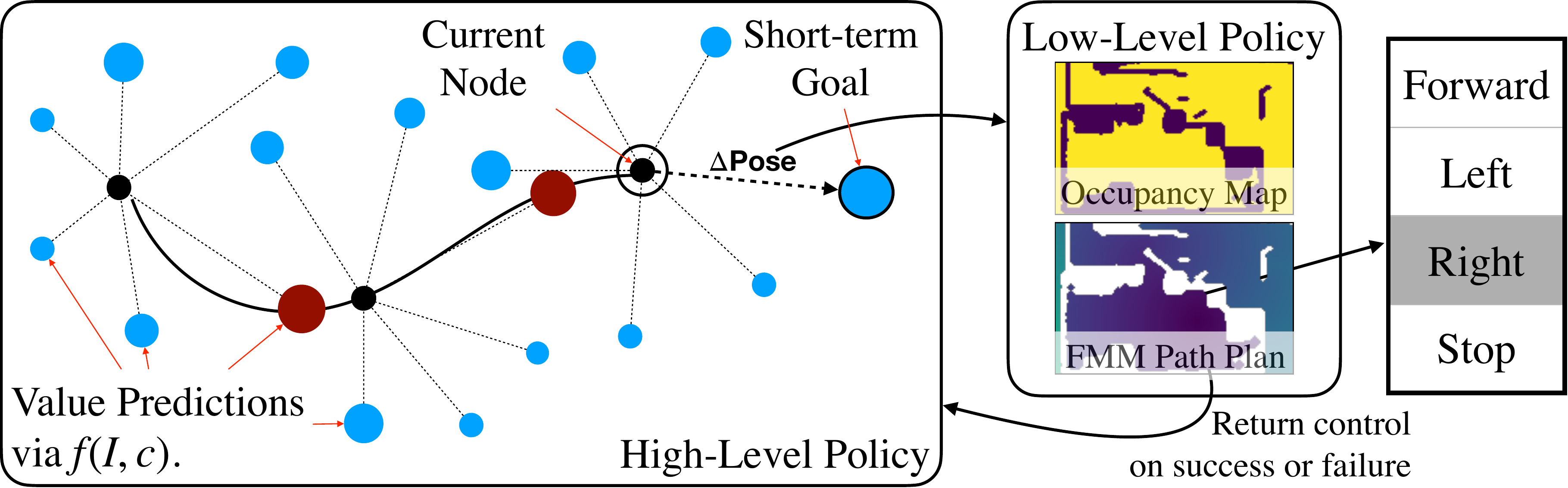}
% \insertWL{1.0}{figures/nav_policy_vertical.pdf}
\caption{\textbf{Hierarchical navigation policy.} High-level policy does
semantic reasoning (using the learned value functions) over images in different
directions and outputs short-term goals, that are consumed by the low-level
policy. The low-level policy employs classical mapping and planning to achieve
the short-term goal, and returns control to the high-level policy if it
achieves the short-term goal, or determines it to be infeasible. Black nodes
depict nodes stored by the high-level policy in the topological graph, and blue 
nodes show the value predictions in different directions from each of the black
nodes (size indicates predicted value, we use 12 uniformly sampled directions
but only show few for clarity). Current location is indicated by the hollow
circle. High-level policy outputs the most promising direction to pursue as the
short-term goal. Relative offset of this location from the current location
($\Delta\text{Pose}$) is passed to the low-level policy. Low-level policy
incrementally builds occupany map. It uses the fast-marching method to plan a path
to the desired short-term goal, and outputs low-level robot actions. Low-level
policy returns control on success (reaching the short-term goal), infeasible
goal (short-term goal determined to be in occupied space), or timeout.}
\figlabel{nav-policy}
\end{figure}
% \end{wrapfigure}

% \begin{figure}[t]
% \centering
% \insertW{1.0}{figures/nav_policy.pdf}
% \caption{An overview of the hierarchical navigation policy that shows the
% topological graph and the local policy, etc. Possibly using the maps that are
% getting generated by the SLAM system, and overlaying images on them and so
% forth.}
% \figlabel{nav-policy}
% \end{figure}

We use the learned value function $f(I,c)$ from \secref{vfv} %Section 3.1 (main paper)
%We use the learned value function $f(I,c)$ from Section 3.1 (main paper)
% \secref{vfv} 
in a hierarchical navigation policy for semantic navigation. Our hierarchical
policy is motivated by Chaplot \etal\cite{chaplot2020neural}, and consists of a
\textit{high-level} policy and a \textit{low-level} policy. The high-level
policy outputs short-term goals that are achieved by the low-level policy. The
high-level policy uses value predictions on images seen so far (at short-term
goal locations), to sample a short-term goal in the most promising direction.
This short-term goal is expressed as a relative offset from the agent's current
location. The low-level policy emits low-level robot actions to navigate to
this short-term goal, or returns that the short-term goal is infeasible. This
process is repeated, \ie, the high-level policy takes feedback from the
low-level policy, along with the image at the agent's new location to sample
the next short-term goal. We describe these two policies in more detail below.
\figref{nav-policy} shows an overview of this navigation policy.

\subsection{High-level policy} 
The high-level policy, $\Pi$ builds a hybrid spatial and topological
representation. 
It stores $360^\circ$ images along with their locations at each short-term goal
location. $360^\circ$ images are obtained by incrementally rotating the agent
12 times by $30^\circ$ each.  High-level policy also stores the value
prediction from $f(I,c)$ on these 12 images, for the category of interest $c$.
These 12 values denote the promise of exploring in the different directions for
reaching the objects of the desired class.  These predicted values are combined
with object detector output and a spatial consistency term to give the final
score:
\begin{equation}
f_{comb}(I,c) = \lambda_1 f(I,c)+ 
\lambda_2 \underbrace{\mathbbm{1}_{\geq
0.5}\left[D_{coco}(I,c)\right] \cdot \left(1+D_{coco}(I,c)\right)}_\text{Object
Detector} + 0.05
\lambda_3 \underbrace{\max\left(10-d,0\right)}_\text{Spatial Consistency}
\eqlabel{combo}
\end{equation}
where $f(I,c)$ is the semantic score for the object class of interest $c$ on
the image $I$, 
$D_{coco}(I,c)$ is the maximum confidence for Mask-RCNN detections
of class $c$ in $I$,
$d$ is the \textit{estimated} geodesic distance (based on the current map) of
the proposed short-term goal from the current agent position in meters, 
and $\mathbbm{1}_{\geq 0.5}$ is an indicator function that outputs $1$ if
$D_{coco}(I,c) \geq 0.5$, and $0$ otherwise. We set $\lambda_1 = \lambda_2 =
\lambda_3 = 1$. 
% variable which is 1 if $D_{coco}(I,c) \ge 0.5$, $0$. otherwise.

As it is expensive to get these images (it costs 12 steps), we only store these
at locations where the short-term policy returns control to the high-level
policy (we call these locations as \textit{semantic reasoning locations}, and
these are marked in \figref{nav-policy} with black dots). 

The high-level policy maintains a priority heap of all of these $12N$ values
(along with their location and associated direction vectors in the agent's
coordinate frame), where $N$ is the number of semantic reasoning nodes
currently stored in the topological graph. At each time step, the high-level
policy pops the highest of these $12N$ values\footnote{As we keep popping
values from the priority heap, there are $11N+1$ (and not $12N$) entries in the
heap at the popping time.} from the priority heap, and samples $k$ ($=100$)
short-term goals in this direction ($\pm 7^\circ$) that are between $1m$ and
$2m$ from the parent node.  These $k$ short-term goals are passed onto the
low-level policy, which pursues the first of these $k$ goals that is not known
to be infeasible, and returns control to the high-level policy if it succeeds, or
determines that the sampled short-term goal is infeasible or too far away.

\subsection{Low-level policy} The low-level policy uses metric occupancy
maps~\cite{elfes1989using} along with fast-marching method (FMM) path
planners~\cite{fmm} to incrementally plan paths to provided short-term goals.
The low-level policy filters the provided $k$ goals for feasibility (using the
current occupancy map). It takes the first one of these filtered short-term
goals, plans a path to it, and outputs planned robot actions. Low-level policy
continues to re-plan when the occupancy map updates.
% Once the short-term goal is selected, we estimate the number of steps required
% to reach it (using the FMM distance estimate). 
Low-level policy executes actions output from the FMM planner. It stops and
returns control when \textbf{a)} it has reached the goal, \textbf{b)} it has already executed
enough steps (based on estimate from original FMM computation), or \textbf{c)} the
short-term goal turns out to be infeasible or much further than originally
anticipated (as more of the map becomes visible).  We assume access to depth
images, and adapt code from the map and plan implementation from
\cite{map-plan-baseline}, to implement the low-level policy.
% FMM distance estimates. The step estimate accounts for a $180^\circ$ rotation
% and an average of one rotation step per forward step thus we estimate
% \[ steps = \left\lceil{\frac{2d}{s} + 6}\right\rceil \]
% where $d$ is the distance estimate from the FMM planner and $s$ is the step
% size of the agent.
% The low level policy takes either a) There is no step which can bring the agent
% closer to the short-term goal according to the FMM, b) the estimated number of
% steps have been executed, or c) the current estimated distance to the goal
% increases by more than $0.1$m between any two steps (i.e.  more of the map is
% uncovered which makes our short-term goal further away). 

%If none of the $k$ short-term goals are
%feasible, or the short-term goal is not achieved in \sg{$15$} steps, the
%low-level policy returns control to the high-level policy.

As our focus is on high-level semantic cues, for simplicity we assume access
to perfect agent pose for this hierarchical policy. This can be achieved using
additional sensors on the robot (depth cameras, and IMU units), or using a SLAM
system~\cite{mur2015orb}, or just with RGB images by using learned pose
estimators and free space estimators~\cite{chaplot2020learning}.

\subsection{Stopping Criteria}
We elaborate on the stopping criteria used for Policy Stop setting.  At every
semantic reasoning step, we compute a proxy measure for whether we are close to
an object of the desired category or not by using the depth image and \D{coco}. For
all high-scoring detections for class $c$ from \D{coco} (detection score more
than $\tau_c = 0.75$), we approximate the distance to the detected object
instance by the median depth value within the predicted instance segmentation
mask. If any detected instance is within a distance $d_c$, the agent emits a
stop signal. $d_c$ is a per-category hyper-parameter (as object sizes vary
drastically across categories). We set it using 100 episodes sampled in
\E{train}.

As noted in Section 4.2, we found that this hand-crafted stopping criteria
also led to best performance for all methods that we compare to (as opposed to
using the method's own stopping method). Threshold $\tau_c$ was fixed to 0.75
for all methods, while $d_c$ was optimized for each category \textit{for each
  method} on the same 100 episodes from \E{train} using the exact same
  procedure. For behavior cloning and RL methods, stopping criteria is
  evaluated at \textit{all} times steps, where as for our method and baselines
  based on our method, it is evaluated at every semantic reasoning step.

\clearpage

\section{Experimental Details}
\subsection{Environment Splits}
\begin{table}[h]
\footnotesize
\setlength{\tabcolsep}{5pt}
\caption{List of Gibson environments in different splits. See \secref{experiments} for details.}
%\caption{List of Gibson environments in different splits. See Section 4 for details.}
\tablelabel{splits}
\begin{tabular}{lp{4.9in}}
  \toprule
  \textbf{Split} & \textbf{Environments} \\ 
  \midrule
  \E{train} & Andover, Annona, Adairsville, Brown, Castor, Eagan, Goodfield,
  Goodwine, Kemblesville, Maugansville, Nuevo, Springerville, Stilwell,
  Sussex \\ \\ 
  \E{test} & Collierville, Corozal, Darden, Markleeville, Wiconisco \\ \\ 
  \E{video} & Airport, Albertville, Allensville, Anaheim, Ancor, Arkansaw,
  Athens, Bautista, Beechwood, Benevolence, Bohemia, Bonesteel, Bonnie,
  Broseley, Browntown, Byers, Chilhowie, Churchton, Clairton, Coffeen, Cosmos,
  Cottonport, Duarte, Emmaus, Forkland, Frankfort, Globe, Goffs, Goodyear,
  Hainesburg, Hanson, Highspire, Hildebran, Hillsdale, Hiteman, Hominy, Irvine,
  Klickitat, Lakeville, Leonardo, Lindenwood, Lynchburg, Maida, Marland,
  Marstons, Martinville, Merom, Micanopy, Mifflinburg, Musicks, Neibert,
  Neshkoro, Newcomb, Newfields, Onaga, Oyens, Pamelia, Parole, Pinesdale,
  Pomaria, Potterville, Ranchester, Readsboro, Rogue, Rosser, Shelbiana,
  Shelbyville, Silas, Soldier, Stockman, Sugarville, Sunshine, Sweatman,
  Thrall, Tilghmanton, Timberon, Tokeland, Tolstoy, Tyler, Victorville,
  Wainscott, Willow, Wilseyville, Winooski, Woodbine \\ 
  \bottomrule
\end{tabular}
\end{table}

\subsection{Difficulty Distribution of Test Episodes}
% \begin{wrapfigure}[h]
\begin{wrapfigure}[10]{R}{0.45\textwidth}
  \vspace{-1.5cm}
  \centering
  \includegraphics[trim=17pt 17pt 153pt 30pt, clip, width=1.0\linewidth]{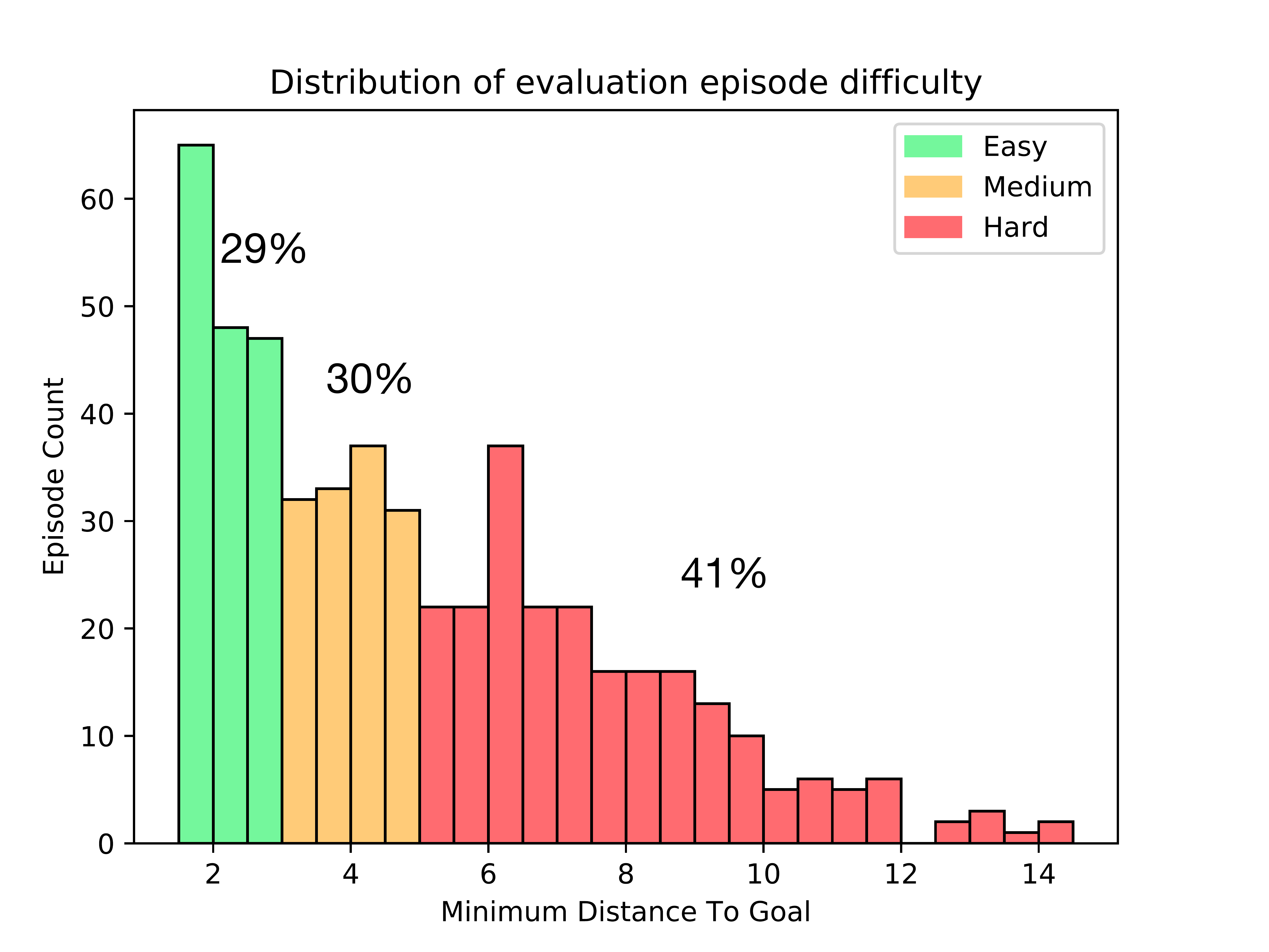}
  % \insertWL{1.0}
\end{wrapfigure}
% \end{figure}
We plot the distribution of difficulty (distance to nearest object of interest)
of the evaluation episodes in \E{test} in figure on right. We group these
episodes into 3 difficulty levels, based on distance to the nearest instance of
the target category: \textit{easy} ($\leq 3m$, green), \textit{medium} ($3m$ to
$5m$, orange), and \textit{hard} ($5m$ to $15m$, red).  In total there were
$313$ easy, $324$ medium and $438$ hard episodes.
There were 200, 250, 200, 125, 300 episodes each for object categories 
\textit{Bed}, \textit{Chair}, \textit{Couch}, \textit{Dining Table},
\textit{Toilet} respectively.

\subsection{Generation of \V{syn}}
We use environments in \E{video} to render out egocentric navigation tours. We
employ a path planner to compute shortest path between \textit{random} pairs of
points in each environment. We render out panorama images (4 images: straight
facing, left facing, back facing, and right facing, relative to the direction
of motion) along these shortest paths and throw out the sequence of actions
that were executed, to arrive at the dataset of videos \V{syn}. To make
these tours more realistic, we execute a random action with $20\%$ probability
at each time step (and replan accordingly). We sample 300 trajectories in each
of the 85 environments. Average trajectory length is 40 steps.

\subsection{More Implementation Details}
We note further implementation details for our method and baselines.
\begin{enumerate}
\item Topological Exploration and Detection Seeker are implemented by setting
$(\lambda_1, \lambda_2, \lambda_3)$ to be $(0,0,1)$, and $(0,1,1)$
respectively in \eqref{combo}. This assures a fair comparison between the three
methods, and tests the effectiveness of our learned function $f(I,c)$.
\item For End-to-End RL, we experimented with different architectures as noted
in Table 1 in the main paper. Baselines as part of Habitat~\cite{habitat} use a
3 layered CNN (denoted by 3CNN and SimpleCNN interchangeably in the main paper)
to represent RGB, Depth or RGB-D input. We report performance with this default
network (RL (RGB-D 3CNN, RL Depth 3CNN)) in Table 1 in main paper. We found that
using a ResNet-18 model (initialized by pre-training on ImageNet) worked better
than using this SimpleCNN to represent RGB images. Thus we additionally also
reported performance with ResNet-18 models (RGB-D ResNet-18+3CNN, RGB
ResNet-18).  For RGB-D models, we could only use ResNet-18 for the RGB part.
Depth is still processed through the same 3-layer CNN (as there is no standard
initialization for Depth models that is commonly used). Output from ResNet-18
for RGB and 3CNN for Depth were concatenated before feeding into the LSTM
model.
\item Our Q-learning models were optimized using Adam with a learning rate of
  $10^{-4}$, $\beta_1=0.9$ and $\beta_2=0.999$. Model was trained for
$300K$ mini-batches of size $16$ and the model after the last update was used for experiments.
\item \textbf{Architecture of Q-network:} The architecture of the Q-network was
based off of ResNet-18. We used a ResNet-18 pretrained on ImageNet removing the
last convolution layer and all later layers. We add to the pre-trained head, an
additional convolution layer with kernel size 3 $\times$ 3 and 64 channels.
After this convolution layer there are 3 fully-connected layers of size
$[512,256,15]$ respectively. The output of the final layer is reshaped to 3
$\times$ 5 to represent the value of taking each of the 3 possible actions with
respect to the 5 possible classes.
\item \textbf{Compute Infrastructure:} All experiments were conducted on a
single GPU server with 8 GPUs (NVidia 2080 Ti). Model training for our method
was done on a single GPU and took 22 hours.
\end{enumerate}

% As noted in the paper
% In the Policy Stop experiments (Section 4, main paper) the stopping criteria
% we found to be most was based on the pre-trained Mask-RCNN detector
% \cite{wu2019detectron2, he2017mask}. Detections are found in the RGB input and
% regions with confidence greater than a tunable threshold (we used 0.75 for all
% experiments) are considered object instances. For each object instance, the
% predicted mask from the Mask-RCNN is registered against the depth input on the
% same frame, and we estimate the distance to that object instance to be the
% median of all depth pixels covered by the mask. If the minimum depth of any of
% the object instances for the class if interest is less than some threshold, the
% stop criteria is met and the episode terminates. To give the most generous
% comparison, we use this stopping criteria for all methods evaluated in the
% Policy Stop setting.
% 
% To determine the distance threshold for each model and object class, each
%   model was run for 500 steps on the same set of 100 episodes sampled in
%   \E{train}. Using the object instance distance estimates, we sweep over
%   potential thresholds from $0.5$ to $2$ to find the threshold for each
%   category that gives the best SPL. Final evaluation is performed using the
%   stopping criteria described above with the thresholds optimized for each model
%   and category on the \E{train} episodes.

\section{Detailed Results}
\begin{figure}[h]
    \centering
    \insertW{0.9}{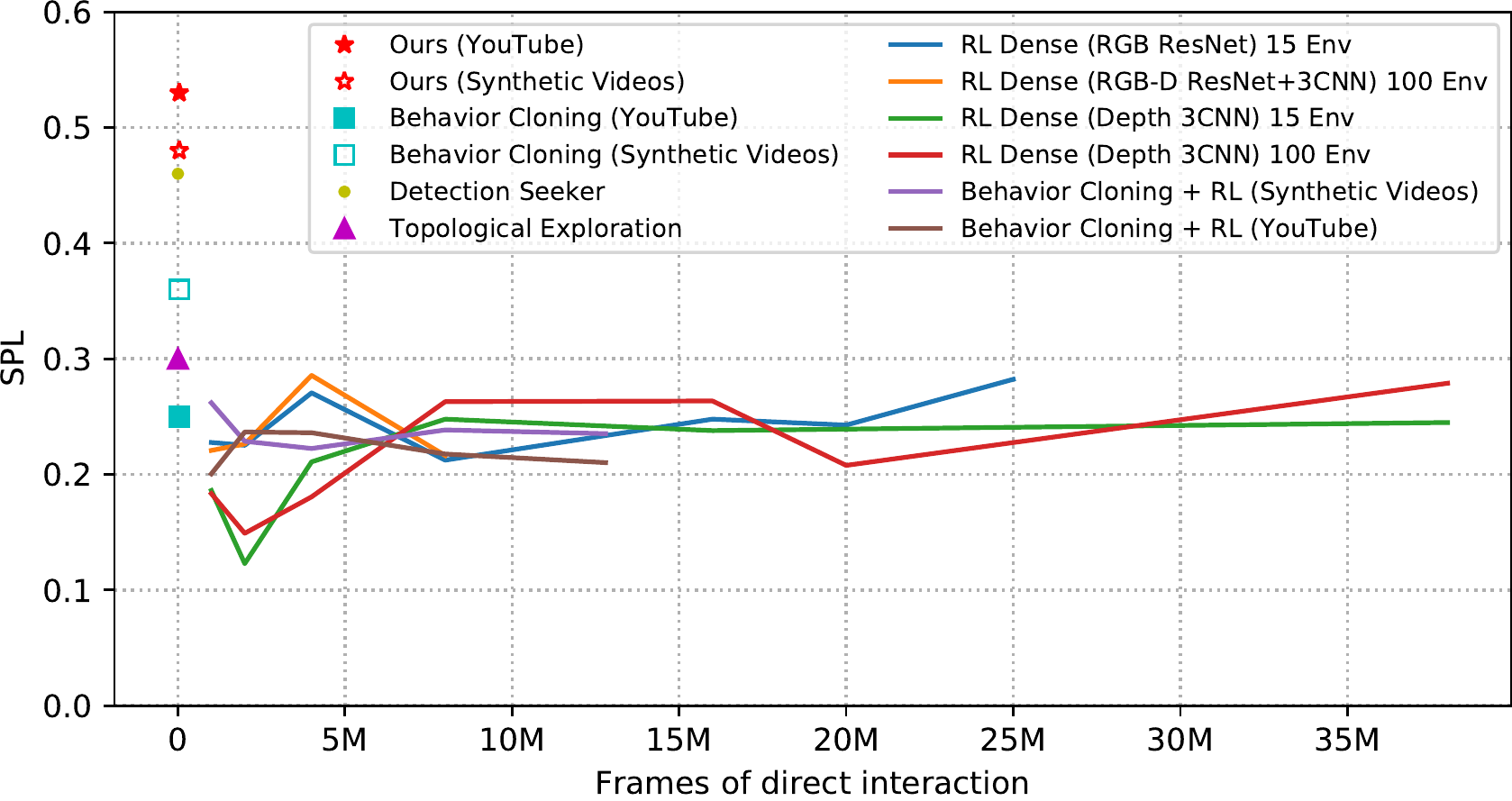}\\
    \caption{Oracle Stop SPL for various methods against the number of direct
    interaction samples used.}
    % \sg{Can we say if these are dense or sparse
    % reward policies? Perhaps also thinner so as to fit on previous page.
    % Perhaps also to note performances by different symbols, colors may not be
    % visible in print. Alternative is to place the text next to the dots (using
    % ax.text)}}
    \figlabel{spl_curve}
  \end{figure}

% \subsection{Main Results (corresponding to Section 4.1)}
% We report category wise results, and also across different difficulty levels.
\renewcommand{\arraystretch}{1.1}
\begin{table}[h]
\setlength{\tabcolsep}{6pt}
\centering
\caption{\textbf{Results}: SPL and Success Rate for ObjectGoal wth
\textbf{Oracle Stop} in novel environments \E{test} by episode difficulty.
Details in \secref{results}.}
%Details in Section 4.2.}
% in oracle stop and policy stop settings. 
\tablelabel{os-diff-eval}
\resizebox{\textwidth}{!}{
\begin{tabular}{lccccccc}
\toprule
\multirow{2}{*}{\textbf{Method}} & \multicolumn{3}{c}{\textbf{Training Supervision}} & \multicolumn{4}{c}{\textbf{SPL}}\\
\cmidrule(lr){2-4} \cmidrule(lr){5-8}
  & \textbf{\# Active Frames} & \textbf{Reward} & \textbf{Other} & \textbf{Easy} & \textbf{Medium} & \textbf{Hard} & \textbf{Overall} \\

\midrule
Topological Exploration           & -                                                        & -               & - & 0.47 $\pm$ 0.03 & 0.31 $\pm$ 0.03 & 0.16 $\pm$ 0.02 & 0.30 $\pm$ 0.02\\
Detection Seeker                  & -                                                        & -               & - & 0.73 $\pm$ 0.03 & 0.53 $\pm$ 0.03 & 0.22 $\pm$ 0.02 & 0.46 $\pm$ 0.02\\
RL (RGB-D ResNet+3CNN)            & 100K (\E{train})                                         & Sparse          & - & 0.30 $\pm$ 0.03 & 0.19 $\pm$ 0.03 & 0.07 $\pm$ 0.02 & 0.17 $\pm$ 0.02\\
RL (RGB-D ResNet+3CNN)            & 10M (\E{train} $\cup$ \E{video})                         & Dense           & - & 0.42 $\pm$ 0.03 & 0.29 $\pm$ 0.03 & 0.11 $\pm$ 0.02 & 0.26 $\pm$ 0.02\\
RL (RGB-D 3CNN)                   & 38M (\E{train} $\cup$ \E{video})                         & Dense           & - & 0.42 $\pm$ 0.03 & 0.32 $\pm$ 0.03 & 0.15 $\pm$ 0.02 & 0.28 $\pm$ 0.02\\
RL (RGB ResNet)                   & 20M (\E{train})                                          & Dense           & - & 0.40 $\pm$ 0.04 & 0.30 $\pm$ 0.03 & 0.21 $\pm$ 0.02 & 0.29 $\pm$ 0.02\\
RL (Depth 3CNN)                   & 38M (\E{train})                                          & Dense           & - & 0.40 $\pm$ 0.04 & 0.24 $\pm$ 0.03 & 0.15 $\pm$ 0.02 & 0.25 $\pm$ 0.02\\
Behavior Cloning                  & 40K (\E{train})                                          & -               & \Vhat{yt} & 0.44 $\pm$ 0.04 & 0.29 $\pm$ 0.03 & 0.07 $\pm$ 0.01 & 0.25 $\pm$ 0.02\\
Behavior Cloning + RL             & 12M (\E{train})                                          & Dense               & \Vhat{yt} & 0.41 $\pm$ 0.03 & 0.26 $\pm$ 0.03 & 0.09 $\pm$ 0.01 & 0.24 $\pm$ 0.02\\
Our (Value Learning from Videos)  & 40K (\E{train})                                          & -               & \Vhat{yt} & \textbf{0.75} $\pm$ 0.03 & \textbf{0.63} $\pm$ 0.03 & \textbf{0.30} $\pm$ 0.02 & \textbf{0.53} $\pm$ 0.02\\ \midrule
Behavior Cloning                  & 40K (\E{train})                                          & -               & \Vhat{syn} & 0.48 $\pm$ 0.04 & 0.34 $\pm$ 0.03 & 0.24 $\pm$ 0.02 & 0.34 $\pm$ 0.02\\
Behavior Cloning + RL             & 12M (\E{train})                                          & Dense               & \Vhat{syn} & 0.42 $\pm$ 0.03 & 0.22 $\pm$ 0.03 & 0.11 $\pm$ 0.02 & 0.24 $\pm$ 0.02\\
Our (Value Learning from Videos)  & 40K (\E{train})                                          & -               & \Vhat{syn} & 0.71 $\pm$ 0.03 & 0.55 $\pm$ 0.03 & 0.26 $\pm$ 0.02 & 0.48 $\pm$ 0.02\\ \midrule
Strong Supervision Values       & \multicolumn{3}{c}{Labeled Maps (\E{video})} & 0.73 $\pm$ 0.03 & 0.60 $\pm$ 0.03 & 0.33 $\pm$ 0.02 & 0.53 $\pm$ 0.02\\
Strong Supervision + VLV (Ours) & \multicolumn{3}{c}{Labeled Maps (\E{video}) + \Vhat{yt}} & 0.80 $\pm$ 0.03 & 0.65 $\pm$ 0.03 & 0.35 $\pm$ 0.03 & 0.57 $\pm$ 0.02\\
\midrule
\midrule
\multirow{2}{*}{\textbf{Method}} & \multicolumn{3}{c}{\textbf{Training Supervision}} & \multicolumn{4}{c}{\textbf{Success Rate}}\\
\cmidrule(lr){2-4} \cmidrule(lr){5-8}
  & \textbf{\# Active Frames} & \textbf{Reward} & \textbf{Other} & \textbf{Easy} & \textbf{Medium} & \textbf{Hard} & \textbf{Overall} \\
\midrule 
Topological Exploration           & -                                                        & -               & - & 0.89 $\pm$ 0.03 & 0.80 $\pm$ 0.04 & 0.41 $\pm$ 0.04 & 0.67 $\pm$ 0.02\\
Detection Seeker                  & -                                                        & -               & - & 0.95 $\pm$ 0.02 & \textbf{0.90} $\pm$ 0.03 & 0.50 $\pm$ 0.04 & 0.75 $\pm$ 0.02\\
RL (RGB-D ResNet+3CNN)            & 100K (\E{train})                                         & Sparse          & - & 0.62 $\pm$ 0.05 & 0.41 $\pm$ 0.05 & 0.15 $\pm$ 0.03 & 0.37 $\pm$ 0.02\\
RL (RGB-D ResNet+3CNN)            & 10M (\E{train} $\cup$ \E{video})                         & Dense           & - & 0.81 $\pm$ 0.04 & 0.65 $\pm$ 0.05 & 0.26 $\pm$ 0.03 & 0.54 $\pm$ 0.03\\
RL (RGB-D 3CNN)                   & 38M (\E{train} $\cup$ \E{video})                         & Dense           & - & 0.79 $\pm$ 0.04 & 0.65 $\pm$ 0.04 & 0.35 $\pm$ 0.04 & 0.57 $\pm$ 0.03\\
RL (RGB ResNet)                   & 20M (\E{train})                                          & Dense           & - & 0.75 $\pm$ 0.04 & 0.59 $\pm$ 0.05 & 0.40 $\pm$ 0.04 & 0.56 $\pm$ 0.03\\
RL (Depth 3CNN)                   & 38M (\E{train})                                          & Dense           & - & 0.73 $\pm$ 0.04 & 0.58 $\pm$ 0.05 & 0.32 $\pm$ 0.04 & 0.52 $\pm$ 0.02\\
Behavior Cloning                  & 40K (\E{train})                                          & -               & \Vhat{yt} & 0.81 $\pm$ 0.04 & 0.68 $\pm$ 0.04 & 0.21 $\pm$ 0.03 & 0.53 $\pm$ 0.02\\
Behavior Cloning + RL             & 12M (\E{train})                                          & Dense               & \Vhat{yt} & 0.86 $\pm$ 0.03 & 0.69 $\pm$ 0.04 & 0.29 $\pm$ 0.03 & 0.58 $\pm$ 0.02\\
Our (Value Learning from Videos)  & 40K (\E{train})                                          & -               & \Vhat{yt} & 0.95 $\pm$ 0.02 & \textbf{0.90} $\pm$ 0.03 & \textbf{0.58} $\pm$ 0.04 & \textbf{0.79} $\pm$ 0.02\\ \midrule
Behavior Cloning                  & 40K (\E{train})                                          & -               & \Vhat{syn} & 0.84 $\pm$ 0.03 & 0.74 $\pm$ 0.04 & 0.55 $\pm$ 0.04 & 0.69 $\pm$ 0.02\\
Behavior Cloning + RL             & 12M (\E{train})                                          & Dense               & \Vhat{syn} & 0.83 $\pm$ 0.03 & 0.62 $\pm$ 0.04 & 0.31 $\pm$ 0.04 & 0.55 $\pm$ 0.02\\
Our (Value Learning from Videos)  & 40K (\E{train})                                          & -               & \Vhat{syn} & \textbf{0.96} $\pm$ 0.02 & 0.88 $\pm$ 0.03 & 0.51 $\pm$ 0.04 & 0.75 $\pm$ 0.02\\ \midrule
Strong Supervision Values       & \multicolumn{3}{c}{Labeled Maps (\E{video})} & 0.95 $\pm$ 0.02 & 0.92 $\pm$ 0.03 & 0.64 $\pm$ 0.04 & 0.81 $\pm$ 0.02\\
Strong Supervision + VLV (Ours) & \multicolumn{3}{c}{Labeled Maps (\E{video}) + \Vhat{yt}} & 0.98 $\pm$ 0.01 & 0.94 $\pm$ 0.02 & 0.62 $\pm$ 0.04 & 0.82 $\pm$ 0.02\\
\bottomrule
\vspace{-15pt}
\end{tabular}}
\end{table}
%
%RL (RGB-D)              & 38M (\E{train})                  & Dense  & -          & 0.25 $\pm$ 0.02 & 0.53 $\pm$ 0.02 &                 & \\
%RL (RGB-D)              & 38M (\E{train})                  & Sparse & -          & 0.15 $\pm$ 0.01 & 0.32 $\pm$ 0.02 &                 & \\
% 1. RL(RGB-D ResNet + SimpleCNN, 100K, sparse, 15), SG trains.
% 2. RL(RGB-D SimpleCNN + SimpleCNN, 38M, 100) Finished training
% 3. RL(Depth SimpleCNN, 38M, 15), Already in paper : in
% 4. RL(RGB-D, ResNet + SimpleCNN, 10M, dense, 100), Finish training in 2 hours : in
% 5. RL(RGB ResNet, 20M, dense, 15), Already trained. :in
%
% rl rgbd sparse
% & 0.28 $\pm$ 0.04 & 0.55 $\pm$ 0.05& 0.16 $\pm$ 0.02 & 0.36 $\pm$ 0.04& 0.06 $\pm$ 0.01 & 0.16 $\pm$ 0.03& 0.00 $\pm$ 0.01 & 0.01 $\pm$ 0.02& 0.15 $\pm$ 0.01 & 0.32 $\pm$ 0.02

% \input{sup-results-diff-sr}
\renewcommand{\arraystretch}{1.1}
\begin{table}[h]
\setlength{\tabcolsep}{4pt}
\centering
\caption{\textbf{Results}: SPL and Success Rate for ObjectGoal wth
\textbf{Oracle Stop} in novel environments \E{test} by object class. Details in
\secref{results}.}
%Section 4.2.}
% in oracle stop and policy stop settings. 
\tablelabel{os-class-eval}
\resizebox{\textwidth}{!}{
\begin{tabular}{lccccccccc}
\toprule
\multirow{2}{*}{\textbf{Method}} & \multicolumn{3}{c}{\textbf{Training Supervision}} & \multicolumn{5}{c}{\textbf{SPL}}\\
\cmidrule(lr){2-4} \cmidrule(lr){5-9}
  & \textbf{\# Active Frames} & \textbf{Reward} & \textbf{Other} & \textbf{Bed} & \textbf{Chair} & \textbf{Couch} & \textbf{Dining Table} & \textbf{Toilet} \\
% & Frames & & & & & & \\

\midrule
Topological Exploration           & -                                                        & -               & - & 0.35 $\pm$ 0.04 & 0.39 $\pm$ 0.03 & 0.29 $\pm$ 0.03 & 0.27 $\pm$ 0.04 & 0.19 $\pm$ 0.03\\
Detection Seeker                  & -                                                        & -               & - & \textbf{0.49} $\pm$ 0.05 & 0.64 $\pm$ 0.04 & 0.48 $\pm$ 0.04 & 0.53 $\pm$ 0.06 & 0.26 $\pm$ 0.03\\
RL (RGB-D ResNet+3CNN)            & 100K (\E{train})                                         & Sparse          & - & 0.12 $\pm$ 0.03 & 0.29 $\pm$ 0.04 & 0.16 $\pm$ 0.03 & 0.20 $\pm$ 0.05 & 0.11 $\pm$ 0.03\\
RL (RGB-D ResNet+3CNN)            & 10M (\E{train} $\cup$ \E{video})                         & Dense           & - & 0.24 $\pm$ 0.04 & 0.37 $\pm$ 0.04 & 0.29 $\pm$ 0.04 & 0.38 $\pm$ 0.05 & 0.09 $\pm$ 0.02\\
RL (RGB-D 3CNN)                   & 38M (\E{train} $\cup$ \E{video})                         & Dense           & - & 0.30 $\pm$ 0.04 & 0.39 $\pm$ 0.04 & 0.26 $\pm$ 0.04 & 0.36 $\pm$ 0.05 & 0.14 $\pm$ 0.03\\
RL (RGB ResNet)                   & 20M (\E{train})                                          & Dense           & - & 0.30 $\pm$ 0.04 & 0.44 $\pm$ 0.04 & 0.24 $\pm$ 0.04 & 0.27 $\pm$ 0.05 & 0.21 $\pm$ 0.03\\
RL (Depth 3CNN)                   & 38M (\E{train})                                          & Dense           & - & 0.29 $\pm$ 0.04 & 0.32 $\pm$ 0.04 & 0.26 $\pm$ 0.04 & 0.32 $\pm$ 0.05 & 0.13 $\pm$ 0.02\\
Behavior Cloning                  & 40K (\E{train})                                          & -               & \Vhat{yt} & 0.24 $\pm$ 0.04 & 0.34 $\pm$ 0.04 & 0.28 $\pm$ 0.04 & 0.36 $\pm$ 0.05 & 0.10 $\pm$ 0.02\\
Behavior Cloning + RL             & 12M (\E{train})                                          & Dense               & \Vhat{yt} & 0.23 $\pm$ 0.04 & 0.33 $\pm$ 0.03 & 0.25 $\pm$ 0.03 & 0.28 $\pm$ 0.04 & 0.13 $\pm$ 0.02\\
Our (Value Learning from Videos)  & 40K (\E{train})                                          & -               & \Vhat{yt} & \textbf{0.49} $\pm$ 0.04 & \textbf{0.68} $\pm$ 0.03 & \textbf{0.60} $\pm$ 0.04 & \textbf{0.71} $\pm$ 0.05 & 0.32 $\pm$ 0.03\\ \midrule
Behavior Cloning                  & 40K (\E{train})                                          & -               & \Vhat{syn} & 0.36 $\pm$ 0.05 & 0.45 $\pm$ 0.04 & 0.37 $\pm$ 0.04 & 0.35 $\pm$ 0.05 & 0.20 $\pm$ 0.03\\
Behavior Cloning + RL             & 12M (\E{train})                                          & Dense               & \Vhat{syn} & 0.21 $\pm$ 0.03 & 0.31 $\pm$ 0.03 & 0.23 $\pm$ 0.03 & 0.35 $\pm$ 0.05 & 0.14 $\pm$ 0.03\\
Our (Value Learning from Videos)  & 40K (\E{train})                                          & -               & \Vhat{syn} & 0.44 $\pm$ 0.04 & 0.60 $\pm$ 0.04 & 0.48 $\pm$ 0.04 & 0.56 $\pm$ 0.06 & \textbf{0.37} $\pm$ 0.04\\ \midrule
Strong Supervision Values       & \multicolumn{3}{c}{Labeled Maps (\E{video})} & 0.46 $\pm$ 0.04 & 0.59 $\pm$ 0.03 & 0.57 $\pm$ 0.04 & 0.68 $\pm$ 0.05 & 0.43 $\pm$ 0.03\\
Strong Supervision + VLV (Ours) & \multicolumn{3}{c}{Labeled Maps (\E{video}) + \Vhat{yt}} & 0.50 $\pm$ 0.04 & 0.69 $\pm$ 0.03 & 0.58 $\pm$ 0.04 & 0.77 $\pm$ 0.04 & 0.43 $\pm$ 0.04\\
\midrule \midrule
\multirow{2}{*}{\textbf{Method}} & \multicolumn{3}{c}{\textbf{Training Supervision}} & \multicolumn{5}{c}{\textbf{Success rate}}\\
\cmidrule(lr){2-4} \cmidrule(lr){5-9}
  & \textbf{\# Active Frames} & \textbf{Reward} & \textbf{Other} & \textbf{Bed} & \textbf{Chair} & \textbf{Couch} & \textbf{Dining Table} & \textbf{Toilet} \\
% & Frames & & & & & & \\
\midrule
Topological Exploration           & -                                                        & -               & - & 0.67 $\pm$ 0.05 & 0.85 $\pm$ 0.04 & 0.71 $\pm$ 0.05 & 0.68 $\pm$ 0.07 & 0.48 $\pm$ 0.05\\
Detection Seeker                  & -                                                        & -               & - & 0.74 $\pm$ 0.05 & 0.92 $\pm$ 0.03 & 0.80 $\pm$ 0.05 & 0.80 $\pm$ 0.06 & 0.57 $\pm$ 0.05\\
RL (RGB-D ResNet+3CNN)            & 100K (\E{train})                                         & Sparse          & - & 0.36 $\pm$ 0.06 & 0.54 $\pm$ 0.05 & 0.34 $\pm$ 0.05 & 0.43 $\pm$ 0.07 & 0.21 $\pm$ 0.04\\
RL (RGB-D ResNet+3CNN)            & 10M (\E{train} $\cup$ \E{video})                         & Dense           & - & 0.48 $\pm$ 0.06 & 0.77 $\pm$ 0.04 & 0.62 $\pm$ 0.05 & 0.86 $\pm$ 0.05 & 0.19 $\pm$ 0.04\\
RL (RGB-D 3CNN)                   & 38M (\E{train} $\cup$ \E{video})                         & Dense           & - & 0.66 $\pm$ 0.05 & 0.74 $\pm$ 0.05 & 0.57 $\pm$ 0.06 & 0.74 $\pm$ 0.07 & 0.30 $\pm$ 0.04\\
RL (RGB ResNet)                   & 20M (\E{train})                                          & Dense           & - & 0.64 $\pm$ 0.05 & 0.74 $\pm$ 0.05 & 0.55 $\pm$ 0.06 & 0.50 $\pm$ 0.07 & 0.38 $\pm$ 0.04\\
RL (Depth 3CNN)                   & 38M (\E{train})                                          & Dense           & - & 0.54 $\pm$ 0.06 & 0.71 $\pm$ 0.05 & 0.54 $\pm$ 0.06 & 0.63 $\pm$ 0.07 & 0.30 $\pm$ 0.04\\
Behavior Cloning                  & 40K (\E{train})                                          & -               & \Vhat{yt} & 0.46 $\pm$ 0.06 & 0.74 $\pm$ 0.05 & 0.63 $\pm$ 0.06 & 0.72 $\pm$ 0.07 & 0.24 $\pm$ 0.04\\
Behavior Cloning + RL             & 12M (\E{train})                                          & Dense               & \Vhat{yt} & 0.59 $\pm$ 0.06 & 0.79 $\pm$ 0.04 & 0.66 $\pm$ 0.06 & 0.67 $\pm$ 0.07 & 0.30 $\pm$ 0.04\\
Our (Value Learning from Videos)  & 40K (\E{train})                                          & -               & \Vhat{yt} & \textbf{0.76} $\pm$ 0.05 & \textbf{0.94} $\pm$ 0.03 & \textbf{0.86} $\pm$ 0.04 & \textbf{0.91} $\pm$ 0.04 & 0.57 $\pm$ 0.05\\ \midrule
Behavior Cloning                  & 40K (\E{train})                                          & -               & \Vhat{syn} & 0.70 $\pm$ 0.05 & 0.84 $\pm$ 0.04 & 0.70 $\pm$ 0.05 & 0.82 $\pm$ 0.06 & 0.51 $\pm$ 0.05\\ 
Behavior Cloning + RL             & 12M (\E{train})                                          & Dense               & \Vhat{syn} & 0.59 $\pm$ 0.06 & 0.77 $\pm$ 0.04 & 0.60 $\pm$ 0.06 & 0.68 $\pm$ 0.07 & 0.25 $\pm$ 0.04\\
Our (Value Learning from Videos)  & 40K (\E{train})                                          & -               & \Vhat{syn} & 0.70 $\pm$ 0.05 & 0.90 $\pm$ 0.03 & 0.77 $\pm$ 0.05 & 0.82 $\pm$ 0.06 & \textbf{0.62} $\pm$ 0.05\\ \midrule
Strong Supervision Values       & \multicolumn{3}{c}{Labeled Maps (\E{video})} & 0.72 $\pm$ 0.05 & 0.91 $\pm$ 0.03 & 0.86 $\pm$ 0.04 & 0.94 $\pm$ 0.03 & 0.71 $\pm$ 0.04\\
Strong Supervision + VLV (Ours) & \multicolumn{3}{c}{Labeled Maps (\E{video}) + \Vhat{yt}} & 0.81 $\pm$ 0.05 & 0.94 $\pm$ 0.02 & 0.82 $\pm$ 0.04 & 0.94 $\pm$ 0.03 & 0.69 $\pm$ 0.04\\
\bottomrule
\vspace{-15pt}
\end{tabular}}
\end{table}

%RL (RGB-D)              & 38M (\E{train})                  & Dense  & -          & 0.25 $\pm$ 0.02 & 0.53 $\pm$ 0.02 &                 & \\
%RL (RGB-D)              & 38M (\E{train})                  & Sparse & -          & 0.15 $\pm$ 0.01 & 0.32 $\pm$ 0.02 &                 & \\
% 1. RL(RGB-D ResNet + SimpleCNN, 100K, sparse, 15), SG trains.
% 2. RL(RGB-D SimpleCNN + SimpleCNN, 38M, 100) Finished training
% 3. RL(Depth SimpleCNN, 38M, 15), Already in paper : in
% 4. RL(RGB-D, ResNet + SimpleCNN, 10M, dense, 100), Finish training in 2 hours : in
% 5. RL(RGB ResNet, 20M, dense, 15), Already trained. :in

% rl rgbd sparse
% & 0.28 $\pm$ 0.04 & 0.55 $\pm$ 0.05& 0.16 $\pm$ 0.02 & 0.36 $\pm$ 0.04& 0.06 $\pm$ 0.01 & 0.16 $\pm$ 0.03& 0.00 $\pm$ 0.01 & 0.01 $\pm$ 0.02& 0.15 $\pm$ 0.01 & 0.32 $\pm$ 0.02

% \input{sup-results-class-sr}

\renewcommand{\arraystretch}{1.1}
\begin{table}[h]
\setlength{\tabcolsep}{8pt}
\centering
\caption{\textbf{Results}: SPL and Success Rate for ObjectGoal wth
\textbf{Policy Stop} in novel environments \E{test} by episode difficulty.
Details in \secref{results}.}
%Details in Section 4.2.}
% in oracle stop and policy stop settings. 
\tablelabel{os-stop-diff-eval}
\resizebox{\textwidth}{!}{
\begin{tabular}{lccccccc}
\toprule
\multirow{2}{*}{\textbf{Method}} & \multicolumn{3}{c}{\textbf{Training Supervision}} & \multicolumn{4}{c}{\textbf{SPL}}\\
\cmidrule(lr){2-4} \cmidrule(lr){5-8}
  & \textbf{\# Active Frames} & \textbf{Reward} & \textbf{Other} & \textbf{Easy} & \textbf{Medium} & \textbf{Hard} & \textbf{Overall} \\
% & Frames & & & & & & \\

\midrule
Topological Exploration           & -                                                        & -               & - & 0.22 $\pm$ 0.03 & 0.12 $\pm$ 0.02 & 0.06 $\pm$ 0.01 & 0.13 $\pm$ 0.01\\
Detection Seeker                  & -                                                        & -               & - & 0.31 $\pm$ 0.04 & 0.22 $\pm$ 0.03 & 0.09 $\pm$ 0.01 & 0.19 $\pm$ 0.02\\
RL (RGB ResNet)                   & 20M (\E{train})                                          & Dense           & - & 0.10 $\pm$ 0.02 & 0.09 $\pm$ 0.02 & 0.05 $\pm$ 0.01 & 0.08 $\pm$ 0.01\\
Behavior Cloning                  & 40K (\E{train})                                          & -               & \Vhat{yt} & 0.16 $\pm$ 0.03 & 0.10 $\pm$ 0.02 & 0.02 $\pm$ 0.01 & 0.08 $\pm$ 0.01\\
Our (Value Learning from Videos)  & 40K (\E{train})                                          & -               & \Vhat{yt} & \textbf{0.32} $\pm$ 0.04 & \textbf{0.29} $\pm$ 0.03 & 0.11 $\pm$ 0.02 & \textbf{0.22} $\pm$ 0.02\\ \midrule
Behavior Cloning                  & 40K (\E{train})                                          & -               & \Vhat{syn} & 0.13 $\pm$ 0.02 & 0.12 $\pm$ 0.02 & 0.07 $\pm$ 0.01 & 0.10 $\pm$ 0.01\\
Our (Value Learning from Videos)  & 40K (\E{train})                                          & -               & \Vhat{syn} & 0.29 $\pm$ 0.04 & 0.23 $\pm$ 0.03 & \textbf{0.13} $\pm$ 0.02 & 0.21 $\pm$ 0.02\\ \midrule
Strong Supervision Values       & \multicolumn{3}{c}{Labeled Maps (\E{video})} & 0.34 $\pm$ 0.04 & 0.25 $\pm$ 0.03 & 0.15 $\pm$ 0.02 & 0.24 $\pm$ 0.02\\
Strong Supervision + VLV (Ours) & \multicolumn{3}{c}{Labeled Maps (\E{video}) + \Vhat{yt}} & 0.31 $\pm$ 0.04 & 0.29 $\pm$ 0.03 & 0.13 $\pm$ 0.02 & 0.23 $\pm$ 0.02\\
\midrule \midrule
\multirow{2}{*}{\textbf{Method}} & \multicolumn{3}{c}{\textbf{Training
Supervision}} & \multicolumn{4}{c}{\textbf{Success Rate}}\\
\cmidrule(lr){2-4} \cmidrule(lr){5-8}
  & \textbf{\# Active Frames} & \textbf{Reward} & \textbf{Other} & \textbf{Easy} & \textbf{Medium} & \textbf{Hard} & \textbf{Overall} \\
% & Frames & & & & & & \\

\midrule
Topological Exploration           & -                                                        & -               & - & 0.43 $\pm$ 0.04 & 0.31 $\pm$ 0.04 & 0.17 $\pm$ 0.03 & 0.29 $\pm$ 0.02\\
Detection Seeker                  & -                                                        & -               & - & 0.52 $\pm$ 0.05 & 0.43 $\pm$ 0.05 & 0.21 $\pm$ 0.03 & 0.37 $\pm$ 0.02\\
RL (RGB ResNet)                   & 20M (\E{train})                                          & Dense           & - & 0.27 $\pm$ 0.04 & 0.26 $\pm$ 0.04 & 0.12 $\pm$ 0.03 & 0.21 $\pm$ 0.02\\
Behavior Cloning                  & 40K (\E{train})                                          & -               & \Vhat{yt} & 0.36 $\pm$ 0.04 & 0.25 $\pm$ 0.04 & 0.05 $\pm$ 0.02 & 0.20 $\pm$ 0.02\\
Our (Value Learning from Videos)  & 40K (\E{train})                                          & -               & \Vhat{yt} & \textbf{0.53} $\pm$ 0.04 & \textbf{0.48} $\pm$ 0.05 & 0.22 $\pm$ 0.03 & \textbf{0.39} $\pm$ 0.02\\ \midrule
Behavior Cloning                  & 40K (\E{train})                                          & -               & \Vhat{syn} & 0.35 $\pm$ 0.04 & 0.28 $\pm$ 0.04 & 0.18 $\pm$ 0.03 & 0.26 $\pm$ 0.02\\
Our (Value Learning from Videos)  & 40K (\E{train})                                          & -               & \Vhat{syn} & 0.50 $\pm$ 0.05 & 0.42 $\pm$ 0.05 & \textbf{0.26} $\pm$ 0.03 & 0.38 $\pm$ 0.02\\ \midrule
Strong Supervision Values       & \multicolumn{3}{c}{Labeled Maps (\E{video})} & 0.58 $\pm$ 0.05 & 0.45 $\pm$ 0.05 & 0.30 $\pm$ 0.04 & 0.43 $\pm$ 0.02\\
Strong Supervision + VLV (Ours) & \multicolumn{3}{c}{Labeled Maps (\E{video}) + \Vhat{yt}} & 0.50 $\pm$ 0.05 & 0.50 $\pm$ 0.05 & 0.27 $\pm$ 0.04 & 0.41 $\pm$ 0.02\\
\bottomrule
\vspace{-15pt}
\end{tabular}}
\end{table}

%RL (RGB-D)              & 38M (\E{train})                  & Dense  & -          & 0.25 $\pm$ 0.02 & 0.53 $\pm$ 0.02 &                 & \\
%RL (RGB-D)              & 38M (\E{train})                  & Sparse & -          & 0.15 $\pm$ 0.01 & 0.32 $\pm$ 0.02 &                 & \\
% 1. RL(RGB-D ResNet + SimpleCNN, 100K, sparse, 15), SG trains.
% 2. RL(RGB-D SimpleCNN + SimpleCNN, 38M, 100) Finished training
% 3. RL(Depth SimpleCNN, 38M, 15), Already in paper : in
% 4. RL(RGB-D, ResNet + SimpleCNN, 10M, dense, 100), Finish training in 2 hours : in
% 5. RL(RGB ResNet, 20M, dense, 15), Already trained. :in

% rl rgbd sparse
% & 0.28 $\pm$ 0.04 & 0.55 $\pm$ 0.05& 0.16 $\pm$ 0.02 & 0.36 $\pm$ 0.04& 0.06 $\pm$ 0.01 & 0.16 $\pm$ 0.03& 0.00 $\pm$ 0.01 & 0.01 $\pm$ 0.02& 0.15 $\pm$ 0.01 & 0.32 $\pm$ 0.02

% \input{sup-results-stop-diff-sr}
\renewcommand{\arraystretch}{1.1}
\begin{table}[h]
\setlength{\tabcolsep}{8pt}
\centering
\caption{\textbf{Results}: SPL and Success Rate for ObjectGoal wth \textbf{Policy Stop} in novel environments 
\E{test} by object class. Details in \secref{results}.}
%\E{test} by object class. Details in Section 4.2.}
% in oracle stop and policy stop settings. 
\tablelabel{os-stop-class-eval}
\resizebox{\textwidth}{!}{
\begin{tabular}{lccccccccc}
\toprule
\multirow{2}{*}{\textbf{Method}} & \multicolumn{3}{c}{\textbf{Training Supervision}} & \multicolumn{5}{c}{\textbf{SPL}}\\
\cmidrule(lr){2-4} \cmidrule(lr){5-9}
  & \textbf{\# Active Frames} & \textbf{Reward} & \textbf{Other} & \textbf{Bed} & \textbf{Chair} & \textbf{Couch} & \textbf{Dining Table} & \textbf{Toilet} \\
% & Frames & & & & & & \\

%Topological Exploration       & 0.18 $\pm$ 0.04 & 0.17 $\pm$ 0.03 & 0.09 $\pm$ 0.02 & 0.11 $\pm$ 0.03 & 0.09 $\pm$ 0.02\\
%Detection Seeker              & 0.25 $\pm$ 0.04 & 0.25 $\pm$ 0.03 & 0.13 $\pm$ 0.03 & 0.23 $\pm$ 0.05 & 0.14 $\pm$ 0.02\\
%RL (RGB ResNet)        20M    & 0.02 $\pm$ 0.01 & 0.15 $\pm$ 0.03 & 0.05 $\pm$ 0.02 & 0.07 $\pm$ 0.03 & 0.08 $\pm$ 0.02\\
%Behavior Cloning Syn          & 0.07 $\pm$ 0.02 & 0.07 $\pm$ 0.02 & 0.04 $\pm$ 0.02 & 0.14 $\pm$ 0.04 & 0.04 $\pm$ 0.01\\
%Our (Value Learning from Videos) Syn  & 0.20 $\pm$ 0.04 & 0.24 $\pm$ 0.04 & 0.20 $\pm$ 0.04 & 0.23 $\pm$ 0.05 & 0.21 $\pm$ 0.03\\
%Behavior Cloning Real         & 0.06 $\pm$ 0.02 & 0.16 $\pm$ 0.03 & 0.09 $\pm$ 0.02 & 0.14 $\pm$ 0.04 & 0.08 $\pm$ 0.02\\
%Our (Value Learning from Videos) Real & 0.21 $\pm$ 0.04 & 0.25 $\pm$ 0.03 & 0.13 $\pm$ 0.03 & 0.17 $\pm$ 0.04 & 0.24 $\pm$ 0.03\\
%Strong Supervision Values            & 0.24 $\pm$ 0.04 & 0.24 $\pm$ 0.03 & 0.22 $\pm$ 0.04 & 0.28 $\pm$ 0.05 & 0.22 $\pm$ 0.03\\

\midrule
Topological Exploration           & -                                                        & -               & - & 0.18 $\pm$ 0.04 & 0.17 $\pm$ 0.03 & 0.09 $\pm$ 0.02 & 0.11 $\pm$ 0.03 & 0.09 $\pm$ 0.02\\
Detection Seeker                  & -                                                        & -               & - & \textbf{0.25} $\pm$ 0.04 & 0.25 $\pm$ 0.04 & 0.13 $\pm$ 0.03 & \textbf{0.23} $\pm$ 0.05 & 0.14 $\pm$ 0.02\\
RL (RGB ResNet)                   & 20M (\E{train})                                          & Dense           & - & 0.02 $\pm$ 0.01 & 0.15 $\pm$ 0.03 & 0.05 $\pm$ 0.02 & 0.07 $\pm$ 0.03 & 0.08 $\pm$ 0.02\\
Behavior Cloning                  & 40K (\E{train})                                          & -               & \Vhat{yt} & 0.13 $\pm$ 0.03 & 0.12 $\pm$ 0.03 & 0.02 $\pm$ 0.01 & 0.12 $\pm$ 0.04 & 0.05 $\pm$ 0.02\\
Our (Value Learning from Videos)  & 40K (\E{train})                                          & -               & \Vhat{yt} & \textbf{0.25} $\pm$ 0.04 & \textbf{0.28} $\pm$ 0.04 & \textbf{0.22} $\pm$ 0.04 & 0.20 $\pm$ 0.05 & 0.17 $\pm$ 0.03\\ \midrule
Behavior Cloning                  & 40K (\E{train})                                          & -               & \Vhat{syn} & 0.06 $\pm$ 0.02 & 0.16 $\pm$ 0.03 & 0.09 $\pm$ 0.02 & 0.14 $\pm$ 0.04 & 0.08 $\pm$ 0.02\\
Our (Value Learning from Videos)  & 40K (\E{train})                                          & -               & \Vhat{syn} & 0.21 $\pm$ 0.04 & 0.25 $\pm$ 0.03 & 0.13 $\pm$ 0.03 & 0.17 $\pm$ 0.04 & \textbf{0.24} $\pm$ 0.03\\ \midrule
Strong Supervision Values       & \multicolumn{3}{c}{Labeled Maps (\E{video})} & 0.24 $\pm$ 0.04 & 0.24 $\pm$ 0.03 & 0.22 $\pm$ 0.04 & 0.28 $\pm$ 0.05 & 0.22 $\pm$ 0.03\\
Strong Supervision + VLV (Ours) & \multicolumn{3}{c}{Labeled Maps (\E{video}) + \Vhat{yt}} & 0.14 $\pm$ 0.04 & 0.31 $\pm$ 0.04 & 0.18 $\pm$ 0.04 & 0.32 $\pm$ 0.05 & 0.24 $\pm$ 0.03\\
\midrule \midrule
\multirow{2}{*}{\textbf{Method}} & \multicolumn{3}{c}{\textbf{Training
Supervision}} & \multicolumn{5}{c}{\textbf{Success Rate}}\\
\cmidrule(lr){2-4} \cmidrule(lr){5-9}
  & \textbf{\# Active Frames} & \textbf{Reward} & \textbf{Other} & \textbf{Bed} & \textbf{Chair} & \textbf{Couch} & \textbf{Dining Table} & \textbf{Toilet} \\
% & Frames & & & & & & \\
%Topological Exploration       & 0.27 $\pm$ 0.05 & 0.35 $\pm$ 0.05 & 0.26 $\pm$ 0.05 & 0.29 $\pm$ 0.07 & 0.27 $\pm$ 0.04\\
%Detection Seeker              & 0.38 $\pm$ 0.05 & 0.48 $\pm$ 0.05 & 0.23 $\pm$ 0.05 & 0.37 $\pm$ 0.07 & 0.36 $\pm$ 0.04\\
%RL (RGB ResNet)        20M    & 0.07 $\pm$ 0.03 & 0.38 $\pm$ 0.05 & 0.12 $\pm$ 0.04 & 0.15 $\pm$ 0.05 & 0.23 $\pm$ 0.04\\
%Behavior Cloning Syn          & 0.12 $\pm$ 0.04 & 0.16 $\pm$ 0.04 & 0.14 $\pm$ 0.04 & 0.37 $\pm$ 0.07 & 0.10 $\pm$ 0.03\\
%Our (Values from Videos) Syn  & 0.32 $\pm$ 0.05 & 0.44 $\pm$ 0.05 & 0.33 $\pm$ 0.06 & 0.45 $\pm$ 0.07 & 0.41 $\pm$ 0.05\\
%Behavior Cloning Real         & 0.11 $\pm$ 0.04 & 0.43 $\pm$ 0.05 & 0.23 $\pm$ 0.05 & 0.34 $\pm$ 0.07 & 0.20 $\pm$ 0.04\\
%Our (Values from Videos) Real & 0.30 $\pm$ 0.05 & 0.49 $\pm$ 0.05 & 0.26 $\pm$ 0.05 & 0.34 $\pm$ 0.07 & 0.43 $\pm$ 0.05\\
%Strong Supervision Values            & 0.34 $\pm$ 0.05 & 0.50 $\pm$ 0.05 & 0.37 $\pm$ 0.06 & 0.54 $\pm$ 0.07 & 0.42 $\pm$ 0.05\\
\midrule
Topological Exploration           & -                                                        & -               & - & 0.27 $\pm$ 0.05 & 0.35 $\pm$ 0.05 & 0.26 $\pm$ 0.05 & 0.29 $\pm$ 0.07 & 0.27 $\pm$ 0.04\\
Detection Seeker                  & -                                                        & -               & - & 0.38 $\pm$ 0.06 & 0.48 $\pm$ 0.05 & 0.23 $\pm$ 0.05 & \textbf{0.37} $\pm$ 0.07 & 0.36 $\pm$ 0.05\\
RL (RGB ResNet)                   & 20M (\E{train})                                          & Dense           & - & 0.07 $\pm$ 0.03 & 0.38 $\pm$ 0.05 & 0.12 $\pm$ 0.04 & 0.15 $\pm$ 0.05 & 0.23 $\pm$ 0.04\\
Behavior Cloning                  & 40K (\E{train})                                          & -               & \Vhat{yt} & 0.23 $\pm$ 0.05 & 0.30 $\pm$ 0.05 & 0.06 $\pm$ 0.03 & 0.33 $\pm$ 0.07 & 0.14 $\pm$ 0.03\\
Our (Value Learning from Videos)  & 40K (\E{train})                                          & -               & \Vhat{yt} & \textbf{0.40} $\pm$ 0.06 & \textbf{0.52} $\pm$ 0.05 & \textbf{0.36} $\pm$ 0.06 & 0.31 $\pm$ 0.07 & 0.32 $\pm$ 0.04\\ \midrule
Behavior Cloning                  & 40K (\E{train})                                          & -               & \Vhat{syn} & 0.11 $\pm$ 0.04 & 0.43 $\pm$ 0.05 & 0.24 $\pm$ 0.05 & 0.34 $\pm$ 0.07 & 0.20 $\pm$ 0.04\\
Our (Value Learning from Videos)  & 40K (\E{train})                                          & -               & \Vhat{syn} & 0.30 $\pm$ 0.05 & 0.49 $\pm$ 0.05 & 0.26 $\pm$ 0.05 & 0.34 $\pm$ 0.07 & \textbf{0.43} $\pm$ 0.05\\ \midrule
Strong Supervision Values       & \multicolumn{3}{c}{Labeled Maps (\E{video})} & 0.34 $\pm$ 0.06 & 0.50 $\pm$ 0.05 & 0.37 $\pm$ 0.05 & 0.54 $\pm$ 0.07 & 0.42 $\pm$ 0.05\\
Strong Supervision + VLV (Ours) & \multicolumn{3}{c}{Labeled Maps (\E{video}) + \Vhat{yt}} & 0.20 $\pm$ 0.05 & 0.53 $\pm$ 0.05 & 0.30 $\pm$ 0.05 & 0.52 $\pm$ 0.08 & 0.47 $\pm$ 0.05\\
\bottomrule                                                                                                                                                                  
\vspace{-15pt}                                                                                                                                                               
\end{tabular}}                                                                                                                                                               
\end{table}                                                                                                                                                                  
                                                                                                                                                                             
%RL (RGB-D)              & 38M (\E{train})                  & Dense  & -          & 0.25 $\pm$ 0.02 & 0.53 $\pm$ 0.02 &                 & \\                                 
%RL (RGB-D)              & 38M (\E{train})                  & Sparse & -          & 0.15 $\pm$ 0.01 & 0.32 $\pm$ 0.02 &                 & \\
% 1. RL(RGB-D ResNet + SimpleCNN, 100K, sparse, 15), SG trains.
% 2. RL(RGB-D SimpleCNN + SimpleCNN, 38M, 100) Finished training
% 3. RL(Depth SimpleCNN, 38M, 15), Already in paper : in
% 4. RL(RGB-D, ResNet + SimpleCNN, 10M, dense, 100), Finish training in 2 hours : in
% 5. RL(RGB ResNet, 20M, dense, 15), Already trained. :in

% rl rgbd sparse
% & 0.28 $\pm$ 0.04 & 0.55 $\pm$ 0.05& 0.16 $\pm$ 0.02 & 0.36 $\pm$ 0.04& 0.06 $\pm$ 0.01 & 0.16 $\pm$ 0.03& 0.00 $\pm$ 0.01 & 0.01 $\pm$ 0.02& 0.15 $\pm$ 0.01 & 0.32 $\pm$ 0.02

% \input{sup-results-stop-class-sr}
%\input{sup-results-diff-sr}
% \clearpage
% \subsection{Ablations (corresponding to Section 4.2)}
%\input{sup-data-teleport}
\begin{table}[h]
\setlength{\tabcolsep}{7pt}
\centering
\caption{We report various ablations of our method, when using automatic
stopping behavior, evaluated on \E{train}. Base setting uses noisy trajectores,
action labels from inverse models and panorama images. We ablate these
settings. See \secref{ablations} for details.}
%settings. See Section 4.3 for details.}
\tablelabel{ablations}
\resizebox{\textwidth}{!}{
\begin{tabular}{lccccccccc}
\toprule
& \multicolumn{4}{c}{SPL} & \multicolumn{4}{c}{Success Rate}\\
\cmidrule(lr){2-5}
\cmidrule(lr){6-9}
Method & Easy & Medium & Hard & Overall & Easy & Medium & Hard & Overall\\ 
\midrule
Base setting      & 0.62$\pm$  0.04 & 0.42$\pm$  0.04 & 0.23$\pm$  0.03 & 0.40$\pm$  0.02 & 0.95$\pm$  0.03 & 0.86$\pm$  0.05 & 0.56$\pm$  0.05 & 0.75$\pm$  0.03\\
True actions                & 0.61$\pm$  0.05 & 0.45$\pm$  0.05 & 0.25$\pm$  0.03 & 0.41$\pm$  0.03 & 0.94$\pm$  0.03 & 0.86$\pm$  0.05 & 0.51$\pm$  0.05 & 0.73$\pm$  0.03\\
True detections             & 0.62$\pm$  0.05 & 0.45$\pm$  0.05 & 0.22$\pm$  0.03 & 0.40$\pm$  0.03 & 0.95$\pm$  0.03 & 0.86$\pm$  0.05 & 0.48$\pm$  0.05 & 0.72$\pm$ 0.03\\
True rewards                & 0.64$\pm$  0.05 & 0.46$\pm$  0.05 & 0.21$\pm$  0.03 & 0.41$\pm$  0.03 & 0.95$\pm$  0.03 & 0.86$\pm$  0.05 & 0.48$\pm$  0.05 & 0.72$\pm$  0.03\\
No noise in videos          & 0.65$\pm$  0.05 & 0.46$\pm$  0.04 & 0.25$\pm$  0.03 & 0.43$\pm$  0.03 & 0.95$\pm$  0.03 & 0.92$\pm$  0.04 & 0.59$\pm$  0.05 & 0.78$\pm$  0.03\\
\D{coco} score              & 0.73$\pm$  0.04 & 0.48$\pm$  0.05 & 0.26$\pm$  0.03 & 0.46$\pm$  0.03 & 0.98$\pm$  0.02 & 0.88$\pm$  0.05 & 0.58$\pm$  0.06 & 0.78$\pm$  0.03 \\
Train on $360^\circ$ videos & 0.66$\pm$  0.04 & 0.51$\pm$  0.05 & 0.32$\pm$  0.03 & 0.47$\pm$  0.02 & 0.98$\pm$  0.02 & 0.92$\pm$  0.04 & 0.66$\pm$  0.05 & 0.82$\pm$  0.03\\
\bottomrule
\end{tabular}}
\end{table}

\clearpage

\section{Visualizations}
%\subsection{Value Maps over the Course of Training}
%\input{sup-train-maps}
\subsection{Value Predictions on Panorama}
\begin{figure}[h]
    \centering
    \insertW{0.86}{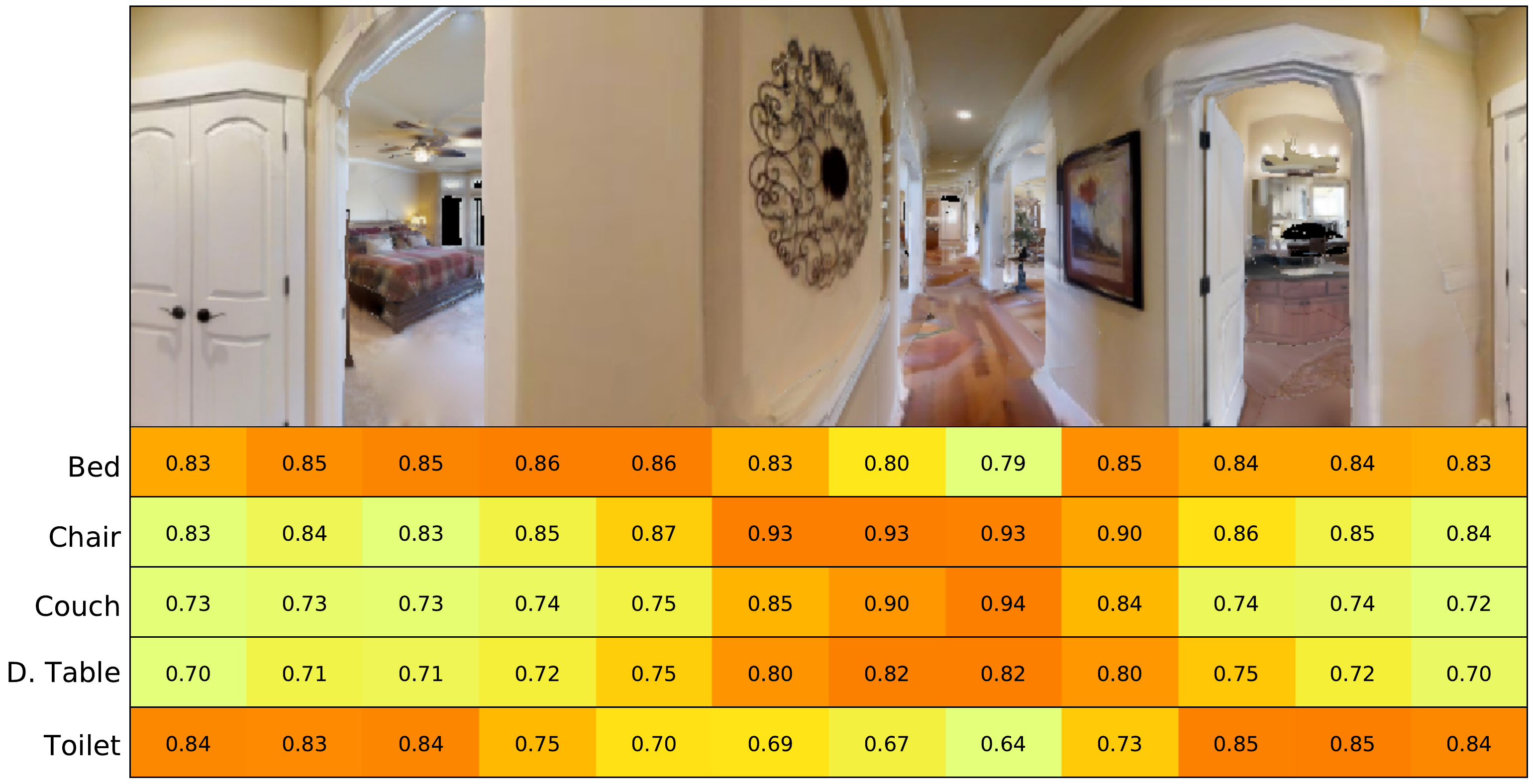}\\
    \insertW{0.86}{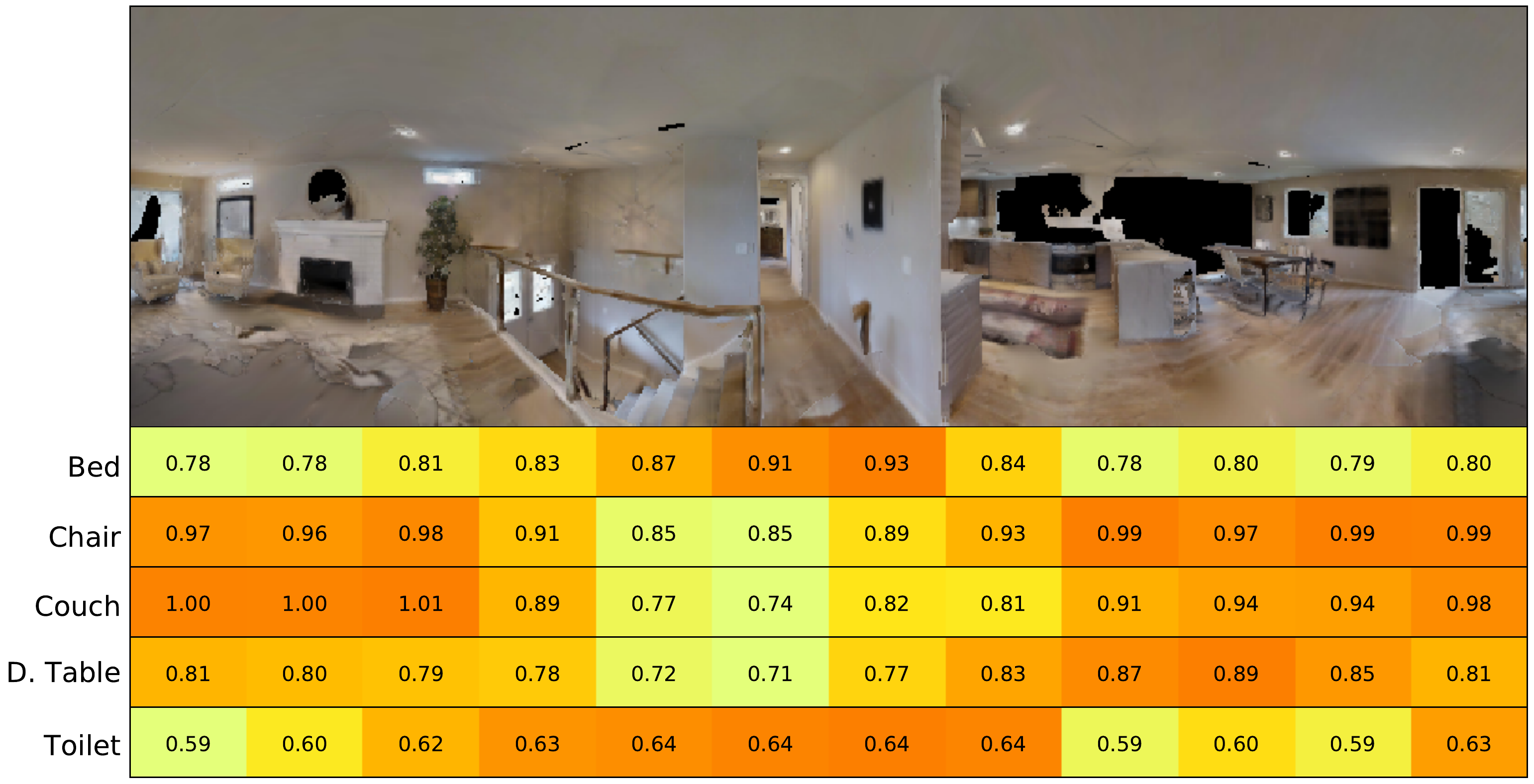}\\
    \insertW{0.86}{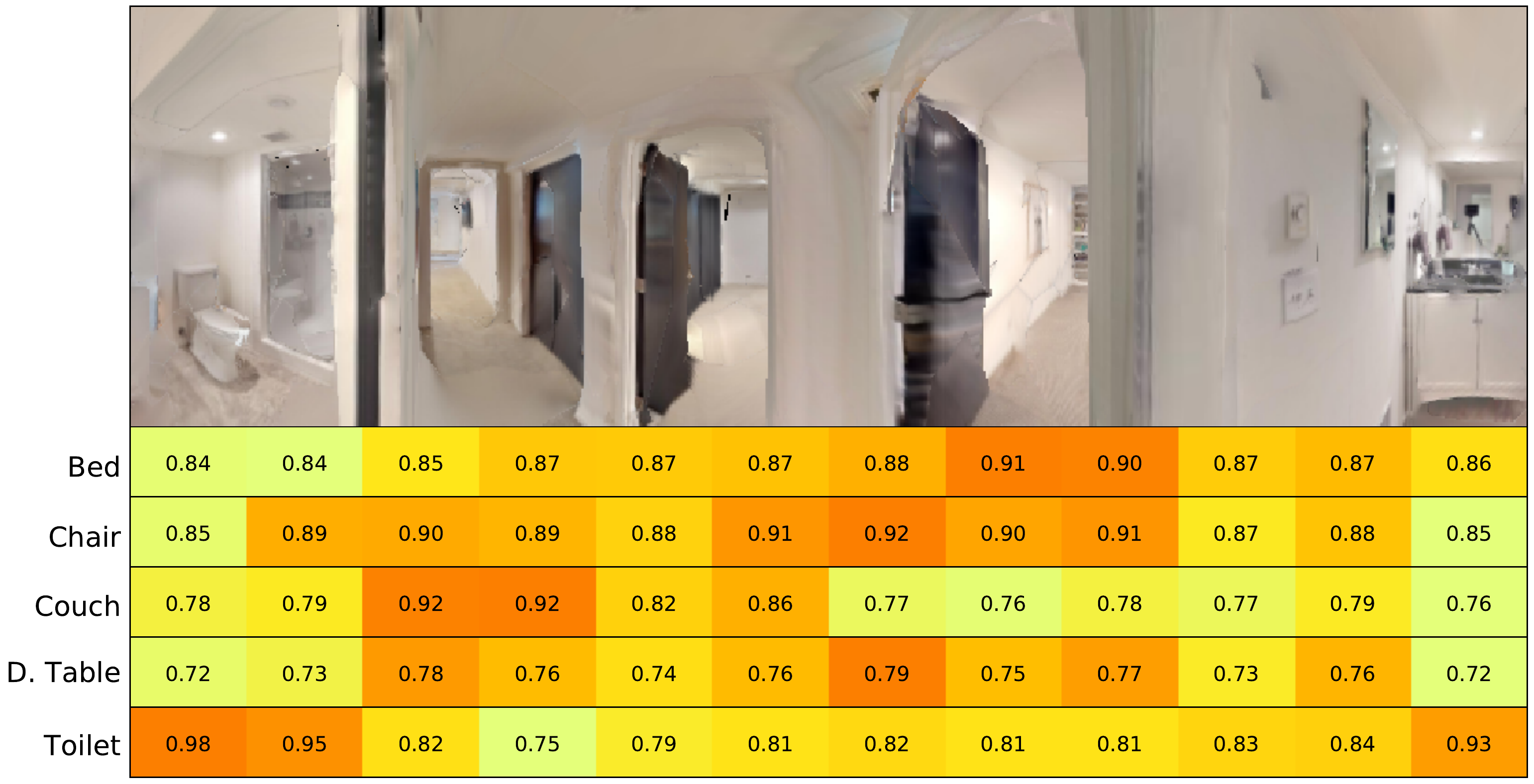}\\
    \caption{Example panoramas from novel environments with scores from our
    value network. Scores for each object class (Bed, Chair, Couch, Dining
    Table, and Toilet) are reported. We can see that value is high in the
    likely direction of objects even if the object is not directly visible.}
    \figlabel{pano-1}
  \end{figure}

\begin{figure}[h]
    \centering
    \insertW{0.86}{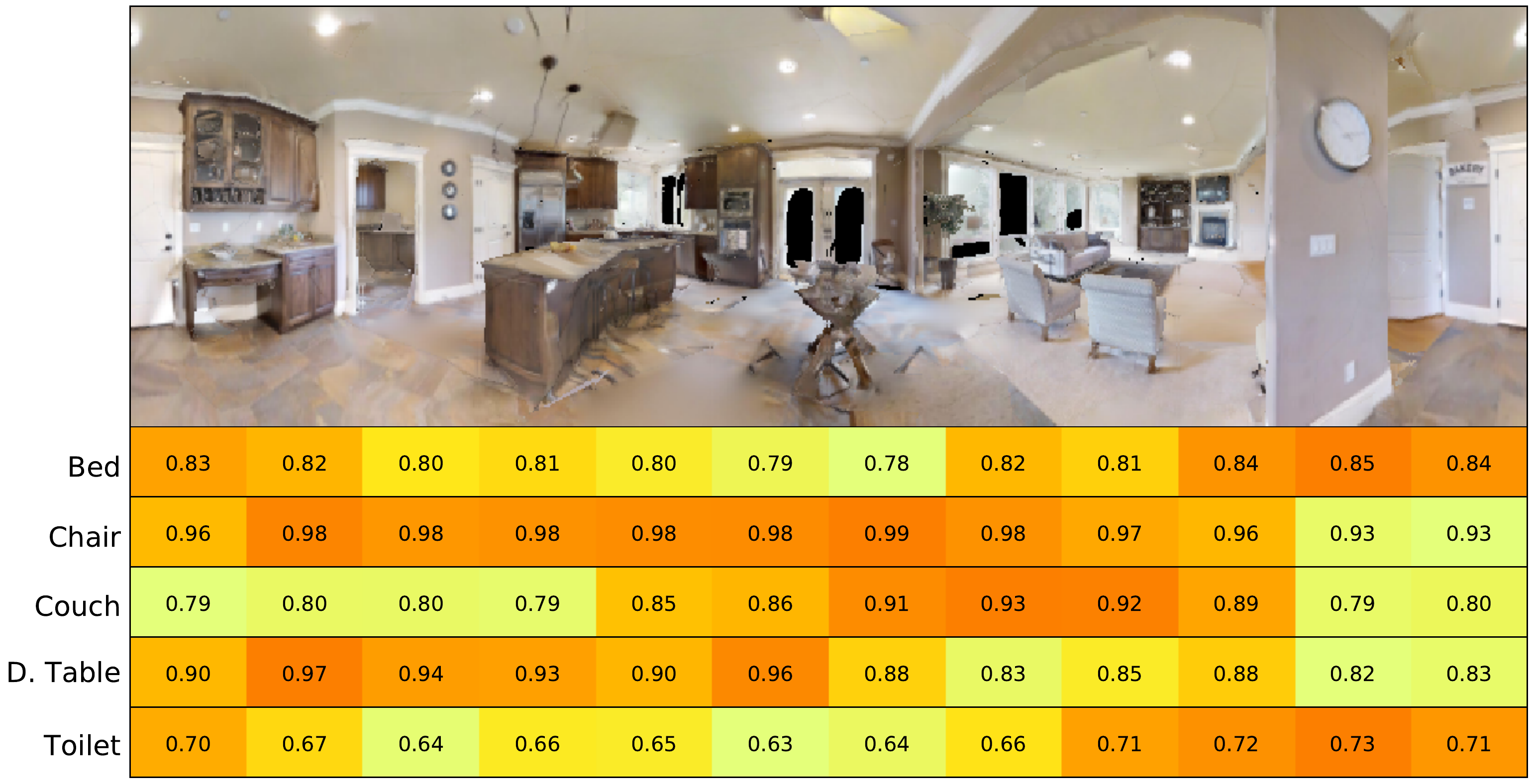}\\
    \insertW{0.86}{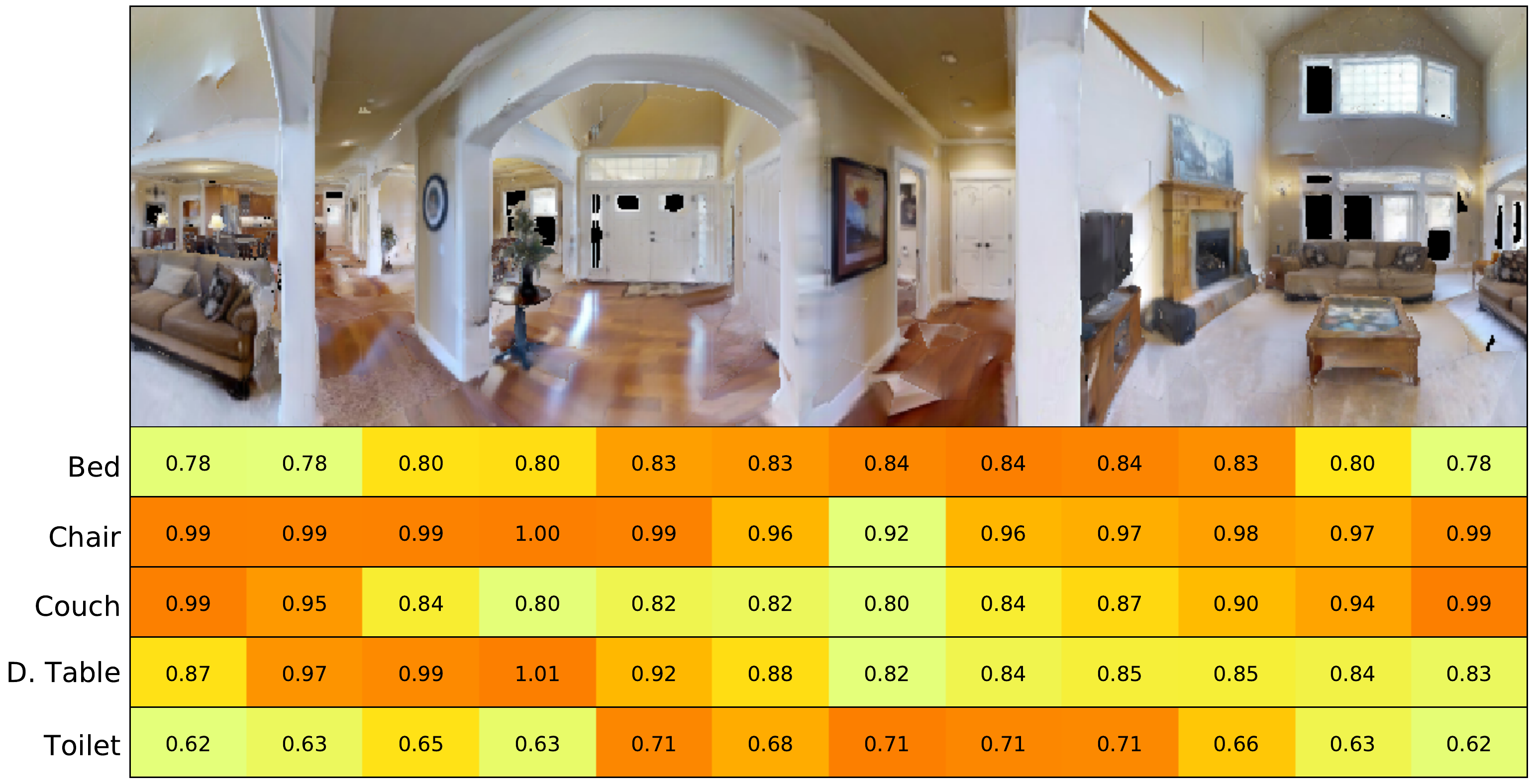}\\
    \insertW{0.86}{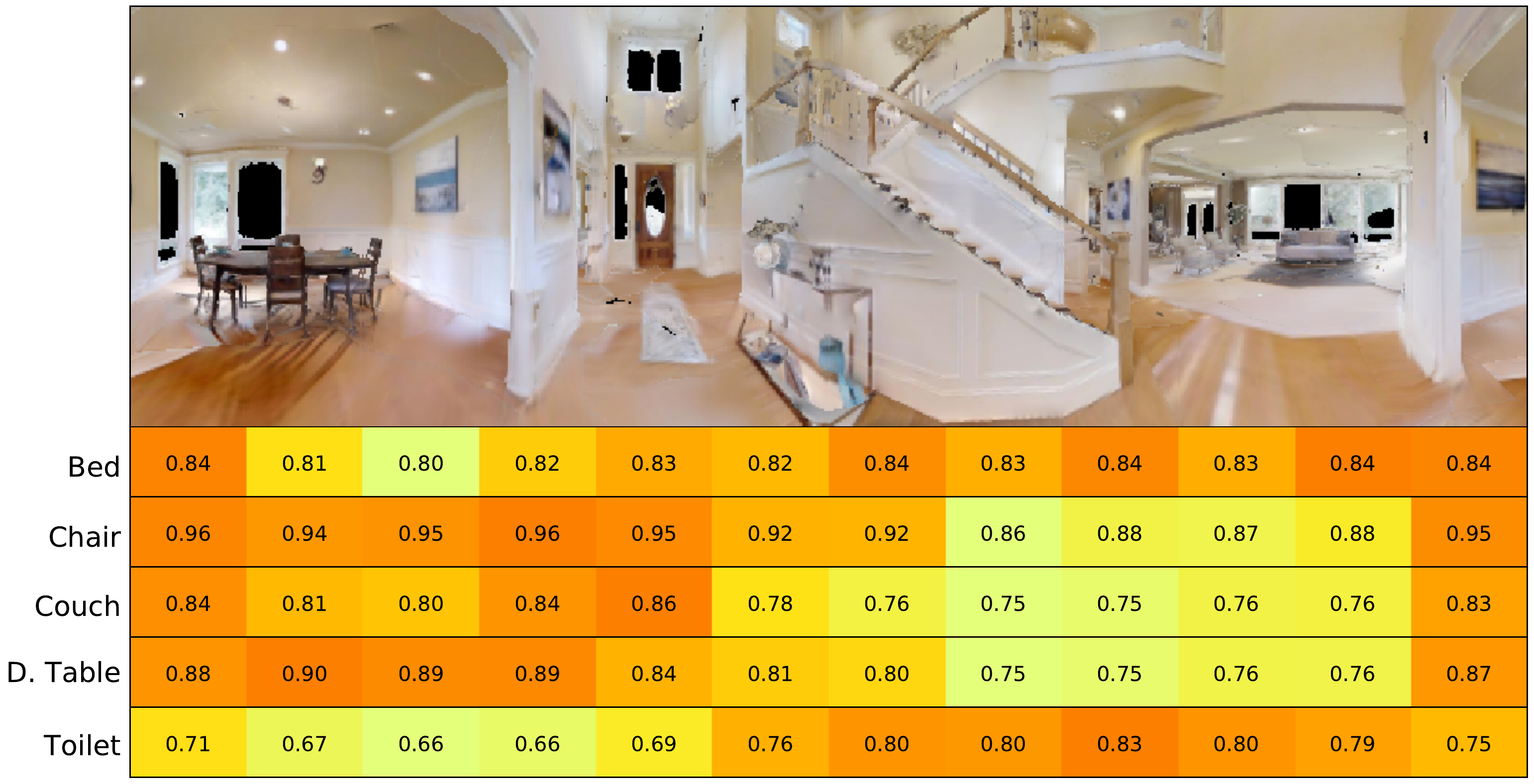}\\
    \caption{Example panoramas from novel environments with scores from our
    value network. Scores for each object class (Bed, Chair, Couch, Dining
    Table, and Toilet) are reported. We can see that value is high in the
    likely direction of objects even if the object is not directly visible.}
    \figlabel{pano-2}
  \end{figure}

\clearpage

\subsection{Executed Trajectories}
\begin{figure}[h]
  % \centering
    % \begin{tabular}{l}
      \insertW{0.32}{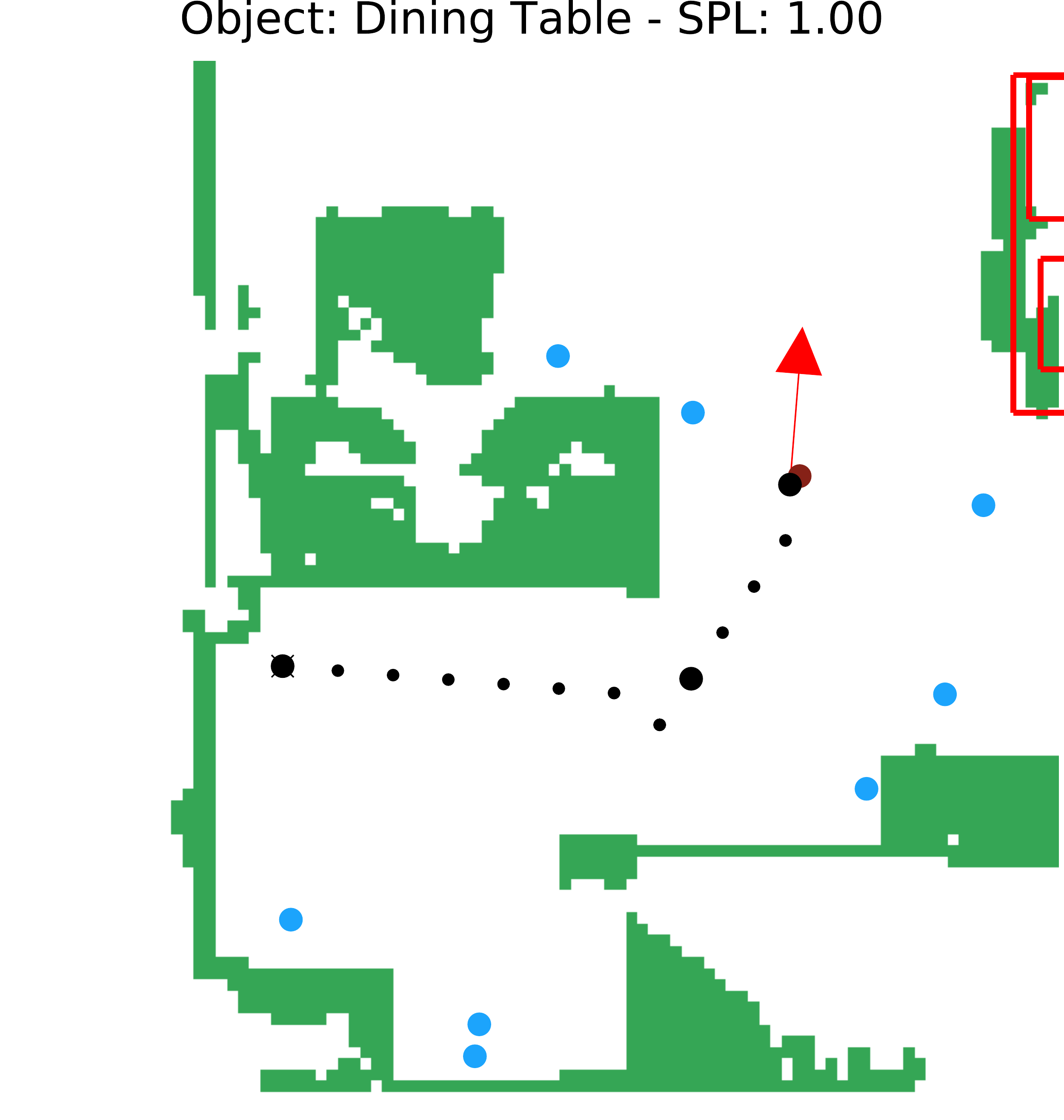}
      \hfill
      \insertW{0.32}{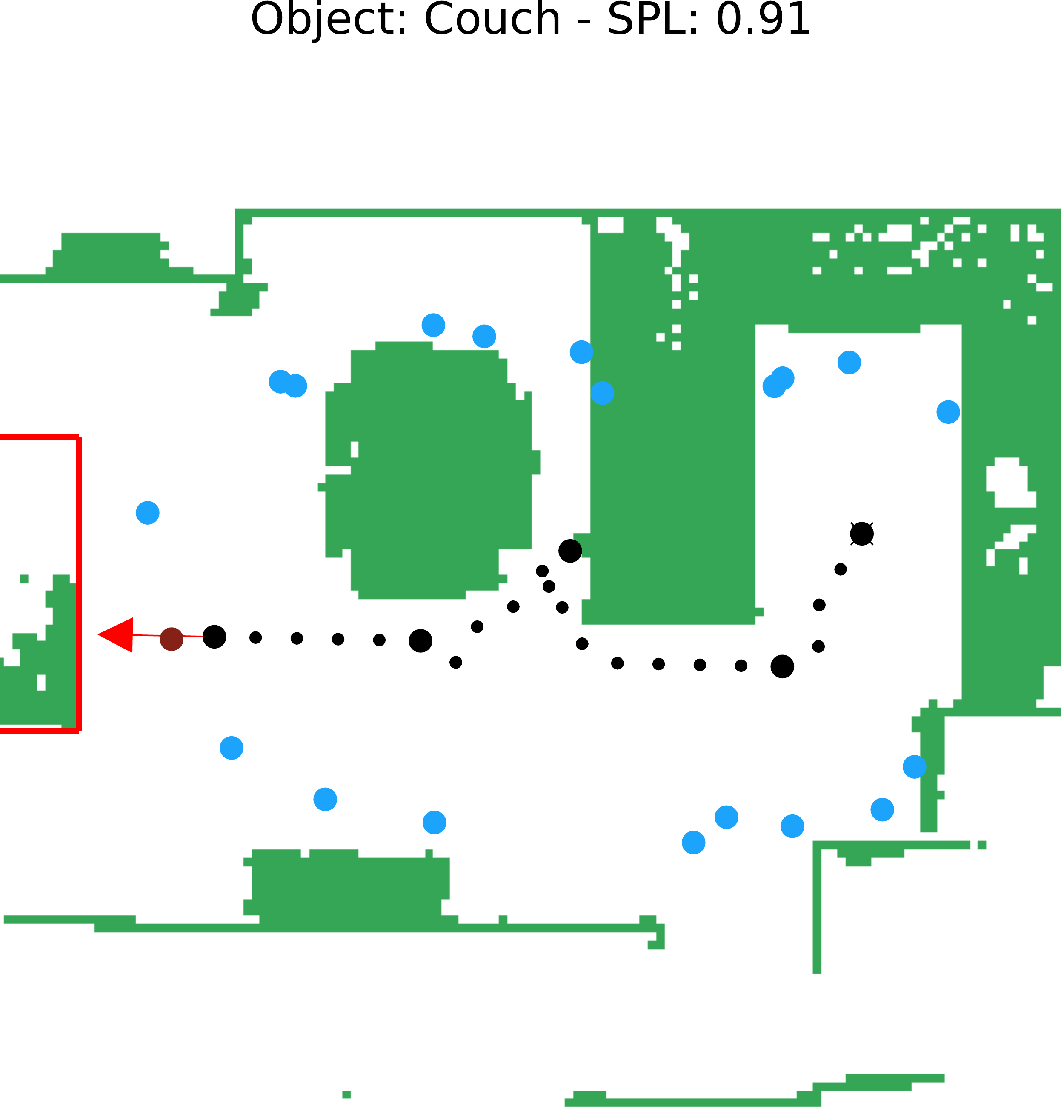}
      \hfill
      \insertW{0.32}{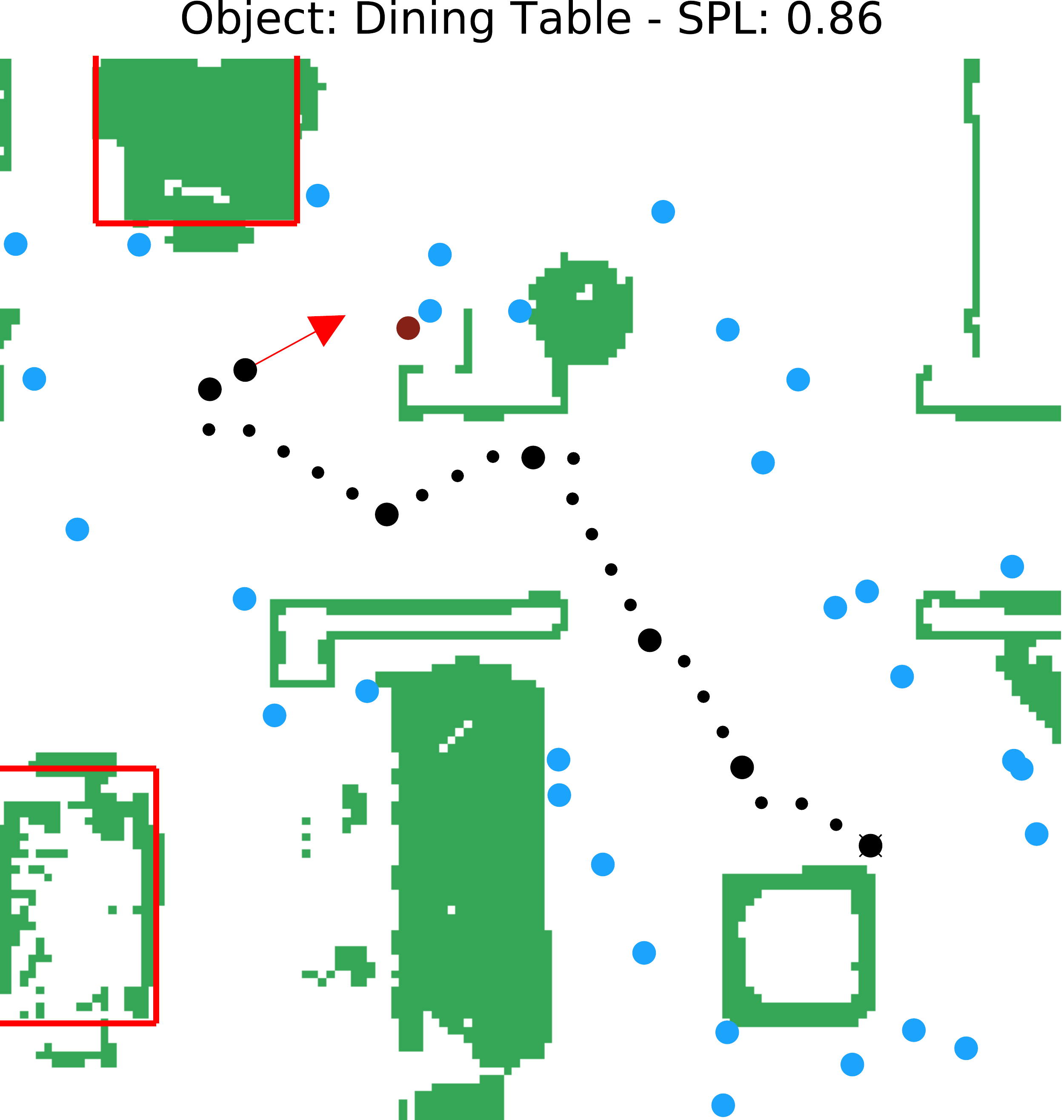}
      \\ \\
      \insertW{0.32}{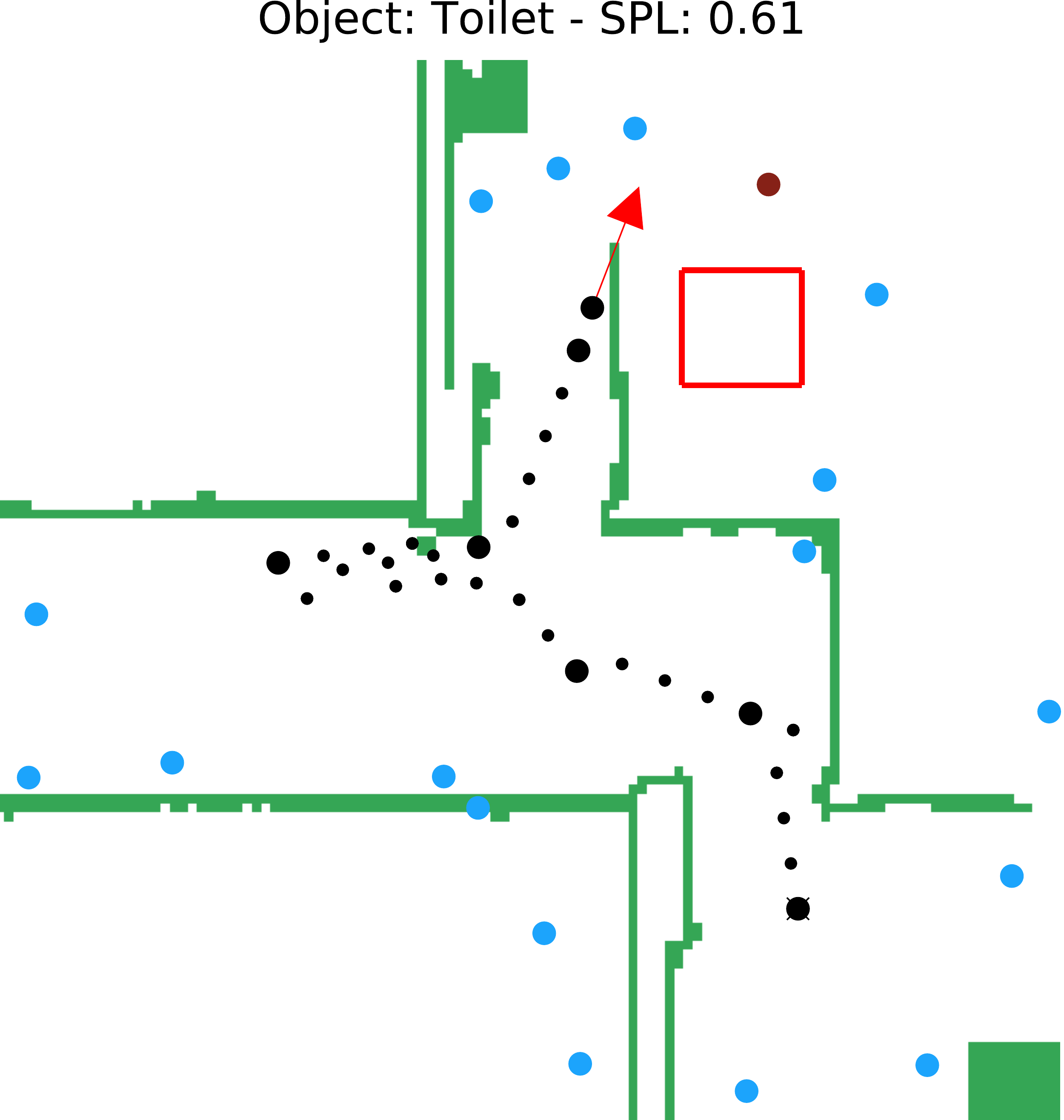}
      \hfill
      \insertW{0.32}{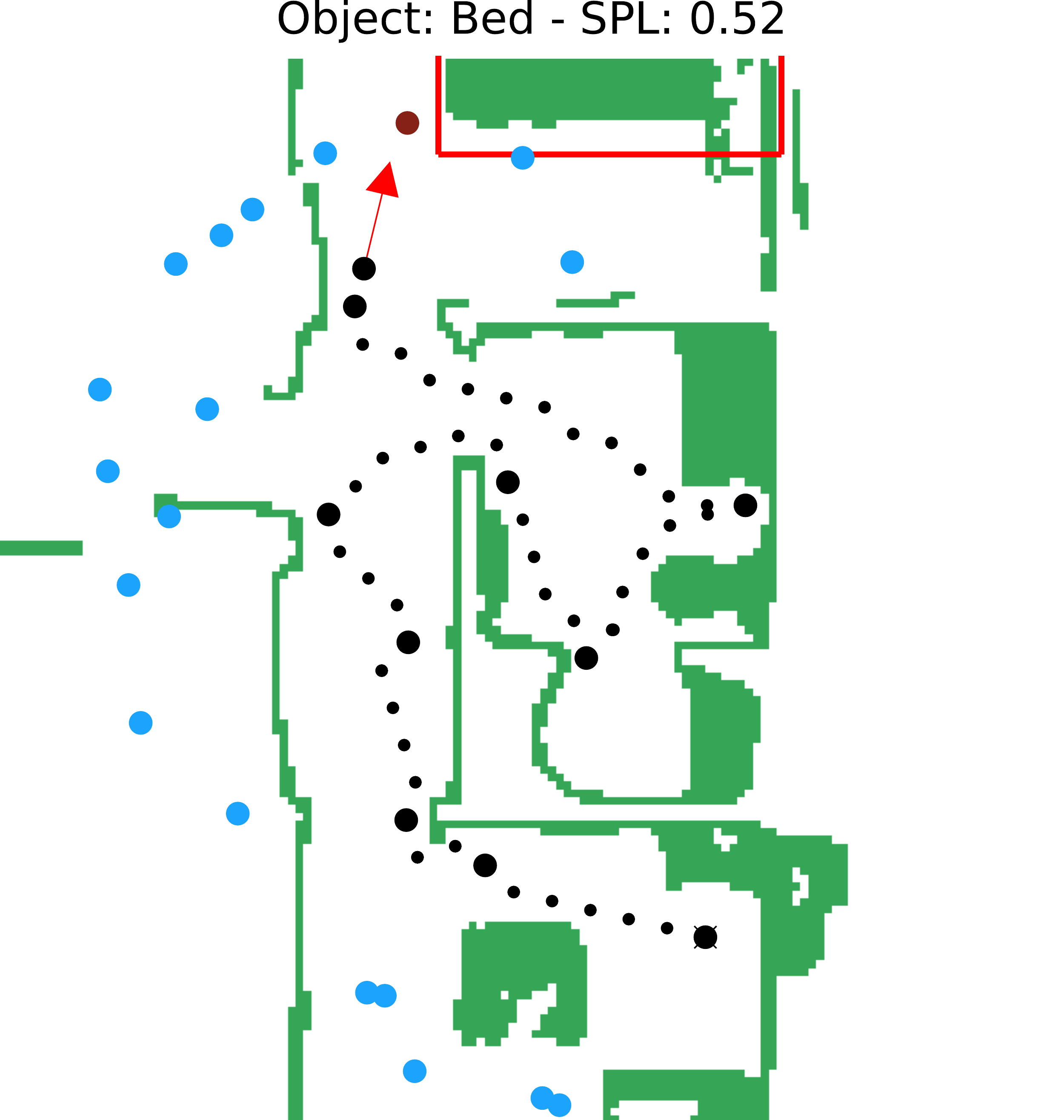}
      \hfill
      \insertW{0.32}{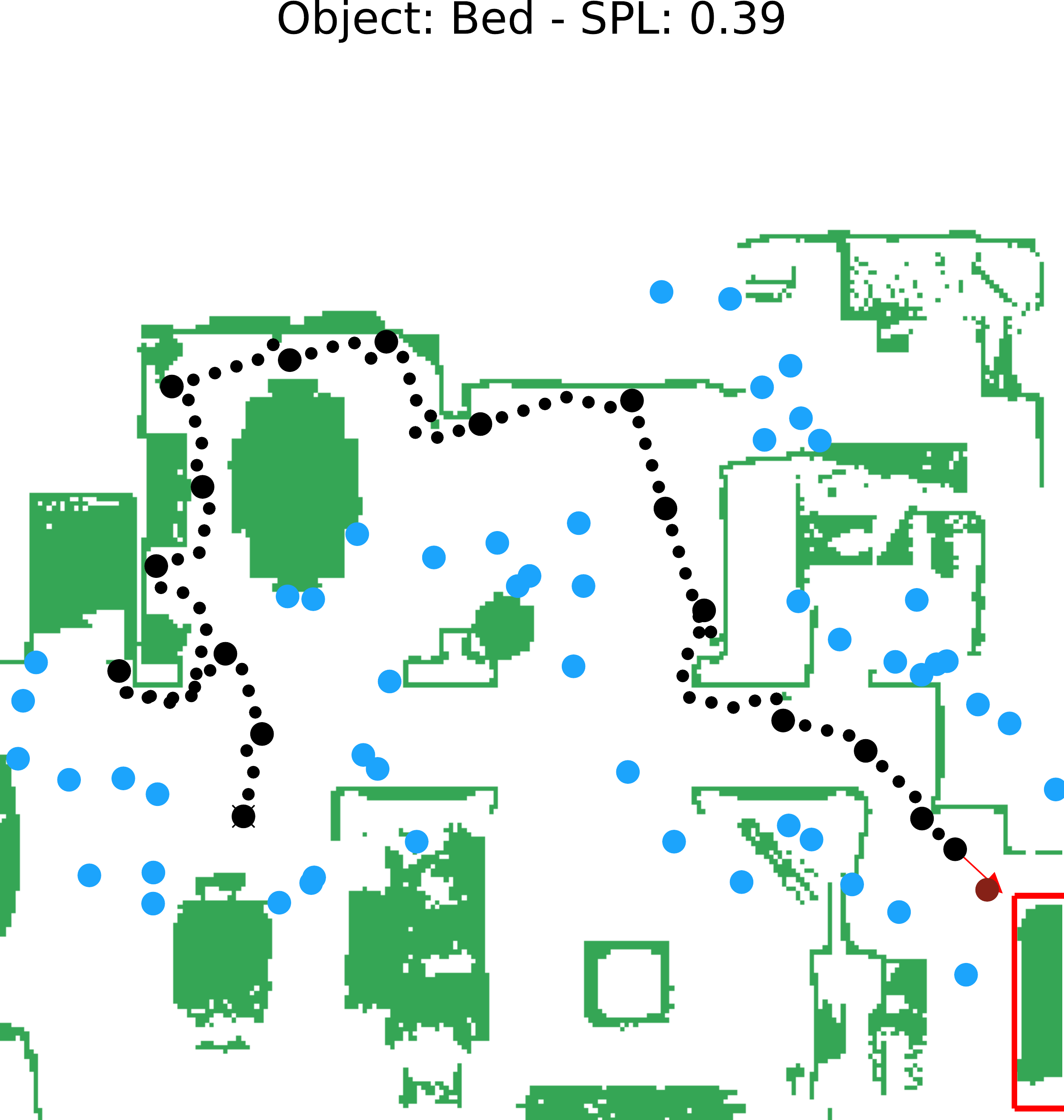}
      \\ \\
      \insertW{0.32}{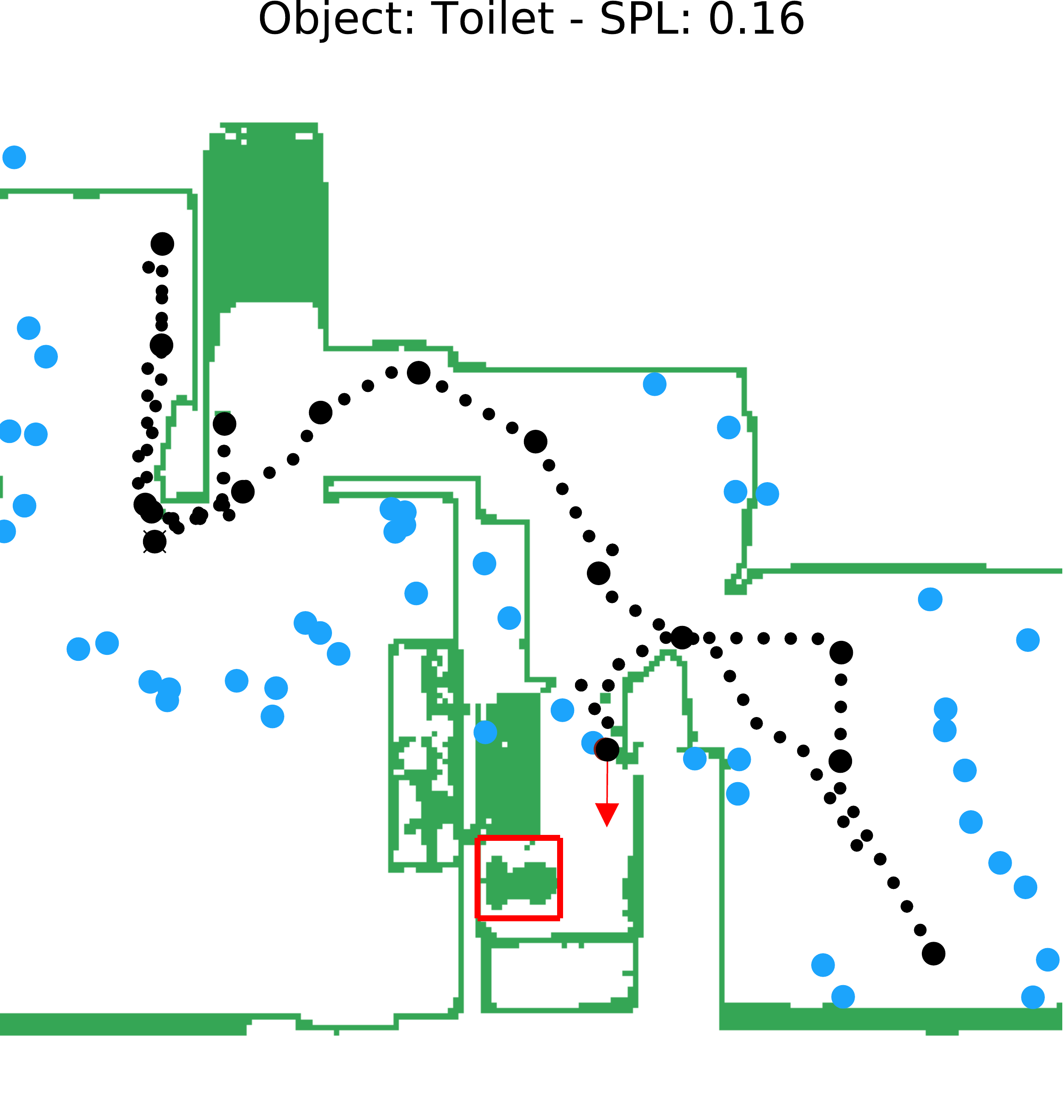}
      \hfill
      \insertW{0.32}{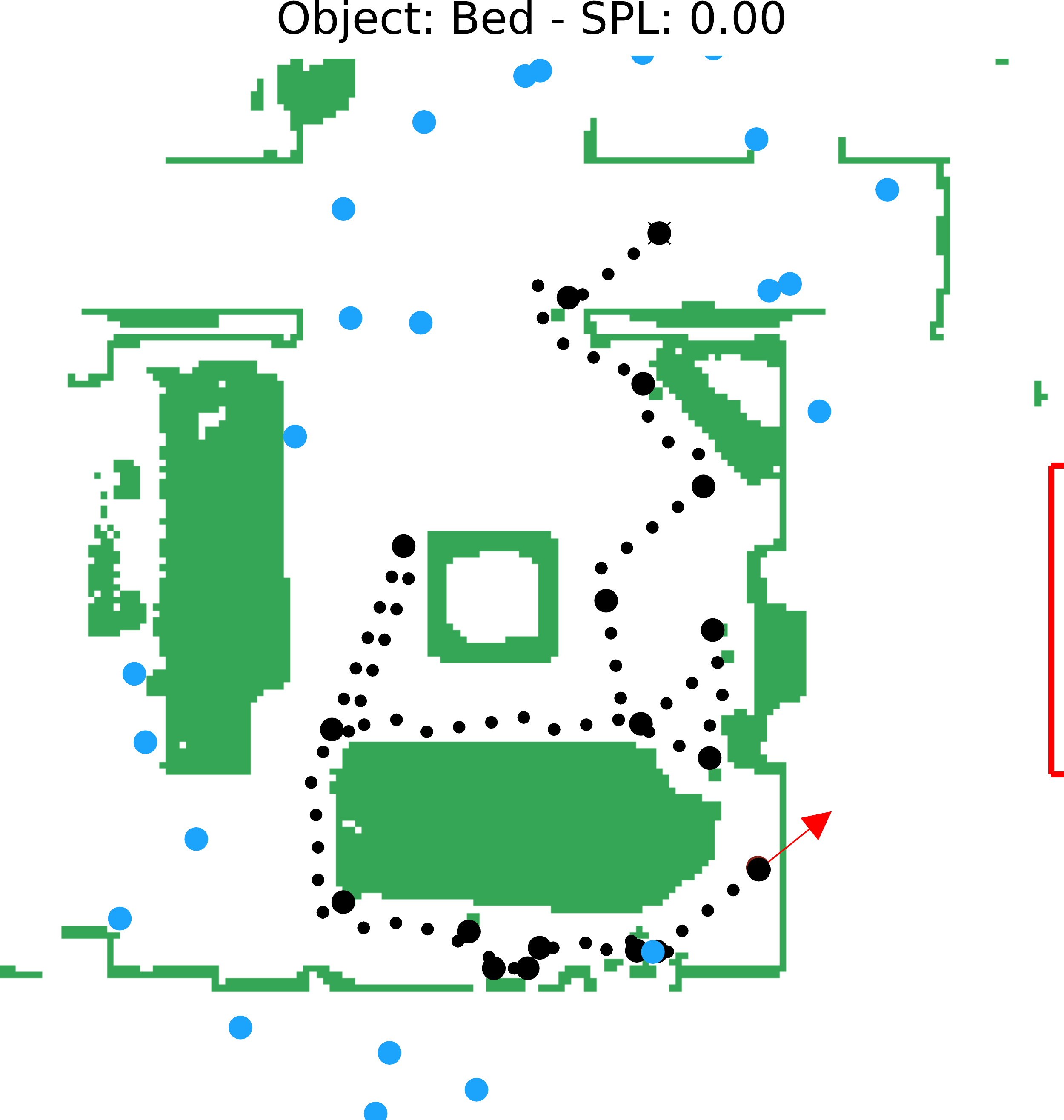}
      \hfill
      \insertW{0.32}{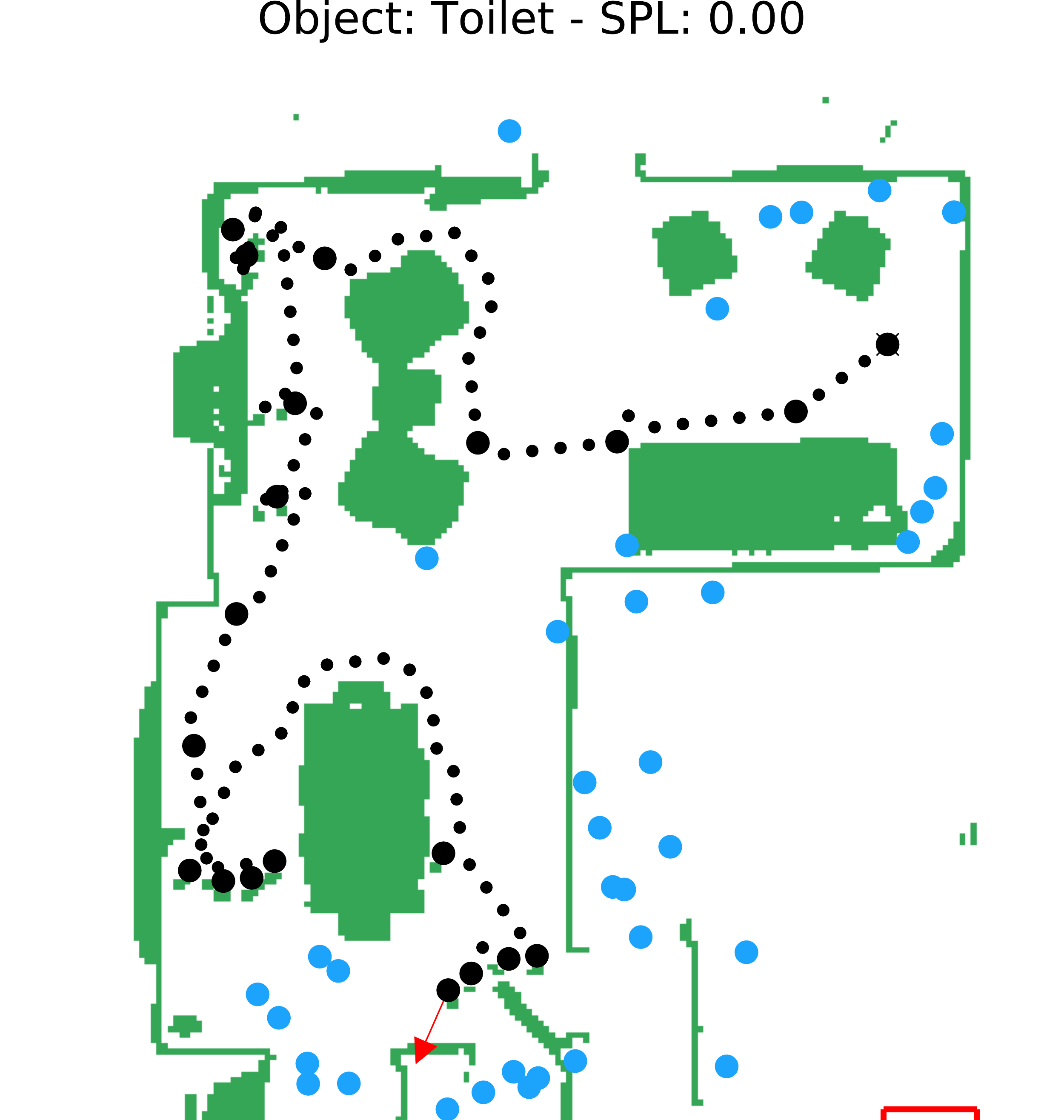}
    %\end{tabular}
    \caption{Example trajectores from our method navigating in novel
    environments, sorted by SPL (first few show successes, last few show
    failures). The black path indicates the trajectory taken by the agent. A
    blue circle indicates potential short-term goal, and a red rectangle
    indicates the object goal.}
    \figlabel{traj-our}
    \end{figure}

\clearpage

\subsection{\E{test} Problem Setup Visualization}
\begin{figure}[h]
  \centering
    \insertW{1.0}{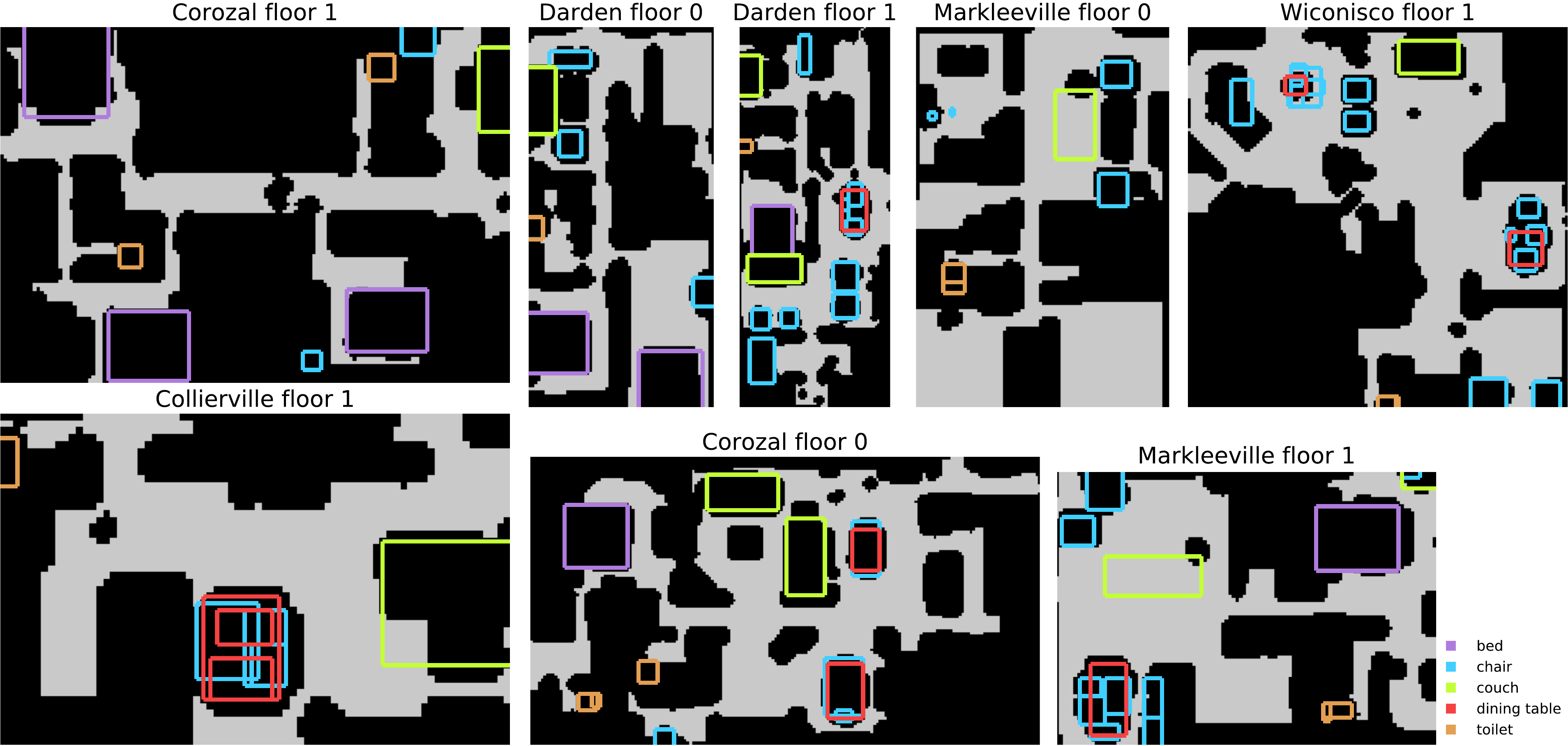}
    % \begin{tabular}{l}
    %   \insertW{0.22}{supplementary_figures/topdown_maps/colors.pdf}
    %   \insertW{0.74}{supplementary_figures/topdown_maps/Corozal0_all.pdf}\\
    %   \insertW{0.15}{supplementary_figures/topdown_maps/Darden0_all.pdf}
    %   \insertW{0.15}{supplementary_figures/topdown_maps/Darden1_all.pdf}
    %   \insertW{0.32}{supplementary_figures/topdown_maps/Markleeville0_all.pdf}
    %   \insertW{0.32}{supplementary_figures/topdown_maps/Wiconisco1_all.pdf}\\
    %   \insertW{0.32}{supplementary_figures/topdown_maps/Markleeville1_all.pdf}
    %   \insertW{0.64}{supplementary_figures/topdown_maps/Corozal1_all.pdf}\\
    % \end{tabular}
    \caption{Top-down maps of selected floors from the \E{test} environments.
    We also show ground truth object locations. Agent does not have access to
    any of these maps or ground truth object locations. Visualizations here are
    provided only to show the difficulty and realism of our problem setup.}
    \figlabel{e-test-top-maps}
    \end{figure}

\clearpage

\subsection{Predicted Values on Held-out Environments}
%\begin{figure}[t]
    %\insertW{0.40}{supplementary_figures/maps_ours/Collierville0bed.pdf}
    %\insertW{0.40}{supplementary_figures/maps_value/Collierville0bed.pdf}
    %\insertHW{0.26}{0.04}{supplementary_figures/maps_ours/colorbar.png} \\
    %\insertW{0.40}{supplementary_figures/maps_ours/Collierville0chair.pdf}
    %\insertW{0.40}{supplementary_figures/maps_value/Collierville0chair.pdf}\\
    %\insertW{0.40}{supplementary_figures/maps_ours/Collierville0couch.pdf}
    %\insertW{0.40}{supplementary_figures/maps_value/Collierville0couch.pdf} \\
    %\insertW{0.40}{supplementary_figures/maps_ours/Collierville0dining_table.pdf}
    %\insertW{0.40}{supplementary_figures/maps_value/Collierville0dining_table.pdf} \\
    %\insertW{0.40}{supplementary_figures/maps_ours/Collierville0toilet.pdf}
    %\insertW{0.40}{supplementary_figures/maps_value/Collierville0toilet.pdf}
    %\caption{Maps representing the value of different locations as predicted by our method and the direct value learning methods. The map is floor 1 from the Gibson environment `Collierville', which was in the validation split.}
    %\figlabel{Collierville0}
    %\end{figure}
\begin{figure}[h]
  % \centering
    % \begin{tabular}{l}
      \insertH{0.28}{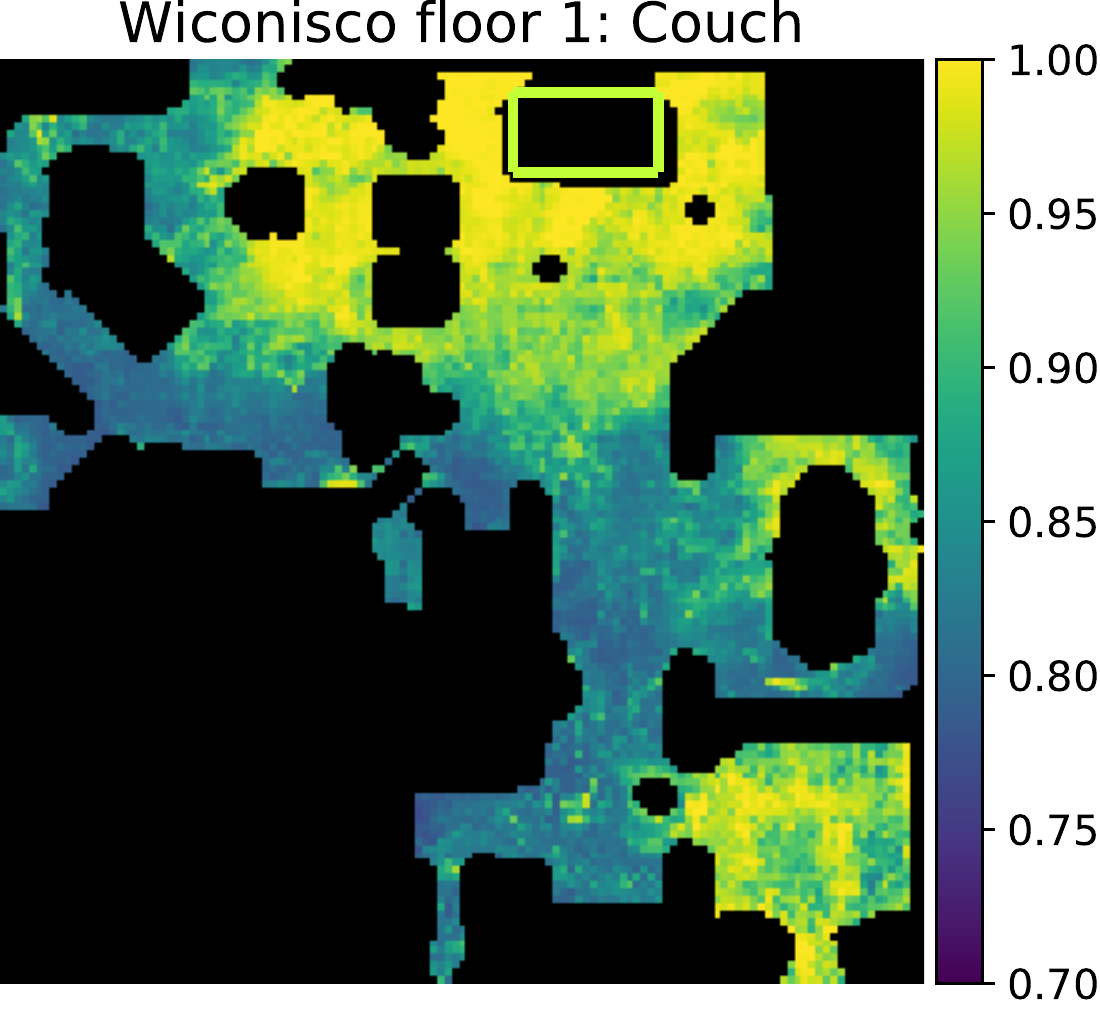}
      \insertH{0.28}{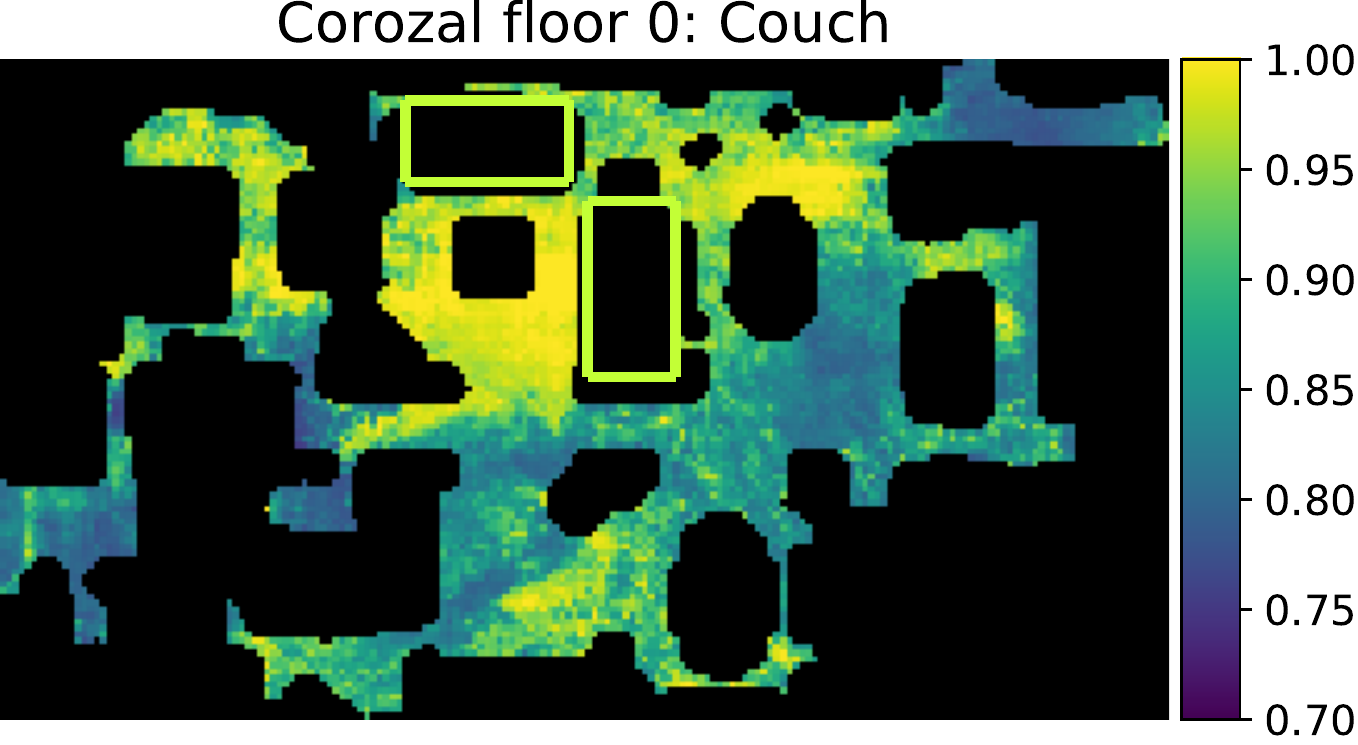}
      \insertH{0.12}{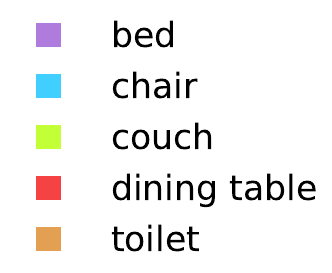}\\
      \insertH{0.30}{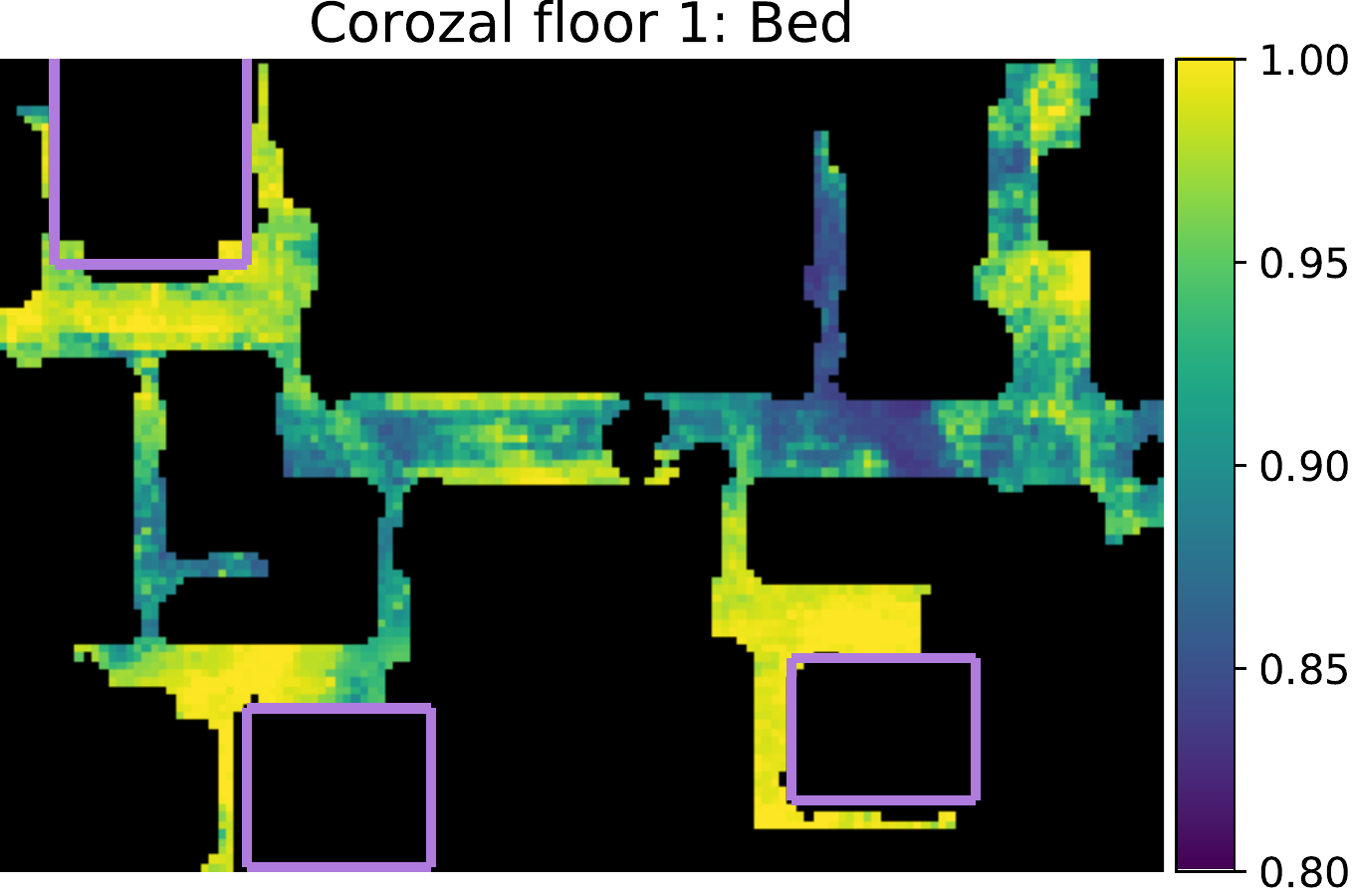}
      \insertH{0.30}{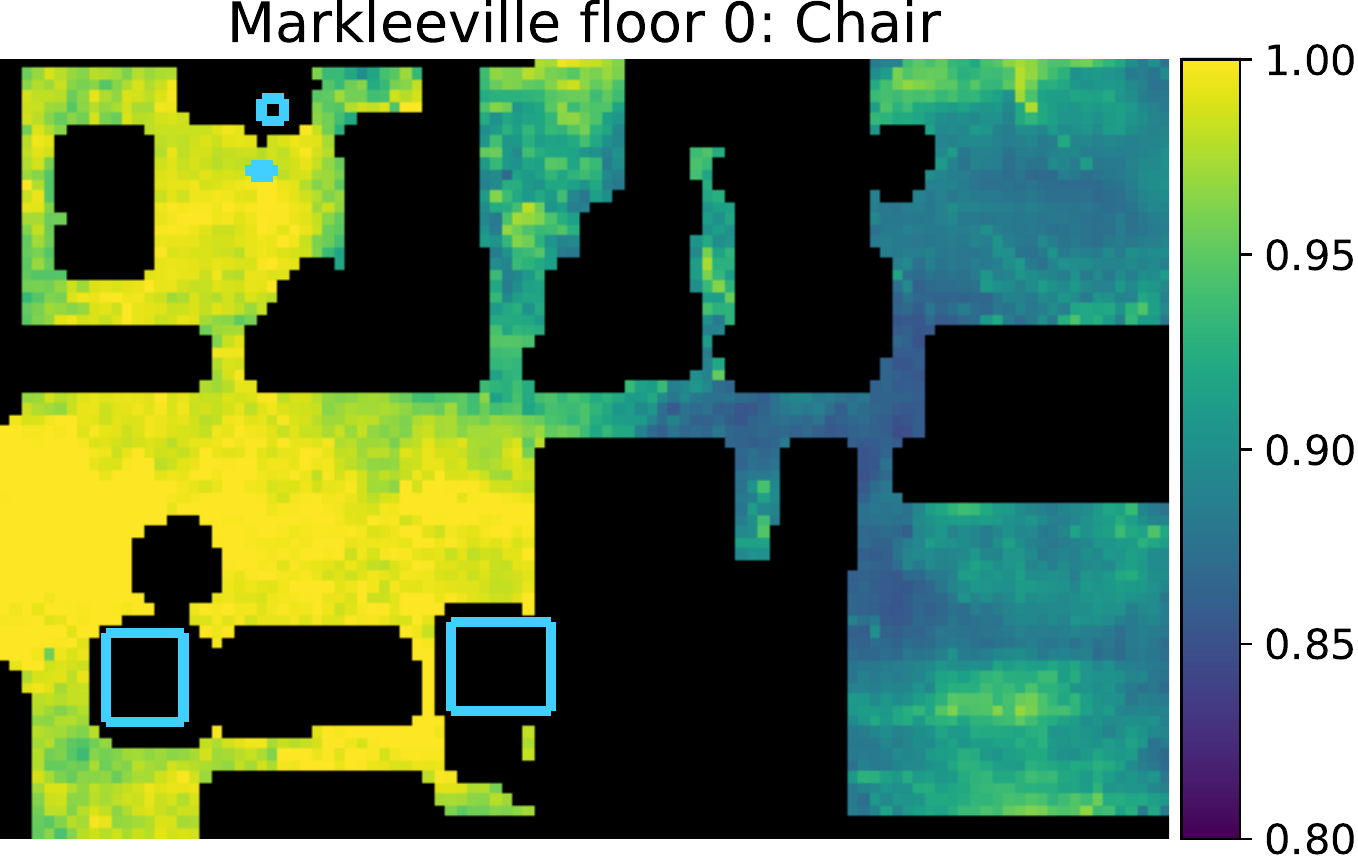} \\
      \insertH{0.32}{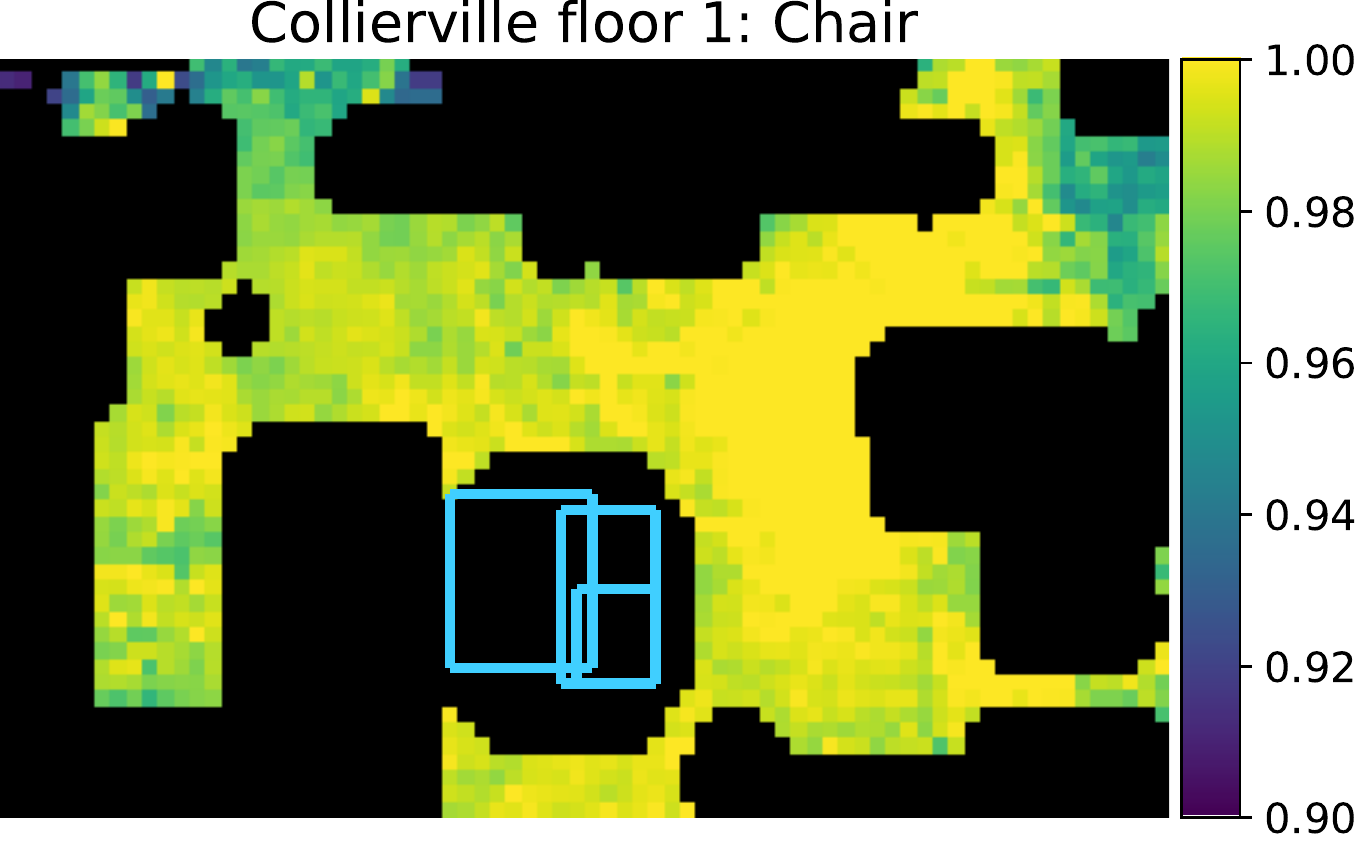}
      \insertH{0.32}{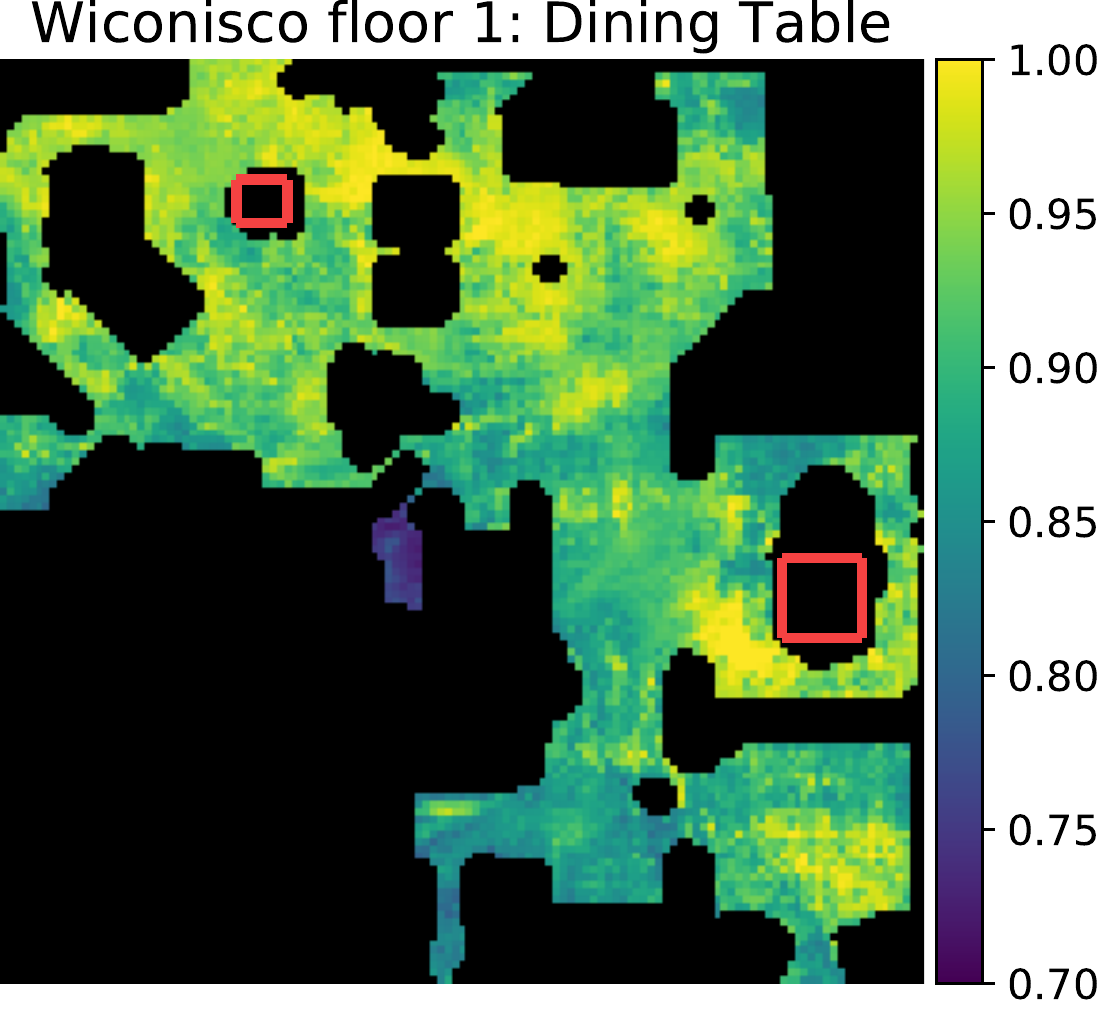}
    % \end{tabular}
    \caption{Maps representing the value of different locations in novel environments as predicted by our method trained on \Vhat{syn}. We can see that high value regions fall off smoothly as the distance from object goals increases.}
    \figlabel{Collierville1}
\end{figure}

% \clearpage

\subsection{Value in Branching Environment}
\begin{figure}[h]
    % \insertW{0.95}{supplementary_figures/branch.pdf}\\
    \insertW{1.00}{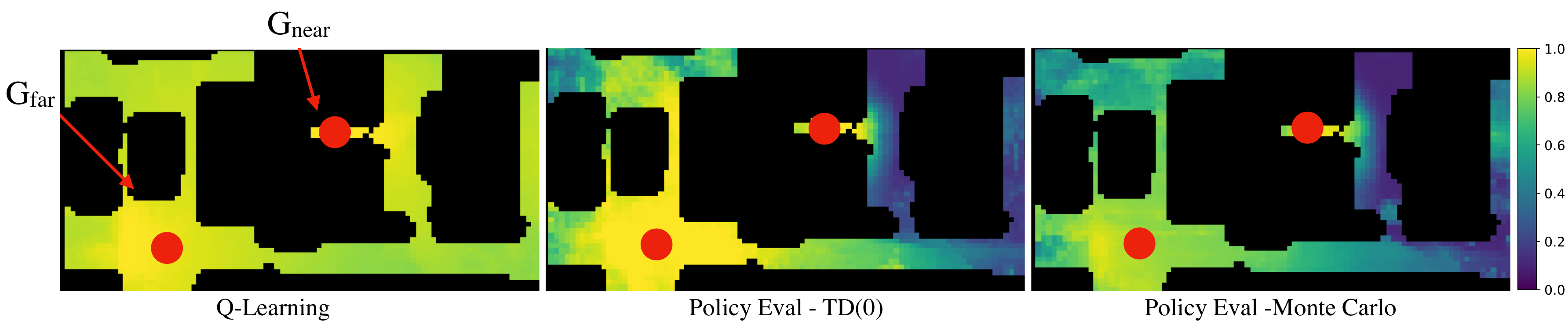}
    \caption{The predicted value in the branching environment using models
    trained with Q-learning, and policy evaluation via TD(0) and Monte Carlo.
    We see that the policy evaluation methods drastically under estimate the
    value in the optimal direction at the branch point. This leads to
    sub-optimal policies for those methods while the Q-learning based value
    function finds the optimal trajectory. See Section 4.3 for
    details.} \figlabel{branch_maps}
  \end{figure}

\end{document}